\documentclass{article}

\usepackage{microtype}
\usepackage{graphicx}
\usepackage{subcaption}
\usepackage{booktabs} 

\usepackage{hyperref}

\usepackage[preprint]{icml2026}

\usepackage{amsmath}
\usepackage{amssymb}
\usepackage{mathtools}
\usepackage{amsthm}

\usepackage[capitalize,noabbrev]{cleveref}

\theoremstyle{plain}

\theoremstyle{definition}

\theoremstyle{remark}

\usepackage[textsize=tiny]{todonotes}

\usepackage{multirow}
\usepackage{graphicx}
\usepackage{subcaption}
\usepackage{fontawesome}
\usepackage{academicons}
\usepackage{pifont}
\usepackage[table]{xcolor}
\usepackage[dvipsnames]{xcolor}
\usepackage{appendix}
\usepackage{stfloats}
\usepackage{wrapfig}
\usepackage{marvosym}

\allowdisplaybreaks[3]

\icmltitlerunning{Multimodal Continual Learning with MLLMs from Multi-scenario Perspectives}

\begin{document}

\twocolumn[
  \icmltitle{Multimodal Continual Learning with MLLMs from Multi-scenario Perspectives}



  \icmlsetsymbol{equal1}{*}
  \icmlsetsymbol{equal2}{\Letter}

  \begin{icmlauthorlist}
    \icmlauthor{Kai Jiang}{equal1,a1,a2}
    \icmlauthor{Siqi Huang}{equal1,a1,a2}
    \icmlauthor{Xiangyu Chen}{a2}
    \icmlauthor{Jiawei Shao}{a2}
    \icmlauthor{Hongyuan Zhang}{equal2,a2,a3}
    \icmlauthor{Ping Luo}{a2,a3}
    \icmlauthor{Xuelong Li}{equal2,a2}
  \end{icmlauthorlist}

  \icmlaffiliation{a1}{Northwestern Polytechnical University}
  \icmlaffiliation{a2}{Institute of Artificial Intelligence (TeleAI) of China Telecom}
  \icmlaffiliation{a3}{University of Hong Kong}

  \icmlcorrespondingauthor{Hongyuan Zhang}{hyzhang98@gmail.com}
  \icmlcorrespondingauthor{Xuelong Li}{xuelong\_li@ieee.org}

  \icmlkeywords{Machine Learning, ICML}

  \vskip 0.3in
]



\printAffiliationsAndNotice{}  

\begin{abstract}
Multimodal large language models (MLLMs) deployed on devices must adapt to continuously changing visual scenarios such as variations in background and perspective, to effectively perform complex visual tasks. 
To investigate catastrophic forgetting under real-world scenario shifts, we construct a multimodal visual understanding dataset (MSVQA), covering four distinct scenarios and perspectives: high-altitude, underwater, low-altitude, and indoor environments.
Furthermore, we propose \textsc{Unifier}~(m\textbf{U}ltimodal co\textbf{N}t\textbf{I}nual learning with MLLMs \textbf{F}rom multi-scenar\textbf{I}o p\textbf{ER}spectives), a continual learning~(CL) framework designed to address visual discrepancies while learning different scenarios. 
Compared to existing CL methods, \textsc{Unifier} enables knowledge accumulation within the same scenario and mutual enhancement across different scenarios via Vision Representation Expansion (VRE) and Vision Consistency Constraint (VCC). Experimental results show that \textsc{Unifier} improves the last-step VQA scores by 2.70\%$\sim$10.62\% and the last-step F{\footnotesize 1} scores by 3.40\%$\sim$7.69\% compared to the state-of-the-art method, QUAD, in 20-step cross-scenario continual learning tasks. MSVQA dataset is available at \url{https://huggingface.co/datasets/Kaij00/MSVQA}.
\end{abstract}

\section{Introduction}
\label{sec:intro}

\begin{figure}[!t]
    \centering
    \includegraphics[width=3.2in]{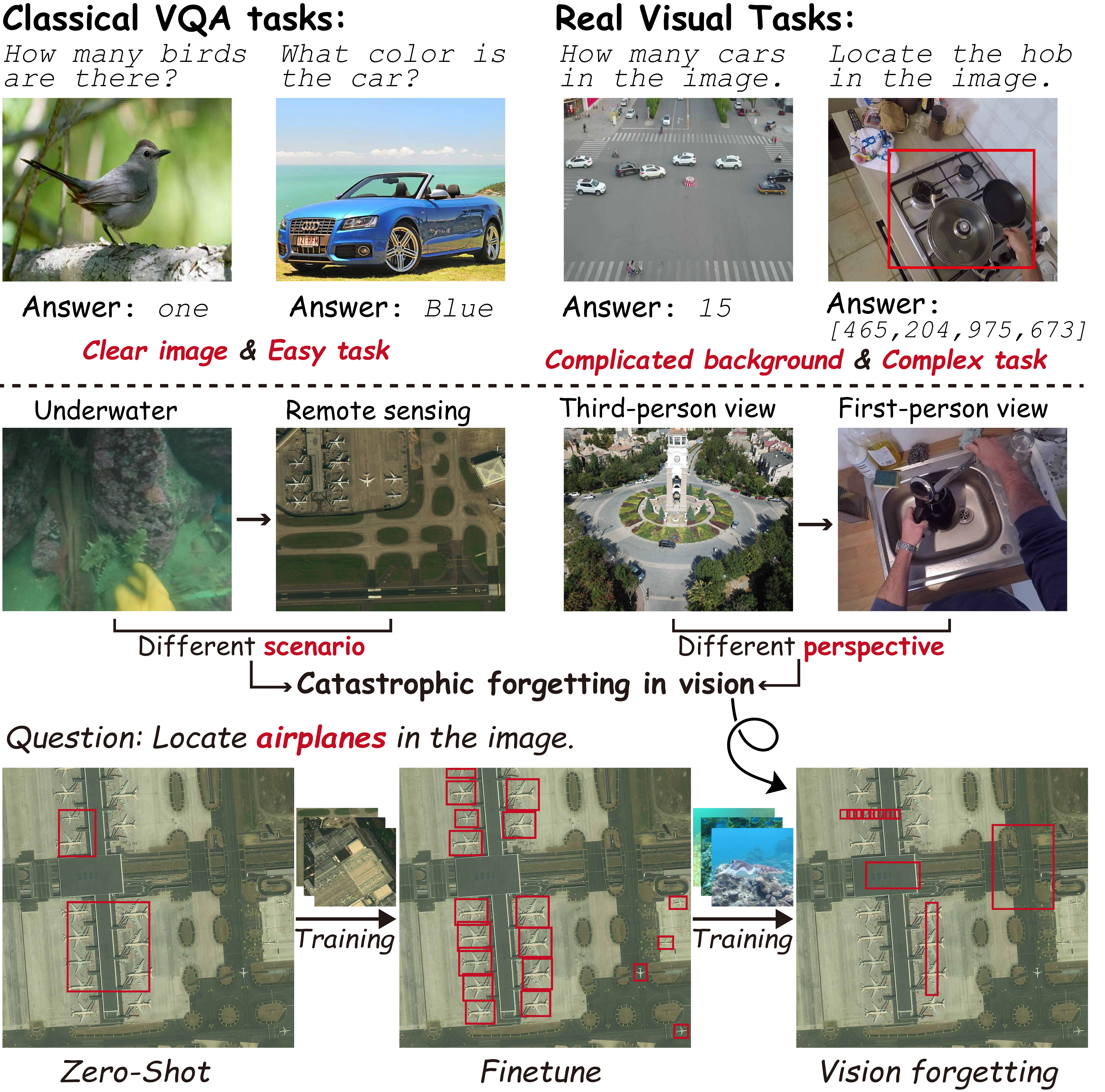}
    \caption{\textbf{Task comparison and vision forgetting}. Classical settings focus on parsing of user intent from text. In real visual tasks, the image background is more complex and the task is more challenging. Untrained model cannot precisely predict the bounding bbox while finetuning with the corresponding scenario can address the issue. But learning a new scenario results in vision forgetting (Severe false positives and false negatives).} \label{fig:motivation}
    \vskip -0.2in
\end{figure}

Real-world data streams are typically collected by various devices across different regions, resulting in diverse visual perspectives and environments conditions.
Consequently, multimodal large language models (MLLMs)~\cite{multimodalsurvey01, multimodalsurvey02} deployed on devices must adapt to continuous scenario shifts in real-world applications~\cite{Catastrophic01, Catastrophic02}. 
However, many existing works \textbf{focus more on language models and neglect visual components}. 
While some continual incremental learning works focus on visual issues~\cite{imageclassificationsurvey01, CLsurvey01, rne}, they are limited to a single modality.
As shown in Fig.~\ref{fig:motivation}, scenario alteration can cause catastrophic forgetting in the visual components.
Therefore, existing continual learning tasks \textbf{lack a benchmark to evaluate performance variations of MLLMs deployed on devices under complex scenario variations}~\cite{continual01, continual02, continual03, continual04}. 
This hinders the development of a unified multimodal large model that can continuously adapt to diverse on-device environments.

For instance, most existing VQA datasets are constructed under fixed scenarios and similar perspectives, where the questions are simple, such as querying the color or quantity of an object. 
In the classical setting~\cite{vqacl}, the core challenge lies in the accurate parsing of user intent from text. 
Such setting neglects \textbf{complicated backgrounds} and \textbf{complex demands} in vision tasks. 
As illustrated in Fig.~\ref{fig:motivation}, images captured by real-world devices typically contains a large number of objects. 
Task-relevant vision information occupies only a small region within the complex backgrounds. 
Moreover, visual tasks usually involve complex requirements, such as identifying the coordinates or categories of a particular target, rather than general information. \textbf{Extracting fine-grained information from the complex background} thus constitutes the core challenge of vision tasks. 
Real-world data streams typically involve varying scenarios and perspectives as environments and sensing devices change.
Such shifts in scenarios or perspectives alter the visual learning behavior of the model,  often leading to catastrophic forgetting in MLLMs.

\begin{figure*}[!t]
    \centering
    \includegraphics[width=6.6in]{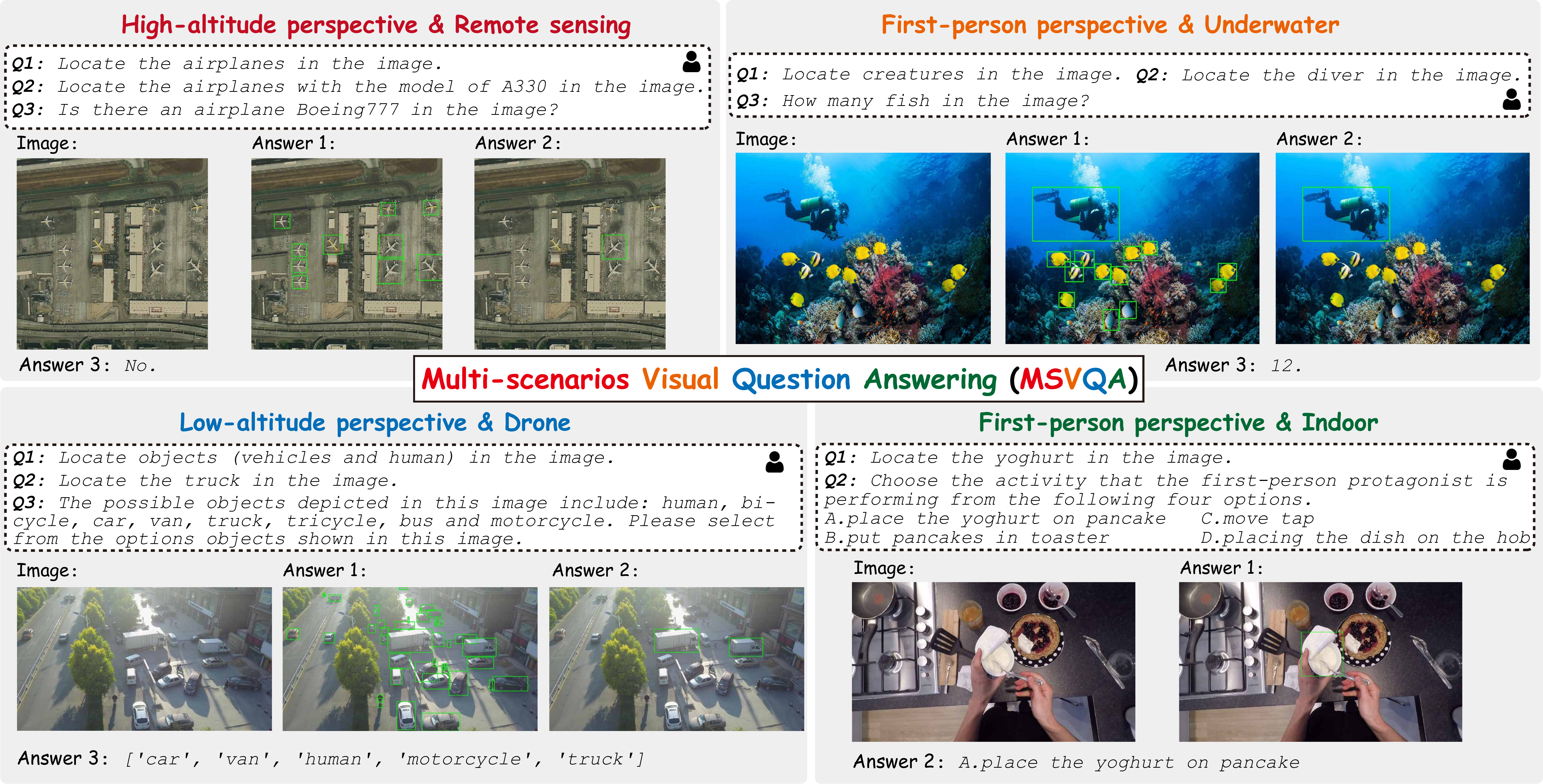}
    \caption{Examples of some image-text pair in the MSVQA dataset. \textbf{More details are provided in Sec.~\ref{sup_sec:dataset}}.} \label{fig:MSVQA}
    \vspace{-5.5pt}
\end{figure*}

\begin{figure}[!t]
    \centering
    \includegraphics[width=3.2in]{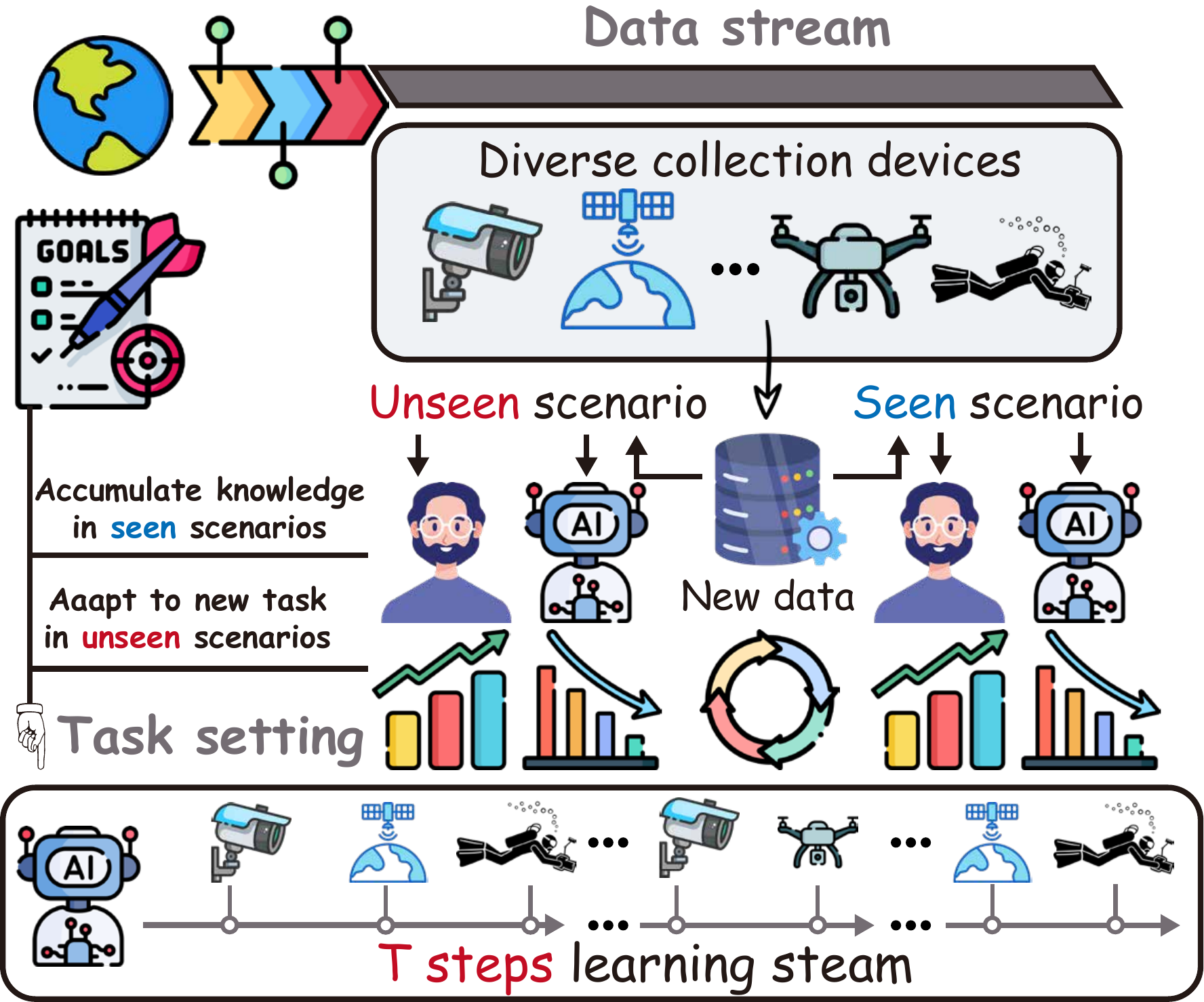}
    \caption{Learning pipeline for MLLMs on devices. Humans can naturally accumulate knowledge in seen scenarios and rapidly adapt to unseen ones. Existing models fail in both cases.} \label{fig:real}
    \vspace{-12pt}
\end{figure}

To this end, \textbf{we construct a Multi-Scenario Visual Question Answering (MSVQA) dataset to evaluate catastrophic forgetting of continual learning} in the shifts of scenarios or perspectives. 
The dataset consists of four distinct scenarios and perspectives, each of which may vary across different locations and time periods.
Across different scenarios, the size and density of objects vary significantly. 
Variations in lighting conditions and occlusions within complex environments further increase the difficulty of MSVQA. 
Therefore, \textbf{the dataset reflects a more realistic real-world data stream}. More details are provided in Sec.~\ref{sup_sec:dataset}. As shown in Fig.~\ref{fig:real}, we aim to construct an MLLM that can continuously adapt to multiple environments, requiring two capabilities: 1) it can \textbf{accumulate knowledge within previously seen scenarios}, with continuously improving performance; 2) it can \textbf{adapt to new tasks in unseen scenarios without forgetting prior knowledge}, ideally enabling \textbf{mutual enhancement across different scenarios}.

However, many existing methods fail to achieve these requirements. 
\textbf{Their performance even degrades when revisiting previously seen scenarios}. 
Moreover, performance on previously learned scenarios often drops rapidly when learning new ones.
To address these challenges,
we propose \textsc{Unifier}(m\textbf{U}ltimodal co\textbf{N}t\textbf{I}nual learning with MLLMs \textbf{F}rom multi-scenar\textbf{I}o p\textbf{ER}spectives).
Expanding LoRA branches to isolate knowledge across scenarios is a natural solution.
However, such models cannot benefit from cross-scenario knowledge sharing and require repeated inference over multiple branches, resulting in inefficiency.
Inspired by multimodal alignment, we introduce additional projection layers to \textbf{re-project features from different branches into a shared subspace, thereby obtaining a unified representation across all scenarios} through a single inference. 
To mitigate catastrophic forgetting, we perform knowledge distillation among various branches. Existing distillation schemes typically align intermediate representations or output logits, however, we find that such constraints significantly limit model plasticity.
Therefore, we design a vision consistency constraint that adopts a softer constraint form that penalizes global changes in the unified representation while aligning representations across scenarios, enabling mutual enhancement of capabilities across different scenes.

The main contributions of our work are three-fold:
\begin{itemize}
    \item We introduce the MSVQA dataset and a corresponding continual learning benchmark to evaluate performance changes of on-device MLLMs under complex scenario variations. The dataset has been open-sourced. The dataset link is omitted due to double-blind review requirements.
    \item We propose \textsc{Unifier} (m\textbf{U}ltimodal co\textbf{N}t\textbf{I}nual learning \textbf{F}rom multi-scenar\textbf{I}o p\textbf{ER}spectives), a framework designed to align visual representations across different scenarios.
    It learns each scenario with a new branch within vision blocks and constrains the features of different scenarios to obtain consistent vision representations, thereby reducing catastrophic forgetting caused by scenario alterations. 
    \item Extensive experiments validated on the MSVQA dataset demonstrate that our approach can alleviate catastrophic forgetting in MLLMs under multi-scenario continual learning, improving the last-step VQA scores by 2.70\%$\sim$10.62\% and the last-step F{\footnotesize 1} score by 3.40\%$\sim$7.69\% in the setting of 20 steps compared to the state-of-the-art approach without additional inference costs.
\end{itemize}

\section{Related Works}
\label{sec:related}

\noindent
\textbf{Multimodal Large Language Models}\quad extend the capabilities of traditional large language models (LLMs) to process and understand multimodal information. They are inspired by representation learning~\cite{bengio2013representation, gu01, pinoise3, pinoise04} such as CLIP~\cite{clip, pinoise6, pinoise7} and integrate specialized encoders to convert various modal inputs to a unified representation. Then, LLMs integrate these unified representations from various modalities and perform complex multimodal reasoning. Existing models such as QwenVL~\cite{qwen2vl} and LLaVA~\cite{LLaVA} have demonstrated that LLMs can handle multimodal information after modal alignment. In this paper, we develop the capabilities of MLLMs in continual learning, to understand new scenarios and respond correctly.

\noindent
\textbf{Continual Learning for MLLMs}\quad aims to develop the capabilities of MLLMs to acquire new data and skills without forgetting. The basic and key challenge lies in catastrophic forgetting due to representation overlap and parameter drift. Some classical CL approaches~\cite{jha2024clap4clip, MiN} penalize parameter drift by restricting the update of key parameters~\cite{ewc} with loss function. Following the same principle, Model Tailor~\cite{tailor} randomly removes updates for most parameters after finetuning and amplifies those for the remaining parameters. Inspired by representation learning, VQACL~\cite{vqacl} introduces a sample-specific and a sample-invariant feature to learn a discriminative and generalizable representation. Inspired by the distillation~\cite{lwf} and rehearsal~\cite{icarl} approaches, QUAD~\cite{quad} only saves all previous questions and uses intermediate attention~\cite{attention} of querying the new image with the old questions for distillation. Although it alleviates forgetting in text modality, the visual discrepancies are ignored across different scenarios.

\section{Methodology}
\label{sec:methodology}
\textbf{Motivation and overview}\quad It is a natural scheme to learn a \textit{single} LoRA\cite{hu2022lora} branch, but it still suffers from severe catastrophic forgetting if we adapt on-device MLLMs to \textit{multiple} scenarios. Therefore, learning multiple LoRA branches to isolate the parameters is an intuitive way to tackle this challenge. However, it also encounters the issue of selecting the correct route. Furthermore, the gating network also encounters catastrophic forgetting if used to select the correct branch, especially without saving previous data. To deal with multiple scenarios while avoiding the need of routing, we propose Vision Representation Expansion (VRE) in Sec.~\ref{subsec:vre} to isolate the parameters for each scenario and then project their representations into a unified feature space. To further avoid drift for new representations, we propose Vision Consistency Constraint (VCC) in Sec.~\ref{subsec:vcc} to align the intermediate representation of branches. Detailed ablation experiments demonstrate that \textbf{VRE outperforms the simple multi-branch (multi-LoRA) one}, and VCC works better on resisting representation drift than strict constraints.

\subsection{Problem Setup}
\label{subsec:problem}
Assume a real-world data stream comprising $T$ tasks, denoted as ${\cal D} = \left\{ {{{\cal D}_1},{{\cal D}_2}, \cdots ,{{\cal D}_T}} \right\}$, with each task data ${{\cal D}_t} = \left\{ {\left( {x_i^t,q_i^t,y_i^t} \right)} \right\}_{i = 1}^{{n_t}}$ collected from different scenarios (e.g., indoor or outdoor) using task-specific methods, such as varying devices and perspectives. $x_i^t \in {{\cal X}_t}$ is the input image, $q_i^t \in {{\cal Q}_t}$ the question, and $y_i^t \in {{\cal Y}_t}$ the corresponding answer. The model is required to learn the current task data ${\cal D}_t$ without forgetting the previous data $\left\{ {{{\cal D}_1},{{\cal D}_2}, \cdots ,{{\cal D}_{T-1}}} \right\}$.

In comparison to the classical setting, this setting introduces the challenge of scenario shift. Cross-scenario tasks require that the model have to \textbf{learn the complex vision information in the new scenario} rather than \textbf{simply reusing the visual abilities} developed in previous tasks. The model should continuously retain the learned knowledge and \textbf{perform increasingly better in the same scenario}. After learning the whole data stream, it is validated on the test set of all scenarios and required to exhibit a performance equivalent to training the model using all data.

\subsection{Vision Representation Expansion} \label{subsec:vre}
Scenario alteration in the real-world data stream requires the MLLMs deployed on devices to learn new representations without forgetting previous ones. Although we can store all textual questions~\cite{quad}, it is impractical to store all images or videos. And visual data has a lower information density and is frequently contaminated with significant noise and irrelevant content. This can lead to overlapping and drifting visual representations when learning different vision data across scenarios, thereby causing catastrophic forgetting. Fig.~\ref{fig:motivation} shows an example of forgetting in the vision representation for MLLMs. Untrained MLLM cannot respond to the user correctly. As illustrated in Fig.~\ref{fig:motivation}, \textbf{relying solely on visual priors tends to lose fine-grained information}, resulting in the omission of small targets. After fine-tuning, the model can locate the target precisely. However, the model cannot correctly extract the visual features of the old scenarios after learning new scenarios. It is caused by the vision representation overlap. The visual representations of old and new scenarios exhibit significant differences, and the representations learned from new scenarios hinder the feature extraction from old scenarios.

Accordingly, our goal is to \textbf{decouple visual representations for different scenarios} in the training process. This process must account for the actual data acquisition conditions, where training data for a single task is typically collected from the same environment, while the test environment is unknown. Inspired by dynamic architecture models~\cite{der, dne, rne}, visual information from different scenarios can be isolated in different model parameters. However, \textbf{expanding the entire backbone is only feasible for small models and is not suitable for MLLMs}. Training with PEFT methods to expand a LoRA branch to learn each task~\cite{ease} also leads to a multiplicative increase in computational overhead during inference.

\begin{figure}[!t]
    \centering
    \includegraphics[width=3.2in]{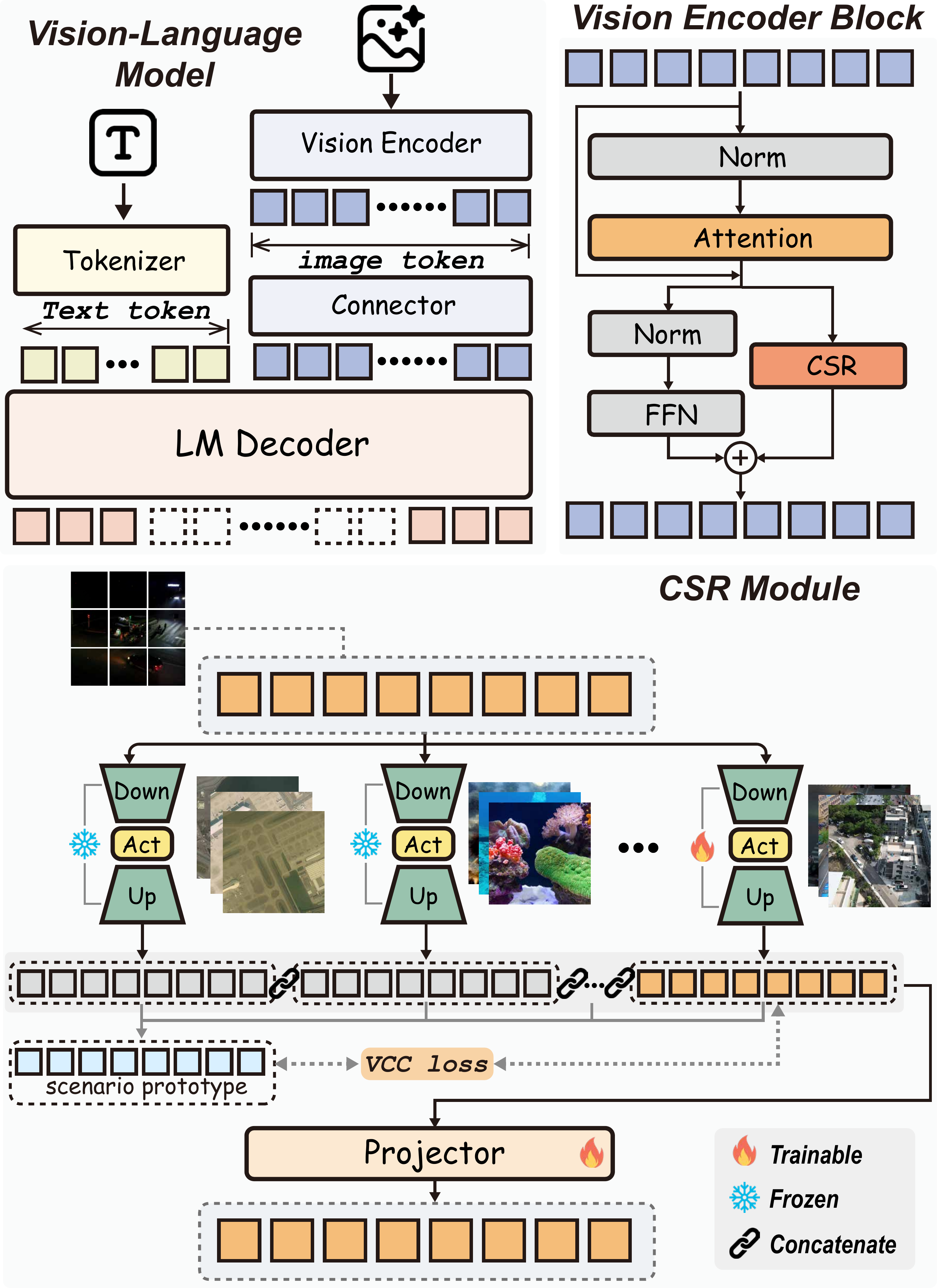}
    \caption{The overview of \textsc{Unifier}. CSR module is added only on the vision encoder to isolate parameters for different scenarios.} \label{fig:overall}
    \vspace{-12pt}
\end{figure}

Therefore, we propose vision representation expansion (VRE) to decouple visual representations for different scenarios without increasing computational load in both training and inference. As illustrated in Fig.~\ref{fig:overall}, a vision-language model consists mainly of a vision encoder and a text encoder. The vision encoder extracts the features of images or videos and then the features are projected on the textual feature space to align with the text embeddings. Thus, new parameters can be expanded in the vision encoder to learn new scenario representations to achieve the isolation of visual information. Assume that a vision encoder ${{\cal F}_v} = \left\{ {{f_1}, \cdots ,{f_L}} \right\}$ consists of $L$ vision blocks ${{f_l}}$, the forward propagation of the vision encoder can be represented as
\begin{align}
    {r_1} &= {f_1}\left( {x_i^t} \right) \label{eq:1}, \\ 
    {r_l} &= {f_l}\left( {{r_{l - 1}}} \right). \label{eq:2}
\end{align}
Each vision block usually consists of an attention module ${A_l}$ and an FFN layer $s_l$. A cross-scenario repressentation~(CSR) module is inserted into the vision block and performs parallel inference with the FFN layer. Then, the visual block can be formulated as
\begin{align}
    {a_l} &= {A_l}\left( {{\rm LN}\left( {{r_{l - 1}}} \right)} \right) + {r_{l - 1}}, \label{eq:3}\\
    {r_l} &= {s_l}\left( {{\rm LN}\left( {{a_l}} \right)} \right) + p_l, \label{eq:4}
\end{align}
where $p_l$ is the output of the CSR module and ${\rm LN}$ is the layer normalization. It consists of multiple branches and a projector to separate different scene information and project them onto the same feature space. It can be represented as
\begin{align}
    {p_l} &= {{\cal P}_l}\left( {\varphi _l^1\left( {{a_l}} \right) \oplus  \cdots  \oplus \varphi _l^K\left( {{a_l}} \right)} \right), \label{eq:5}\\
    \varphi \left(  \cdot  \right) &= {\phi _{\rm up}}\left( {o\left( {{\phi _{\rm down}}\left(  \cdot  \right)} \right)} \right), \label{eq:6}
\end{align}
where $K$ denotes the number of scenarios, ${{\cal P}_l} \in {\mathbb{R}^{K \times {d_1} \to {d_1}}}$ denotes the projector to integrate all representations of various scenarios, ${\phi}_{\rm down}$ denotes the down-projection layer, ${\phi}_{\rm up}$ denotes the up-projection layer and $o$ denotes the activation function. ${\phi _{\rm down}} \in {\mathbb{R}^{{d_1} \to {d_2}}}$ reduces the dimension of the input feature vector from $d_1$ to $d_2$, while ${\phi _{\rm up}} \in {\mathbb{R}^{{d_2} \to {d_1}}}$ restores it to $d_1$, where ${d_1} \gg {d_2}$. As illustrated in Fig.~\ref{fig:overall}, only one branch is trainable in the training of the new scenario, while the parameters of other branches remain frozen. In real-world applications, continuously updated data streams under similar scenarios are more common. Using VRE to \textbf{obtain a unified visual representation has inherent advantages for ambiguous scenarios}, offering better robustness compared to routing-based approaches.

\subsection{Vision Consistency Constraint} \label{subsec:vcc}
Although we reduce the mutual interference of visual information among different scenarios through parameter isolation, \textbf{directly learning new scenarios fails to constrain internal representation}~\cite{pelosin2022towards, wang2022learning}, leading to \textbf{feature drift} and \textbf{destroying the consistency of representations} across tasks. This inconsistency shifts self-attention~\cite{attn01, rne, attn02}, thus the model focuses on an irrelevant visual region to previous scenarios. Some distillation-based CL approaches maintain cross-task consistency by comparing intermediate representations of the new and old models~\cite{podnet, cscct, xu2023multi}. These explicit regularization methods are proven to be effective in aligning internal representations. However, this strict constraint within layers limits the plasticity of the model~\cite{quad}. Especially for images of an unseen scenario, it fails to learn new scenario representations if aligning the intermediate features using strict regularization due to the lack of visual prior. Therefore, maintaining the representation consistency across different scenarios without compromising the model plasticity lies a challenge to learn a new vision representation in continual learning.

To address these challenges, it learns a consistency constraint within the CSR module in each vision block of the model to \textbf{learn a consistent vision representation in different scenarios}. Due to the multi-branch structure, which retains visual scenario information from previous tasks, it can \textbf{align the representations of different branches} without distillation between models. Specifically, scenario-specific vision representations are reserved in $\varphi _l^k$. For a random scenario input, $K$ different vision features $\left\{ {\varphi _l^1\left( {{a_l}} \right), \cdots ,\varphi _l^K\left( {{a_l}} \right)} \right\}$ can be obtained. We can calculate a scenario prototype $\mu _l$ as an explicit multi-scenario unified representation, and then align the representations of each branch with this prototype. 
\begin{align}
    {\mu _l} = \frac{1}{K}\sum\nolimits_{k = 1}^K {\varphi _l^k\left( {{a_l}} \right)} \label{eq:7}
\end{align}
The simplest way constrains the representations of all branches by the ${\ell}^2$-distance between scenario prototype $\mu _l  \in {\mathbb{R}^{{\rm seq} \times d_1}}$ and the representation $\varphi _l^k\left( {{a_l}} \right) \in {\mathbb{R}^{{\rm seq} \times d_1}}$, resulting in the lowest model plasticity. Similarly to knowledge distillation, we compute the relative entropy using soft labels as a constraint instead of ${\ell}^2$-distance. By computing the mean vectors $\bar \varphi _{l}^{k,{\rm fe}}\left( {{a_l}} \right) \in {\mathbb{R}^{{d_1}}}$ and $\bar \varphi _{l}^{k,{\rm em}}\left( {{a_l}} \right) \in {\mathbb{R}^{{\rm seq}}}$ along the feature and embedding channels, we aim to penalize the global changes of representations and allow them to reorganize across the channels:
\begin{align}
    {\cal L}_{c}^{l,k} = {\rm KL}\left( {\frac{{\bar \varphi _l^{k,{\rm fe}}}}{\tau }\left| {\frac{{\bar \mu _l^{\rm fe}}}{\tau }} \right.} \right) + {\rm KL}\left( {\frac{{\bar \varphi _l^{k,{\rm em}}}}{\tau }\left| {\frac{{\bar \mu _l^{\rm em}}}{\tau }} \right.} \right), \label{eq:8}
\end{align}
where ${\bar \mu _l^{\rm fe}}$ and ${\bar \mu _l^{\rm em}}$ denote the mean vectors of the scenario prototype $\mu _l$ along the feature and embedding channels, $\tau$ the distillation coefficient, and ${\rm KL}\left( { \cdot \left|  \cdot  \right.} \right)$ the relative entropy. To prevent parameter drift of projectors, we align the intermediate representations $p_l$ of both the old and new models in the same manner as in Eq.~\eqref{eq:9},
\begin{align}
    {\cal L}_p^l = {\rm KL}\left( {\frac{{\bar p_{l,{\rm new}}^{\rm fe}}}{\tau }\left| {\frac{{\bar p_{l,{\rm old}}^{\rm fe}}}{\tau }} \right.} \right) + {\rm KL}\left( {\frac{{\bar p_{l,{\rm new}}^{\rm em}}}{\tau }\left| {\frac{{\bar p_{l,{\rm old}}^{\rm em}}}{\tau }} \right.} \right) \label{eq:9}
\end{align}
where ${\bar p_{l}^{\rm fe}}$ and ${\bar p_{l}^{\rm em}}$ denote the mean vectors of $p_l$ along the feature and embedding channels. The vision consistency constraint loss ${\cal L}_{vcc}$ is obtained by the averages of ${\cal L}_{c}^{l,k}$ and ${\cal L}_p^{l}$ in all layers:
\begin{align}
    {{\cal L}_{vcc}} = \frac{1}{L}\sum\nolimits_{l = 1}^L {\left( {{\cal L}_p^l + \sum\nolimits_{k = 1}^K {{\cal L}_{c}^{l,k}} } \right)}. \label{eq:10}
\end{align}

\begin{table*}[tbp]
\renewcommand{\arraystretch}{1.0}
\centering
\caption{Average and Last performance comparison with $T=5$ on MSVQA. \textbf{Mem.} denotes the type of exemplar set, including questions~(~\faFileTextO~) and images~(~\faFileImageO~). \ding{56} indicates that no data from previous tasks is saved. Best results are in \textbf{bold}, second-best are \underline{underlined}.}
\label{tab:5steps}
\resizebox{1\linewidth}{!}{
\begin{tabular}{lccccc|cccc|cccc|cccc}
\toprule
\multirow{4}{*}{{\large \textbf{Methods}}} & \multirow{4}{*}{{\large \textbf{Mem.}}} & \multicolumn{4}{c|}{{\textbf{High altitude}}} & \multicolumn{4}{c|}{\textbf{Underwater}} & \multicolumn{4}{c|}{\textbf{Low altitude}} & \multicolumn{4}{c}{\textbf{Indoor}}\\ 
\cmidrule(lr){3-18}
 & & \multicolumn{2}{c}{\textbf{VQA}} & \multicolumn{2}{c|}{\textbf{F{\footnotesize 1}}} & \multicolumn{2}{c}{\textbf{VQA}} & \multicolumn{2}{c|}{\textbf{F{\footnotesize 1}}} & \multicolumn{2}{c}{\textbf{VQA}} & \multicolumn{2}{c|}{\textbf{F{\footnotesize 1}}} & \multicolumn{2}{c}{\textbf{VQA}} & \multicolumn{2}{c}{\textbf{F{\footnotesize 1}}}\\
 \cmidrule(lr){3-18}
 & & $\overline{A}$ & $A_T$ & $\overline{A}$ & $A_T$ & $\overline{A}$ & $A_T$ & $\overline{A}$ & $A_T$ & $\overline{A}$ & $A_T$ & $\overline{A}$ & $A_T$ & $\overline{A}$ & $A_T$ & $\overline{A}$ & $A_T$\\
\midrule[0.5pt]
\rowcolor{blue!5}
Zero-shot & \ding{56} & -- & 20.55 & -- & 23.74 & -- & 19.30 & -- & 35.41 & -- & 14.94 & -- & 14.67 & -- & 52.40 & -- & 49.89\\
\rowcolor{blue!5}
Joint & \ding{56} & -- & 64.97 & -- & 75.41 & -- & 84.27 & -- & 75.32 & -- & 59.80 & -- & 34.33 & -- & 87.20 & -- & 74.93\\
\midrule[0.5pt]
\rowcolor{red!5}
Finetune & \ding{56} & 36.35 & 30.09 & 41.45 & 36.36 & 54.08 & 74.98 & 45.62 & 63.50 & 36.44 & 32.27 & 18.51 & 18.01 & 62.00 & 51.40 & 47.99 & 47.84\\
\rowcolor{red!5}
EWC~\cite{ewc} & \ding{56} & 35.23 & 31.70 & 44.05 & 37.27 & 55.38 & 76.14 & 49.05 & 66.94 & 38.72 & 35.27 & 20.47 & 23.34 & 66.44 & 55.00 & 48.69 & 50.95\\
\rowcolor{red!5}
Tailor~\cite{tailor} & \ding{56} & 38.15 & 33.40 & 43.88 & 35.27 & 55.98 & 76.97 & 47.34 & 67.31 & 37.76 & 34.56 & 20.11 & 20.99 & 63.64 & 53.60 & 49.11 & 50.05\\
\midrule[0.5pt]
\rowcolor{yellow!10}
ER~\cite{er} & \faFileTextO ~+ \faFileImageO & 50.01 & 43.64 & 64.04 & 58.73 & 68.24 & 78.16 & 54.69 & 68.55 & 49.63 & 48.12 & 24.60 & 22.91 & 67.20 & 61.40 & 54.77 & 55.98\\
\rowcolor{yellow!10}
PODNet~\cite{podnet} & \faFileTextO ~+ \faFileImageO & 56.87 & 52.95 & 70.77 & 68.03 & 79.89 & 79.38 & 69.99 & 69.81 & \underline{54.88} & \underline{52.87} & 31.41 & 30.95 & 82.36 & 81.20 & 64.57 & \underline{64.89}\\
\rowcolor{yellow!10}
VQACL~\cite{vqacl} & \faFileTextO ~+ \faFileImageO & 50.42 & 41.22 & 68.02 & 64.71 & 76.22 & 78.60 & 67.09 & 69.29 & 52.58 & 47.21 & 28.50 & 24.65 & 81.70 & 82.00 & 60.21 & 58.95\\
\rowcolor{yellow!10}
QUAD~\cite{quad} & \faFileTextO & \underline{57.83} & \underline{56.59} & \underline{71.64} & \underline{70.33} & \underline{80.43} & \underline{79.62} & \underline{70.61} & \underline{70.61} & 54.58 & 52.15 & \underline{31.54} & \underline{31.17} & \underline{82.76} & \underline{82.80} & \underline{65.27} & 63.60\\
\midrule[0.5pt]
\rowcolor{cyan!10}
\textbf{\textsc{Unifier}} & \ding{56} & \textbf{62.24} & \textbf{63.13} & \textbf{72.86} & \textbf{73.76} & \textbf{82.01} & \textbf{82.20} & \textbf{71.95} & \textbf{72.07} & \textbf{56.75} & \textbf{57.81} & \textbf{32.61} & \textbf{33.16} & \textbf{84.72} & \textbf{85.70} & \textbf{66.59} & \textbf{69.85}\\
\rowcolor{cyan!10}
~~~\textcolor{NavyBlue}{\textbf{($\Delta$\%)}} & \faLineChart & \textcolor{NavyBlue}{\textbf{(+4.41)}} & \textcolor{NavyBlue}{\textbf{(+6.54)}} & \textcolor{NavyBlue}{\textbf{(+1.22)}} & \textcolor{NavyBlue}{\textbf{(+3.43)}} & \textcolor{NavyBlue}{\textbf{(+1.58)}} & \textcolor{NavyBlue}{\textbf{(+2.58)}} & \textcolor{NavyBlue}{\textbf{(+1.34)}} & \textcolor{NavyBlue}{\textbf{(+1.46)}} & \textcolor{NavyBlue}{\textbf{(+1.87)}} & \textcolor{NavyBlue}{\textbf{(+4.94)}} & \textcolor{NavyBlue}{\textbf{(+1.07)}} & \textcolor{NavyBlue}{\textbf{(+1.99)}} & \textcolor{NavyBlue}{\textbf{(+1.96)}} & \textcolor{NavyBlue}{\textbf{(+2.90)}} & \textcolor{NavyBlue}{\textbf{(+1.32)}} & \textcolor{NavyBlue}{\textbf{(+4.96)}}\\
\bottomrule
\end{tabular}
}
\end{table*}
\begin{figure*}[!t]
    \centering
    \begin{subfigure}[b]{0.495\textwidth}
        \centering
        \includegraphics[width=\textwidth]{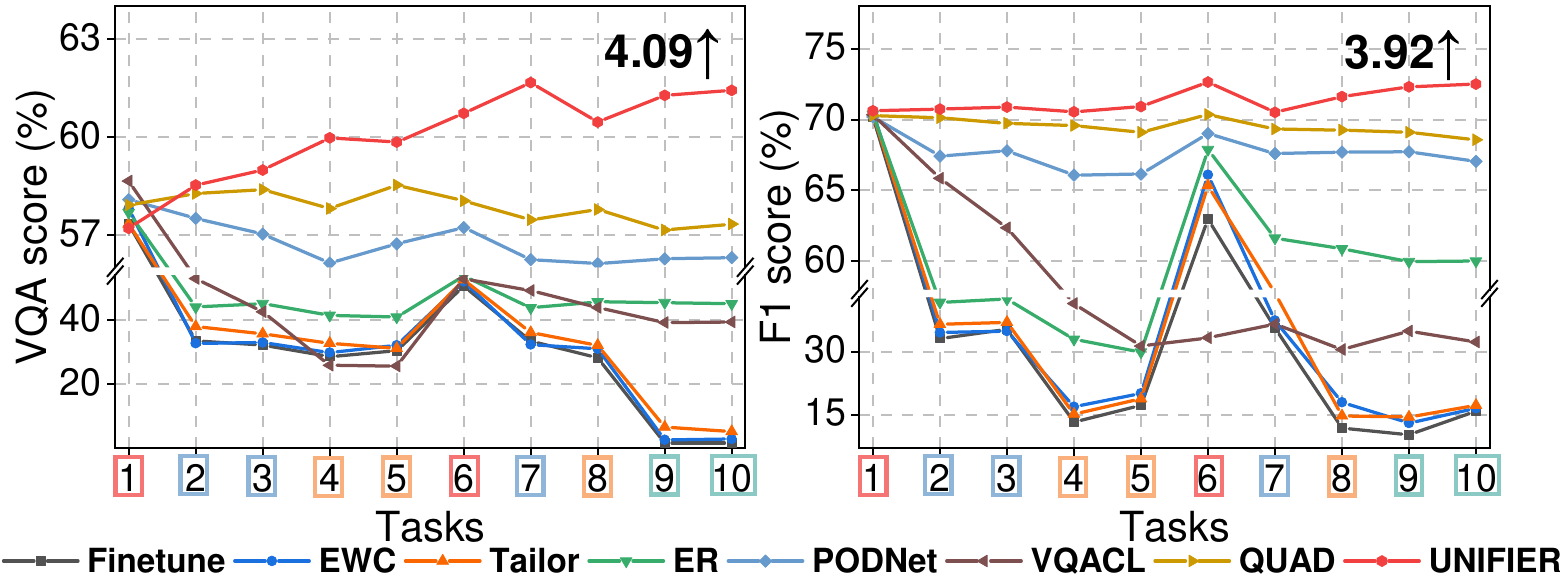}
        \caption{High altitude}
        \label{fig:cmp_sub1}
    \end{subfigure}
    \begin{subfigure}[b]{0.495\textwidth}
        \centering
        \includegraphics[width=\textwidth]{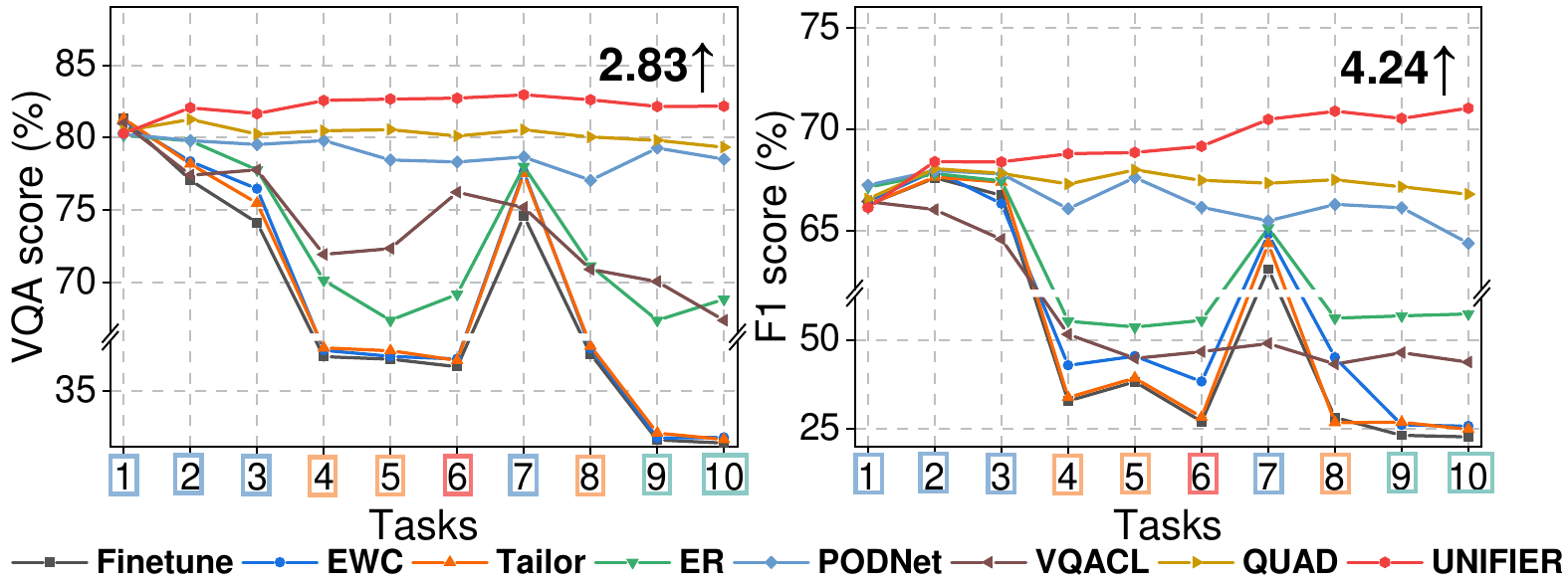}
        \caption{Underwater}
        \label{fig:cmp_sub2}
    \end{subfigure}
    \\
    \begin{subfigure}[b]{0.495\textwidth}
        \centering
        \includegraphics[width=\textwidth]{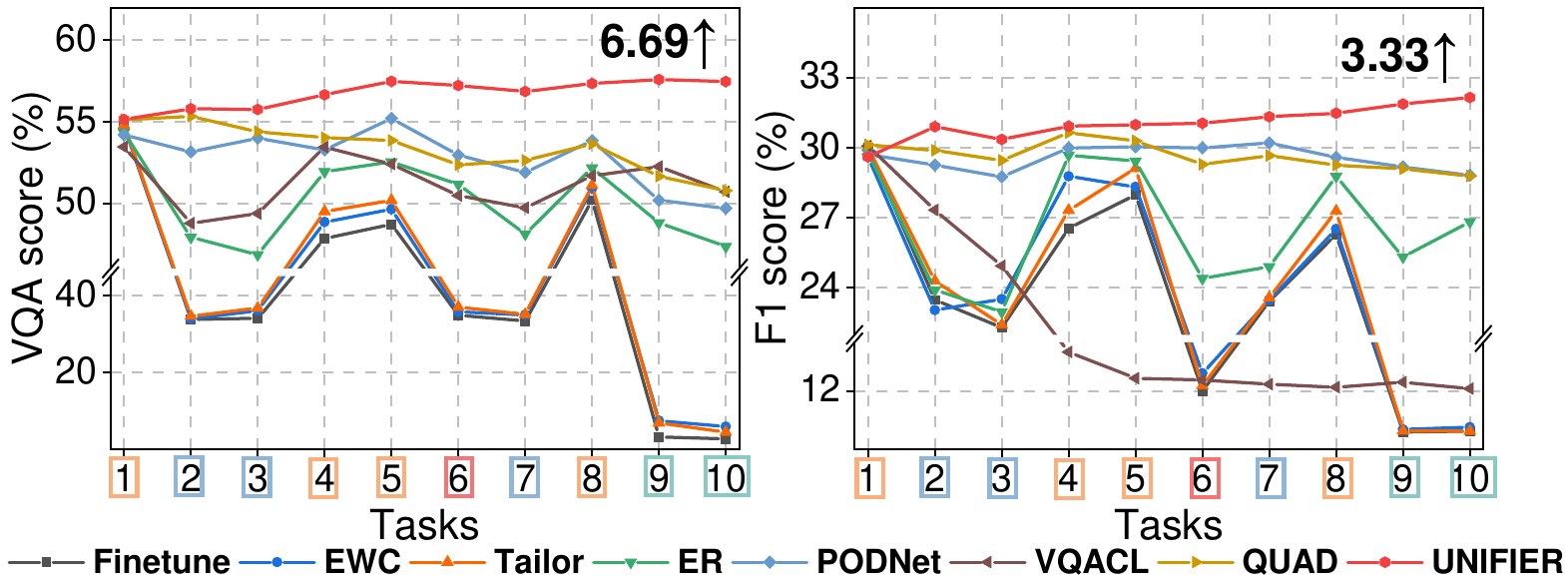}
        \caption{Low altitude}
        \label{fig:cmp_sub3}
    \end{subfigure}
    \begin{subfigure}[b]{0.495\textwidth}
        \centering
        \includegraphics[width=\textwidth]{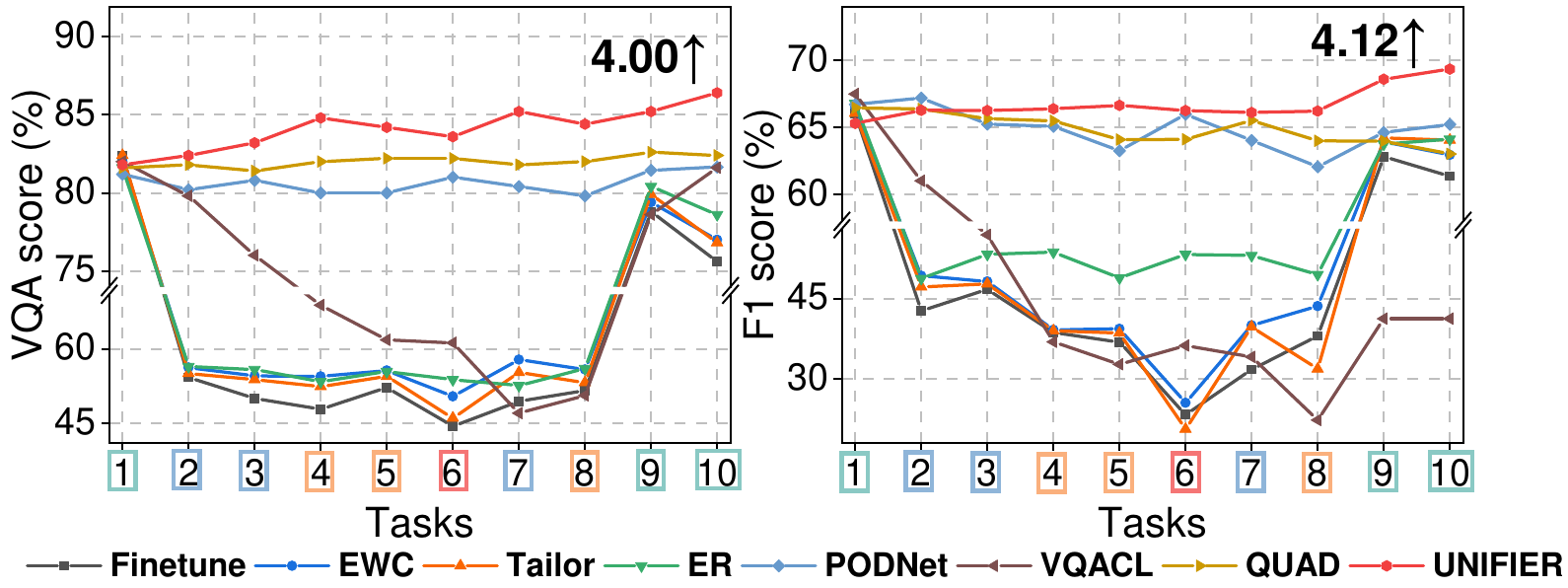}
        \caption{Indoor}
        \label{fig:cmp_sub4}
    \end{subfigure}
    \caption{Incremental trends on 10 steps setting in different scenarios. The performance gap is annotated at the end of each curve. The color of the boxes on the horizontal axis indicates the task scenario: \textcolor{red!50}{High altitude}, \textcolor{blue!50}{Underwater}, \textcolor{orange!50}{Low altitude} and \textcolor{SeaGreen!50}{Indoor}.}
    \label{fig:comparison_10steps}
    \vspace{-12pt}
\end{figure*}

\section{Experiments}
\label{sec:exp}
\subsection{Experimental Settings}
\label{subsec:exp_setting}
\noindent
\textbf{Datasets.}\quad All experiments are conducted on MSVQA datasets, which consist of 4 different scenarios, with each scenario data collected from various perspectives using different devices. The diversity of perspectives and devices forces the model to develop its visual feature extraction capability to perform well in different scenarios. Detailed information can be found in Sec.~\ref{sup_sec:dataset}.

\noindent
\textbf{Protocols.}\quad We split the whole dataset into $T$ tasks, denoted as $T$ steps. Data from each scenario is divided equally into multiple subsets and allocated to individual steps, with each step containing data from only one scenario. We set $T$=5, 10 and 20 to further investigate the performance of all methods at various step sizes. To prevent leakage of image information, we partition the data at the image level as opposed to the annotation level, ensuring that the images in each step are mutually exclusive. The model should learn each task in sequence and be validated in the test set of corresponding scenarios after each step. 

\noindent
\textbf{Evaluation metric.}\quad We incorporate the F{\footnotesize 1} score from object detection~\cite{zou2023object} and combine it with the VQA score for a joint evaluation of MLLM performance in complex visual tasks. Following previous work, let $A_{t}$ denote the VQA score or F{\footnotesize 1} score after learning the task $t$, we use the average performance $\bar A = \left( {1/T} \right)\sum\nolimits_{t = 1}^T {{A_t}}$ across $T$ steps and the last performance $A_T$ as metrics for continual learning. Detailed calculation procedures for all metrics are provided in Sec.~\ref{sup_sec:metric}.

\textbf{Comparison methods.}\quad The proposed approach is compared with six established continual learning methods, which consist of two rehearsal-free approaches: EWC~\cite{ewc} and Tailor~\cite{tailor}, and four rehearsal-based approaches: ER~\cite{er}, PODNet~\cite{podnet}, VQACL~\cite{vqacl}, and QUAD~\cite{quad}. Furthermore, we compare the theoretical performance bounds in continual learning: the lower bound (Finetune), and the upper bound (Joint). We also report the results of zero-shot to demonstrate the necessity of fine-tuning.

\begin{table*}[!t]
\renewcommand{\arraystretch}{1.0}
\centering
\caption{Average and Last performance comparison with $T=20$ on MSVQA. \textbf{Mem.} denotes the type of exemplar set, including questions~(~\faFileTextO~) and images~(~\faFileImageO~). \ding{56} indicates that no data from previous tasks is saved. Best results are in \textbf{bold}, second-best are \underline{underlined}.}
\label{tab:20steps}
\resizebox{1\linewidth}{!}{
\begin{tabular}{lccccc|cccc|cccc|cccc}
\toprule
\multirow{4}{*}{{\large \textbf{Methods}}} & \multirow{4}{*}{{\large \textbf{Mem.}}} & \multicolumn{4}{c|}{{\textbf{High altitude}}} & \multicolumn{4}{c|}{\textbf{Underwater}} & \multicolumn{4}{c|}{\textbf{Low altitude}} & \multicolumn{4}{c}{\textbf{Indoor}}\\ 
\cmidrule(lr){3-18}
 & & \multicolumn{2}{c}{\textbf{VQA}} & \multicolumn{2}{c|}{\textbf{F{\footnotesize 1}}} & \multicolumn{2}{c}{\textbf{VQA}} & \multicolumn{2}{c|}{\textbf{F{\footnotesize 1}}} & \multicolumn{2}{c}{\textbf{VQA}} & \multicolumn{2}{c|}{\textbf{F{\footnotesize 1}}} & \multicolumn{2}{c}{\textbf{VQA}} & \multicolumn{2}{c}{\textbf{F{\footnotesize 1}}}\\
 \cmidrule(lr){3-18}
 & & $\overline{A}$ & $A_T$ & $\overline{A}$ & $A_T$ & $\overline{A}$ & $A_T$ & $\overline{A}$ & $A_T$ & $\overline{A}$ & $A_T$ & $\overline{A}$ & $A_T$ & $\overline{A}$ & $A_T$ & $\overline{A}$ & $A_T$\\
\midrule[0.5pt]
\rowcolor{blue!5}
Zero-shot & \ding{56} & -- & 20.55 & -- & 23.74 & -- & 19.30 & -- & 35.41 & -- & 14.94 & -- & 14.67 & -- & 52.40 & -- & 49.89\\
\rowcolor{blue!5}
Joint & \ding{56} & -- & 64.97 & -- & 75.41 & -- & 84.27 & -- & 75.32 & -- & 59.80 & -- & 34.33 & -- & 87.20 & -- & 74.93\\
\midrule[0.5pt]
\rowcolor{red!5}
Finetune & \ding{56} & 26.96 & 40.67 & 26.90 & 48.52 & 45.56 & 35.40 & 32.17 & 25.14 & 30.04 & 20.30 & 13.87 & 8.80 & 59.36 & 46.40 & 35.99 & 21.27\\
\rowcolor{red!5}
EWC~\cite{ewc} & \ding{56} & 28.19 & 43.41 & 34.02 & 51.41 & 48.64 & 46.29 & 43.67 & 37.88 & 33.75 & 38.43 & 17.96 & 17.71 & 62.89 & 54.80 & 47.42 & 46.46\\
\rowcolor{red!5}
Tailor~\cite{tailor} & \ding{56} & 29.00 & 41.62 & 28.00 & 49.77 & 47.86 & 37.87 & 34.32 & 28.03 & 30.34 & 20.80 & 14.34 & 5.04 & 59.91 & 47.40 & 37.13 & 22.40\\
\midrule[0.5pt]
\rowcolor{yellow!10}
ER~\cite{er} & \faFileTextO ~+ \faFileImageO & 45.04 & 45.02 & 53.48 & 58.48 & 70.64 & 65.24 & 56.13 & 50.53 & 49.92 & 46.13 & 25.28 & 24.00 & 64.21 & 66.80 & 56.18 & 61.56\\
\rowcolor{yellow!10}
PODNet~\cite{podnet} & \faFileTextO ~+ \faFileImageO & 48.30 & \underline{49.49} & 64.14 & \underline{63.35} & 75.61 & 76.75 & \underline{63.14} & \underline{62.04} & \underline{52.62} & \underline{52.42} & 28.67 & \underline{27.62} & 79.30 & 79.20 & \underline{62.29} & \underline{64.47}\\
\rowcolor{yellow!10}
VQACL~\cite{vqacl} & \faFileTextO ~+ \faFileImageO & 42.71 & 48.47 & 31.10 & 25.71 & 75.50 & 74.70 & 42.25 & 35.52 & 51.52 & 51.58 & 12.34 & 7.80 & 70.13 & 78.80 & 34.44 & 32.23\\
\rowcolor{yellow!10}
QUAD~\cite{quad} & \faFileTextO & \underline{50.77} & 47.32 & \underline{65.74} & 62.85 & \underline{77.06} & \underline{77.00} & 62.55 & 57.44 & 49.20 & 44.73 & \underline{29.19} & 25.86 & \underline{80.42} & \underline{80.60} & 54.70 & 50.14\\
\midrule[0.5pt]
\rowcolor{cyan!10}
\textbf{\textsc{Unifier}} & \ding{56} & \textbf{59.20} & \textbf{60.11} & \textbf{68.17} & \textbf{69.39} & \textbf{78.12} & \textbf{80.79} & \textbf{69.06} & \textbf{69.73} & \textbf{54.36} & \textbf{55.12} & \textbf{30.71} & \textbf{31.02} & \textbf{82.44} & \textbf{83.60} & \textbf{66.51} & \textbf{69.23}\\
\rowcolor{cyan!10}
~~~\textcolor{NavyBlue}{\textbf{($\Delta$\%)}} & \faLineChart & \textcolor{NavyBlue}{\textbf{(+8.43)}} & \textcolor{NavyBlue}{\textbf{(+10.62)}} & \textcolor{NavyBlue}{\textbf{(+2.43)}} & \textcolor{NavyBlue}{\textbf{(+6.04)}} & \textcolor{NavyBlue}{\textbf{(+1.06)}} & \textcolor{NavyBlue}{\textbf{(+3.79)}} & \textcolor{NavyBlue}{\textbf{(+5.92)}} & \textcolor{NavyBlue}{\textbf{(+7.69)}} & \textcolor{NavyBlue}{\textbf{(+1.74)}} & \textcolor{NavyBlue}{\textbf{(+2.70)}} & \textcolor{NavyBlue}{\textbf{(+1.52)}} & \textcolor{NavyBlue}{\textbf{(+3.40)}} & \textcolor{NavyBlue}{\textbf{(+2.02)}} & \textcolor{NavyBlue}{\textbf{(+3.00)}} & \textcolor{NavyBlue}{\textbf{(+4.22)}} & \textcolor{NavyBlue}{\textbf{(+4.76)}}\\
\bottomrule
\end{tabular}
}
\end{table*}

\begin{figure*}[!t]
    \centering
    \begin{subfigure}[b]{0.495\textwidth}
        \centering
        \includegraphics[width=\textwidth]{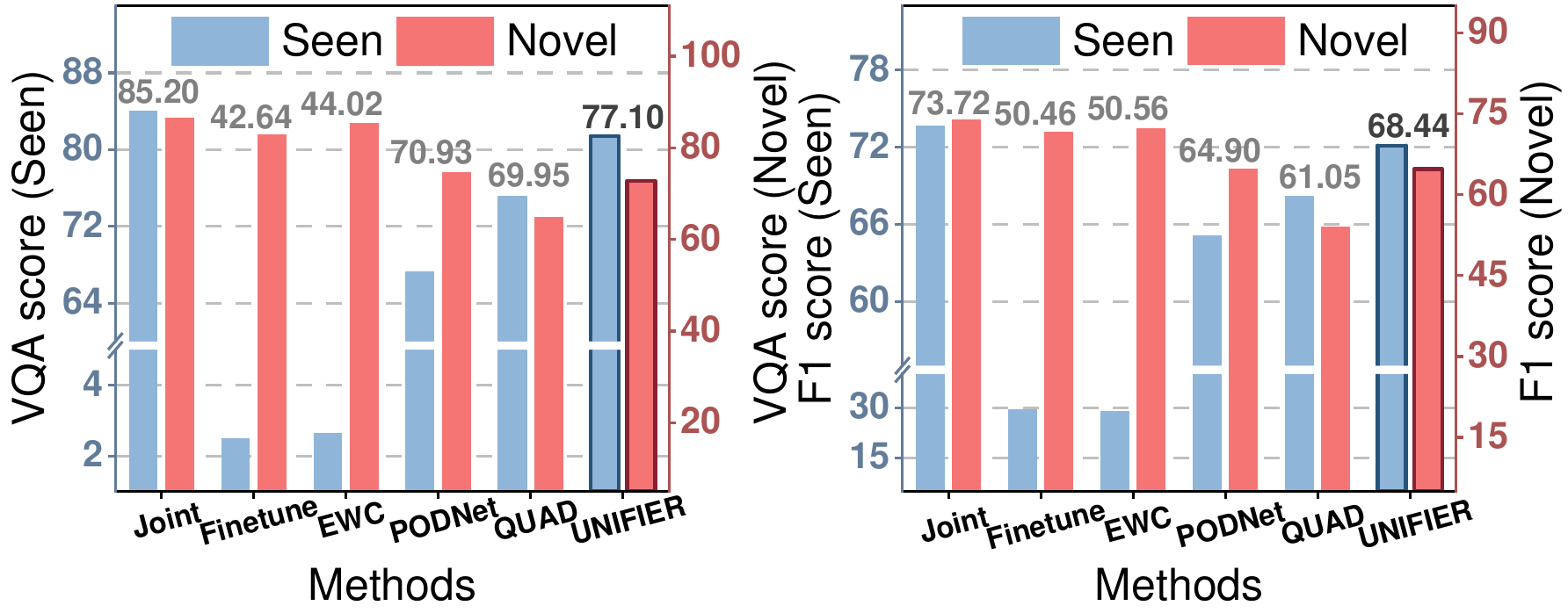}
        \caption{Underwater$\rightarrow$Indoor}
        \label{fig:crs_sub1}
    \end{subfigure}
    \begin{subfigure}[b]{0.495\textwidth}
        \centering
        \includegraphics[width=\textwidth]{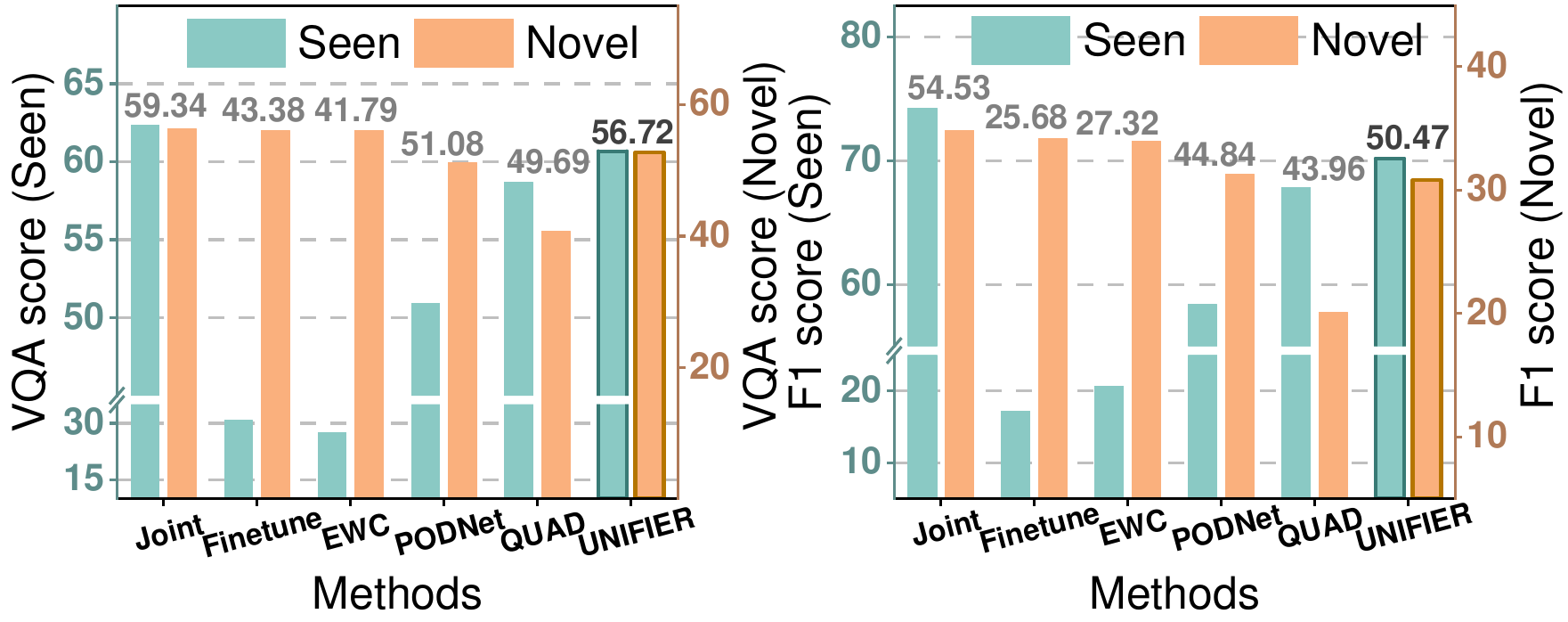}
        \caption{High altitude$\rightarrow$Low altitude}
        \label{fig:crs_sub2}
    \end{subfigure}
    \\
    \begin{subfigure}[b]{0.495\textwidth}
        \centering
        \includegraphics[width=\textwidth]{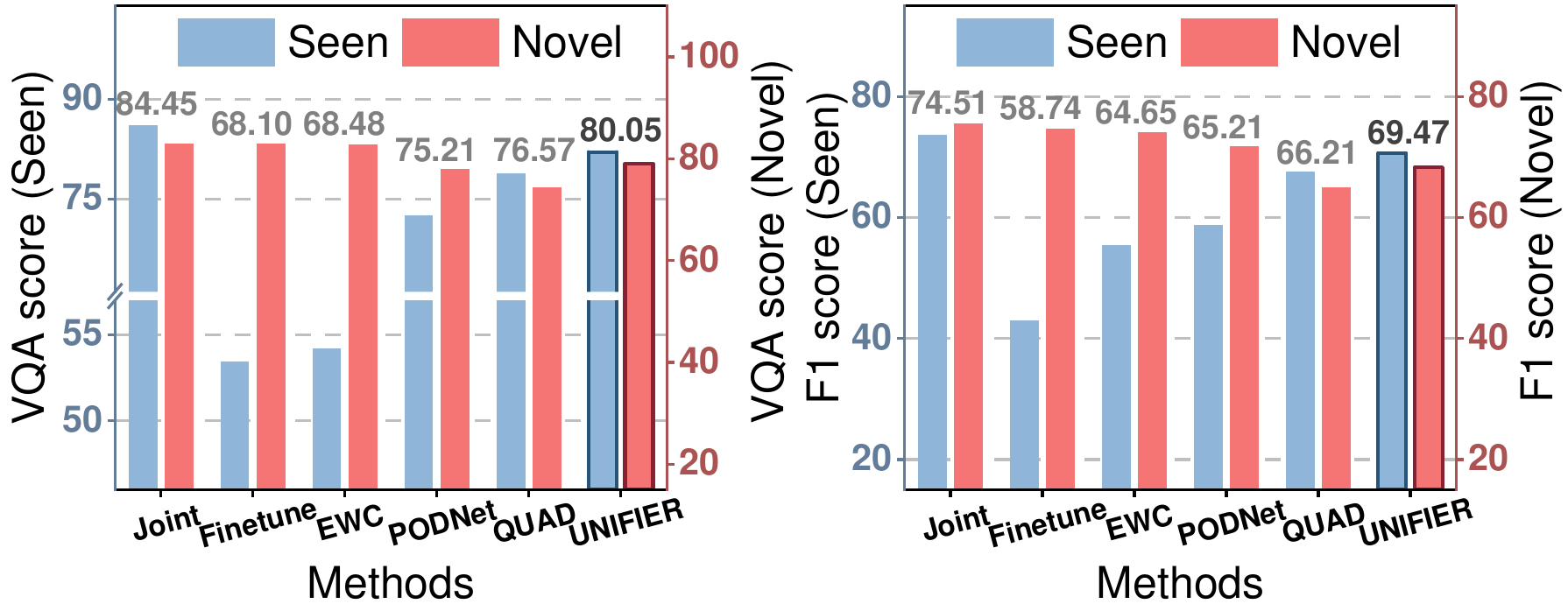}
        \caption{Indoor$\rightarrow$Underwater}
        \label{fig:crs_sub3}
    \end{subfigure}
    \begin{subfigure}[b]{0.495\textwidth}
        \centering
        \includegraphics[width=\textwidth]{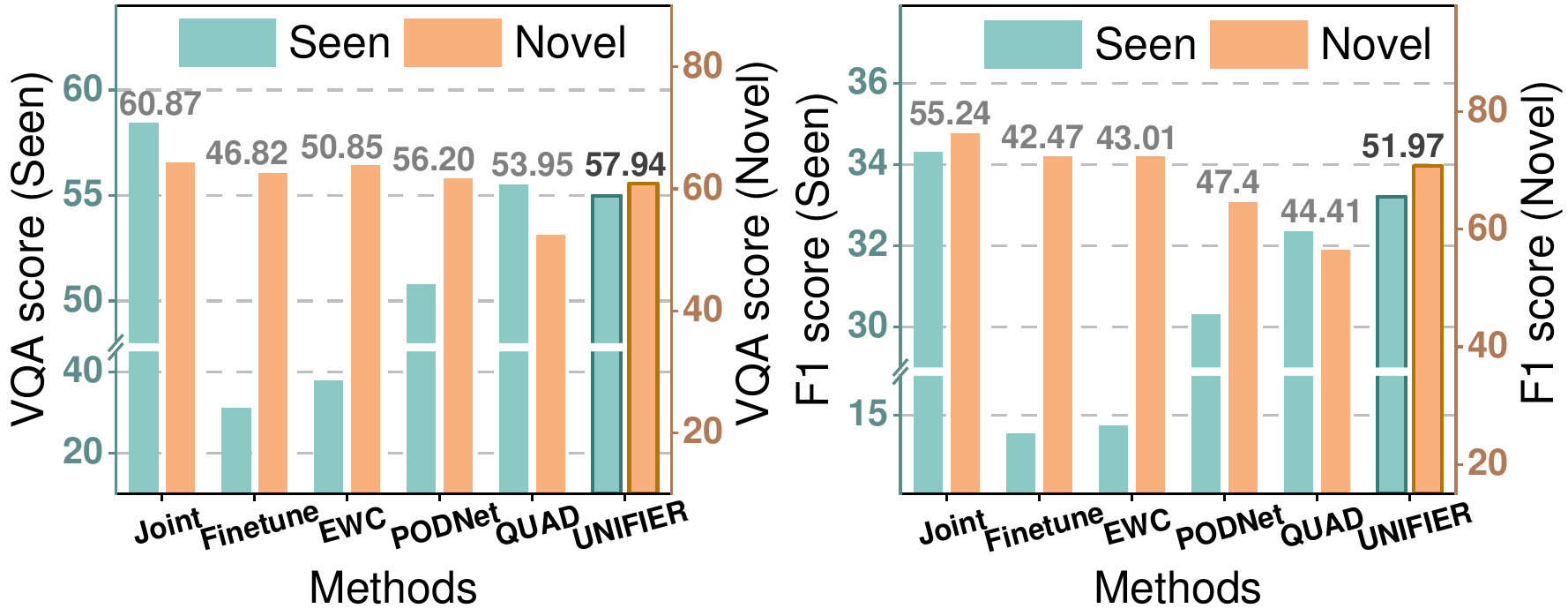}
        \caption{Low altitude$\rightarrow$High altitude}
        \label{fig:crs_sub4}
    \end{subfigure}
    \caption{Performance comparison in scenario alteration. Average score of each method in the seen and novel scenarios is annotated on the top of each bar.}
    \label{fig:crs}
    \vspace{-7.35pt}
\end{figure*}

\textbf{Implementation details.}\quad To facilitate deployment on mobile platforms, we use Qwen2.5VL-3B in our experiments. We utilize the AdamW optimizer with an initial learning rate of $10^{-5}$. The batch size is set to 64 and is allocated to 8 Nvidia H200 GPUs. For each method, the model is trained for 20 epochs in the initial task and 10 epochs in each subsequent task.

\subsection{Main Results}
\label{subsec:main_res}
\textbf{Comparison in standard settings.}\quad Tab.~\ref{tab:5steps} and Tab.~\ref{tab:20steps} compare various methods with $T=5$ and $T=20$. The \textsc{Unifier} achieves superior performance in both settings. For $T=5$, it outperforms the runner-up approach by 6.54\% on last-step VQA score in high altitude scenario and by 4.96\% on the last-step F{\footnotesize 1} score in indoor scenario. For $T=20$, it outperforms the runner-up approach by 10.62\% on the last-step VQA score in high altitude scenario and by 7.69\% on the last-step F{\footnotesize 1} score in underwater scenario. Fig.~\ref{fig:comparison_10steps} shows the incremental trends in different scenarios with $T=10$. Most compared approaches show a significant decline in performance when learning new scenario. In contrast, \textsc{Unifier} improves its performance in both seen and unseen scenarios. It indicates that \textbf{\textsc{Unifier} achieves knowledge accumulation without forgetting} across scenarios.

\begin{table*}[!t]
\renewcommand{\arraystretch}{0.85}
\centering
\caption{Ablation study with $T=10$ on MSVQA.}
\label{tab:ablation}
\resizebox{1\linewidth}{!}{
\begin{tabular}{lcccc|cccc|cccc|cccc}
\toprule
\multirow{4}{*}{{\large \textbf{Methods}}} & \multicolumn{4}{c|}{{\textbf{High altitude}}} & \multicolumn{4}{c|}{\textbf{Underwater}} & \multicolumn{4}{c|}{\textbf{Low altitude}} & \multicolumn{4}{c}{\textbf{Indoor}}\\ 
\cmidrule(lr){2-17}
 & \multicolumn{2}{c}{VQA} & \multicolumn{2}{c|}{F{\footnotesize 1}} & \multicolumn{2}{c}{VQA} & \multicolumn{2}{c|}{F{\footnotesize 1}} & \multicolumn{2}{c}{VQA} & \multicolumn{2}{c|}{F{\footnotesize 1}} & \multicolumn{2}{c}{VQA} & \multicolumn{2}{c}{F{\footnotesize 1}}\\
 \cmidrule(lr){2-17}
 & $\overline{A}$ & $A_T$ & $\overline{A}$ & $A_T$ & $\overline{A}$ & $A_T$ & $\overline{A}$ & $A_T$ & $\overline{A}$ & $A_T$ & $\overline{A}$ & $A_T$ & $\overline{A}$ & $A_T$ & $\overline{A}$ & $A_T$\\
\midrule[0.5pt]
\rowcolor{red!5}
Baseline & 29.74 & 1.60 & 30.54 & 15.82 & 53.26 & 2.48 & 43.62 & 22.77 & 34.33 & 2.62 & 19.94 & 3.72 & 58.64 & 75.60 & 44.80 & 61.38\\
\midrule[0.5pt]
\rowcolor{yellow!10}
w/ VRE & 51.03 & 51.82 & 53.46 & 54.77 & 75.36 & 74.07 & 56.16 & 55.45 & 52.07 & 48.28 & 25.15 & 24.33 & 81.52 & 82.00 & 61.95 & 62.91\\
\rowcolor{yellow!10}
w/ VRE+${\ell}^2$-dis. & 53.38 & 53.47 & 59.94 & 58.79 & 76.49 & 75.16 & 57.47 & 57.45 & 52.34 & 49.18 & 25.62 & 24.86 & 82.62 & 82.75 & 63.05 & 64.27\\
\rowcolor{yellow!10}
w/ VRE+${\ell}^2$-spa. & 56.29 & 56.29 & 67.08 & 65.36 & 78.23 & 78.15 & 65.89 & 65.98 & 52.60 & 49.32 & 28.93 & 28.46 & 82.84 & 83.03 & 63.24 & 64.57\\
\rowcolor{yellow!10}
w/ VRE+KL-spa. & 59.49 & 60.46 & 70.59 & 70.57 & 81.14 & 81.05 & 68.47 & 69.70 & 55.25 & 55.51 & 29.53 & 30.23 & 81.73 & 83.79 & 64.87 & 66.62\\
\rowcolor{yellow!10}
w/o Projector & 59.07 & 60.43 & 70.41 & 71.24 & 81.28 & 81.01 & 68.49 & 70.09 & 55.64 & 55.98 & 30.06 & 31.30 & 83.18 & 85.44 & 65.88 & 68.73\\
\midrule[0.5pt]
\rowcolor{cyan!10}
\textbf{w/ VRE+VCC} & \textbf{60.00} & \textbf{61.42} & \textbf{71.32} & \textbf{72.49} & \textbf{82.18} & \textbf{82.16} & \textbf{69.26} & \textbf{71.02} & \textbf{56.73} & \textbf{57.47} & \textbf{31.06} & \textbf{32.15} & \textbf{84.12} & \textbf{86.40} & \textbf{66.75} & \textbf{69.36}\\
\bottomrule
\end{tabular}
}
\end{table*}

\begin{figure*}[!t]
    \centering
    \includegraphics[width=6.5in]{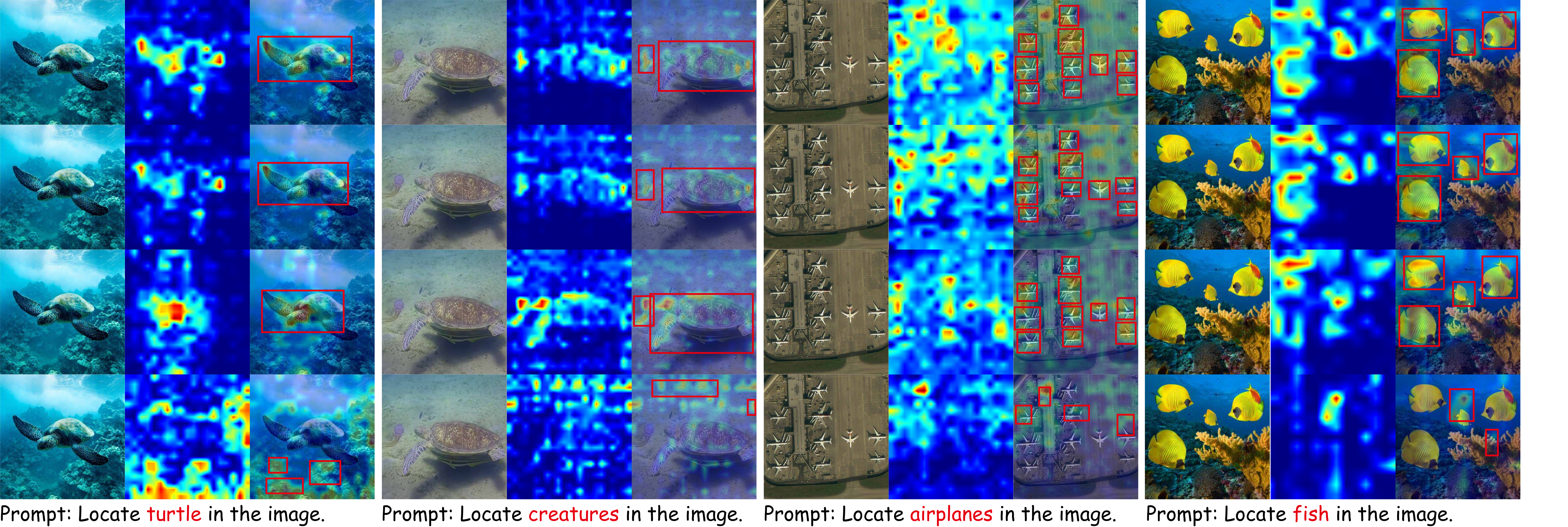}
    \caption{Visualization comparisons between \textsc{Unifier} and Finetune. Row 1 and 2 of each group denote the visualization comparisons of \textsc{Unifier} before and after it learns a new scenario. Row 3 and 4 of each group denote the visualization comparisons of Finetune before and after it learns a new scenario.} \label{fig:vis}
\end{figure*}

\noindent
\textbf{Comparison in scenario alteration}\quad We further compare the model stability and plasticity in learning new scenarios. The model undergoes sequential training on two different scenarios. We evaluate its performance in both seen and novel scenarios. Fig.~\ref{fig:crs} shows that \textsc{Unifier} achieves a higher average score on both scenarios compared to other methods. It demonstrates that \textbf{\textsc{Unifier} achieves a better trade-off between the seen and novel scenarios}.

\subsection{Ablation Study and Analysis}
\label{subsec:ablation_analysis}
\textbf{Ablation study.}\quad We perform a detailed ablation study in Tab.~\ref{tab:ablation}. Row~1 reports the results of the baseline, which fine-tunes the model without any strategies. Row~2 adds a VRE structure on the baseline and Row~7 adds the VCC constraint on Row~2. Comparison of rows~1, 2 and 7 illustrates that each component of \textsc{Unifier} effectively improves the model’s ability to resist forgetting.
Rows~3 to 6 show some alternatives to the structure or loss design. Row~3 is the simplest design as an alternative to Eq.~\eqref{eq:8}, which constrains the intermediate representations using ${\ell}^2$-distance without channel reduction. And row~4 uses a channel reduction based on row~3 to penalize only global changes. The comparison shows that \textbf{looser constraints better balance between stability and plasticity}. Row~5 outperforms row~4 by replacing ${\ell}^2$-distance with relative entropy to use soft logits to constrain the representations of different scenarios. Row~6 shows the results of removing the projector. Although feature alignment maintains the representation consistency between the new and old models, simple feature addition constrains the model capacity to learn the new scenario. The comparison shows that \textbf{the projection layer is effective}.

\noindent
\textbf{Visualizations.}\quad Fig.~\ref{fig:vis} shows the visualizations of \textsc{Unifier} and Finetune. Comparison between rows~1 and 2 of each group demonstrates that \textsc{Unifier} maintains a stable representation of the old scenario throughout learning a new scenario, with attentional shifts confined to only a few regions. Therefore, the proposed method \textbf{exhibits only minor missed detections and bounding box offsets after adaptation to new scenes}. However, visualization of the Finetune shows that \textbf{its attention undergoes a significant shift to some unrelated areas after learning a new scenario}, which leads to a large number of false alarms and missed detections.

\noindent
\textbf{Complexity analysis.}\quad For autoregressive generative MLLMs, the main factor affecting model inference efficiency lies in the text decoder. Therefore, \textbf{adding the VRE module does not impact the inference efficiency}. We provide detailed experiments in Sec.~\ref{sup_sec:complexity} analyzing the inference complexity and parameter count. The results show that \textbf{the proposed method incurs no additional inference cost}, and the \textbf{increase in parameters is entirely acceptable}. In addition, we also try various foundation model in different size and report the results in Sec.~\ref{sup_sec:upper} and~\ref{sup_sec:qwen3vl}.

\section{Conclusion}
\label{sec:conclusion}
Existing investigation on continual learning overlooks visual forgetting for MLLMs. Therefore, we construct a multimodal visual understanding datatset (MSVQA) to evaluate continual learning ability for MLLMs deployed on devices in scenario alteration. Then, we propose \textsc{Unifier} to obtain a unified representation for different scenarios, and align the representations from all individual branches with a soft consistency constraint. Experimental results demonstrate that \textbf{\textsc{Unifier} achieves knowledge accumulation in data streams containing various scenarios}.

\section*{Acknowledgement}
This paper is partially supported by the National Key R\&D Program of China No.2022ZD0161000.

\section*{Impact Statement}
This paper presents work whose goal is to advance the field of Machine Learning. There are many potential societal consequences of our work, none which we feel must be specifically highlighted here.

\bibliography{example_paper}
\bibliographystyle{icml2026}

\newpage
\appendix
\onecolumn
\setcounter{page}{1}
\setcounter{figure}{0}
\renewcommand*{\thefigure}{A\arabic{figure}}
\renewcommand*{\thetable}{A\arabic{table}}
\setcounter{equation}{0}
\renewcommand*{\theequation}{A\arabic{equation}}
\begin{appendices}
\section*{Appendix}
\section{Overview} \label{Appendix}
In this supplementary material, we provide more details about \textsc{Unifier}. We first introduce details of the multimodal visual understanding dataset MSVQA~(in Sec.~\ref{sup_sec:dataset}) and provide the corresponding evaluation metrics~(in Sec.~\ref{sup_sec:metric}). Then, we provide descriptions of all comparison methods~(in Sec.~\ref{sup_sec:cmp}). Moreover, we provide the hardware information for all experiments~(in Sec.~\ref{sup_sec:hardware}) and summarize the computational resource requirements with complexity analysis~(in Sec.~\ref{sup_sec:complexity}).
Furthermore, we compare the upper bounds of Qwen2.5VL-3B, Qwen2.5VL-7B and Qwen3VL-4B to investigate the influence of different model sizes and foundation models in Sec.~\ref{sup_sec:upper}, which detailed experiments on setting of 10 steps using Qwen3VL-4B are performed in Sec.~\ref{sup_sec:qwen3vl}. 
Finally, we provide complete comparison experiments~(in Sec.~\ref{sup_sec:complete_cmp}) and complete scenario alteration experiments~(in Sec.~\ref{sup_sec:scenario_alteration}). In addition, explained in Sec.~\ref{sup_sec:order} that the effect of learning order is eliminated.

\begin{figure*}[bp]
    \centering
    \includegraphics[width=6.8in]{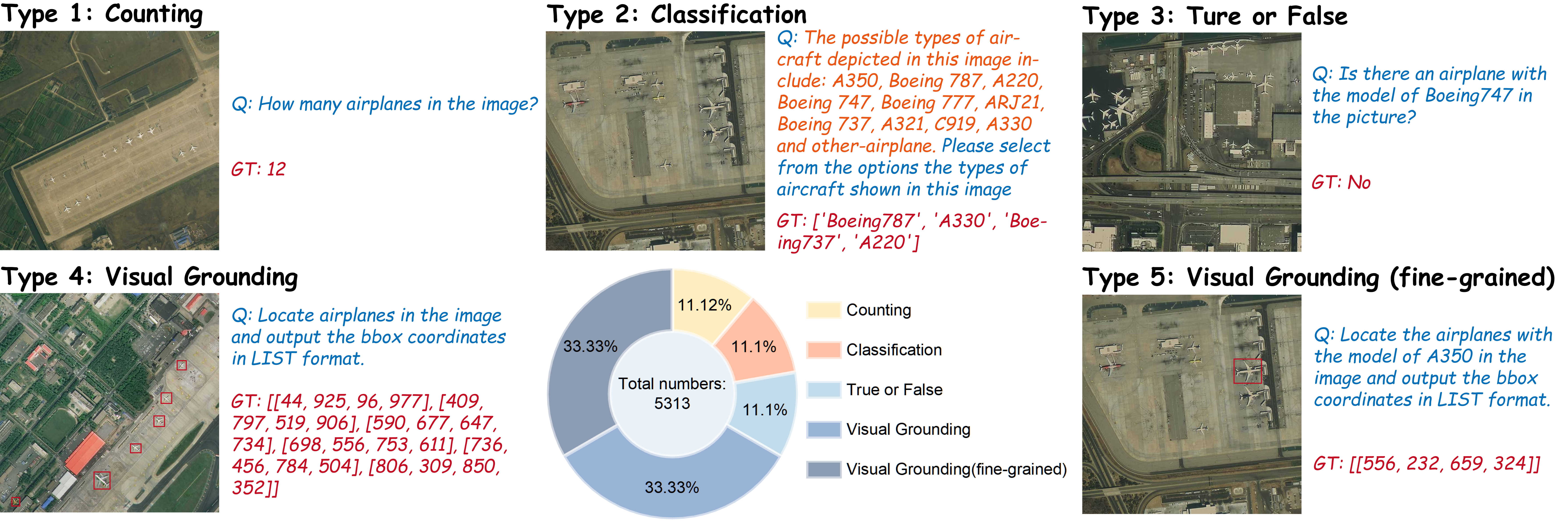}
    \caption{Illustrations of high altitude data.} \label{sup_fig:high_data}
\end{figure*}

\section{Dataset details} \label{sup_sec:dataset}
\subsection{High altitude.}

\noindent
The source data of high altitude scenario are from a large-scale optical remote sensing dataset~\cite{2022FAIR1M}.

\noindent
\textbf{Data Preprocess}: We mainly execute four operations.
\begin{enumerate}
    \item Separate image containing airports from the source data.
    
    \item Perform sliding window cropping on high-resolution images to ensure that the maximum size of the image slices does not exceed 1500 pixels, and set the backtracking step size to 200 pixels to ensure that each target appears completely in at least one slice.
    
    \item Recalculate the coordinates for each target using the cropping parameters.
    
    \item Transform coordinate data into a standard format and populate a JSON file accordingly with the designed VQA template.
\end{enumerate}

\noindent
Fig.~\ref{sup_fig:high_data} presents data from high altitude scenarios, comprising five types of question, i.e., Counting, Classification, True or False, Visual Grounding, and Fine-grained Visual Grounding. 

\noindent
\textbf{Easy visual tasks}:
\begin{enumerate}
    \item \textbf{Counting}: Determine the number of airplanes in the input image (Maximum$\ge30$).
    
    \item \textbf{Classification}: Determine the aircraft models presented in the image. A set of aircraft models is provided in the question (\textcolor{orange}{orange text}) as prior knowledge. MLLM is required to comprehend aircraft models associated with the image from the training data and to indicate the specific models shown in the image.
    
    \item \textbf{True or False}: Determine whether there is a specific aircraft model in the image. MLLM needs to first extract the specific aircraft model from the query, then assess whether that aircraft model is present in the input image, and finally output a standard response of 'yes' or 'no'.
\end{enumerate}

\noindent
\textbf{Complex visual tasks}:
\begin{enumerate}
    \item \textbf{Visual Grounding}: Locate all the airplanes in the input image and then output the rectangular coordinates in LIST format. This requires MLLMs to accurately interpret the query and locate the target precisely within the image. Precise localization capability is a critical step for MLLMs in practical applications.
    
    \item \textbf{Visual Grounding (fine-grained)}: Locate the specified airplane model in the image and output the bounding box coordinates in LIST format. In contrast to task 4, the fine-grained detection task requires MLLMs to accurately identify the target airplane model and exhibits fine-grained classification capabilities. The small size of the targets makes fine-grained detection particularly challenging.
\end{enumerate}

\begin{figure*}[t]
    \centering
    \includegraphics[width=6.8in]{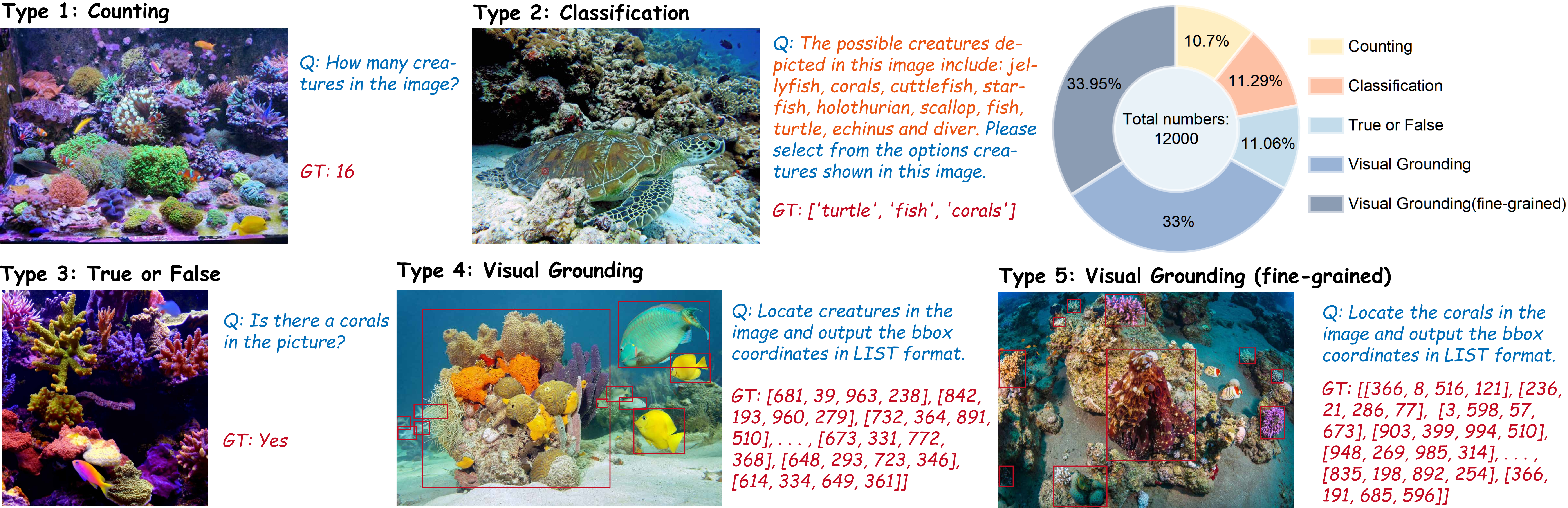}
    \caption{Illustrations of underwater data.} \label{sup_fig:underwater}
\end{figure*}

\subsection{Underwater}

\noindent
The source data of underwater scenario are from~\cite{RUOD23}.

\noindent
\textbf{Data Preprocess}: We mainly execute three operations.
\begin{enumerate}  
    \item Perform adaptive scaling on the image to ensure that its dimensions are less than 1500$\times$1500 pixels.
    
    \item Adjust the coordinates of all targets according to the scaling parameters.
    
    \item Transform coordinate data into a standard format and populate a JSON file accordingly with the designed VQA template.
\end{enumerate}

\noindent
Fig.~\ref{sup_fig:underwater} presents data from underwater scenarios, comprising five types of question, i.e., Counting, Classification, True or False, Visual Grounding, and Fine-grained Visual Grounding. 

\noindent
\textbf{Easy visual tasks}:
\begin{enumerate}
    \item \textbf{Counting}: Determine the number of creatures in the input image (Maximum$\ge80$).
    
    \item \textbf{Classification}: Determine the species presented in the image. A set of species is provided in the question (\textcolor{orange}{orange text}) as prior knowledge. MLLM is required to comprehend the creatures associated with the image from the training data and to indicate the specific species shown in the image.
    
    \item \textbf{True or False}: Determine whether there is a specific creature in the image. MLLM needs to first extract the specific creatures from the query, then assess whether that creature is present in the input image, and finally output a standard response of 'yes' or 'no'.
\end{enumerate}

\noindent
\textbf{Complex visual tasks}:
\begin{enumerate}
    \item \textbf{Visual Grounding}: Locate all the creatures in the input image and then output the rectangular coordinates in LIST format. Due to the light attenuation and scattering, as well as the camouflage colors of marine creatures, object detection in underwater scenario is highly challenging.
    
    \item \textbf{Visual Grounding (fine-grained)}: Locate the specified creature in the image and output the bounding box coordinates in LIST format. The diminutive size of certain marine organisms (such as echinus and fish) complicates the detection of such small targets against a complex background.
\end{enumerate}

\begin{figure*}[t]
    \centering
    \includegraphics[width=6.8in]{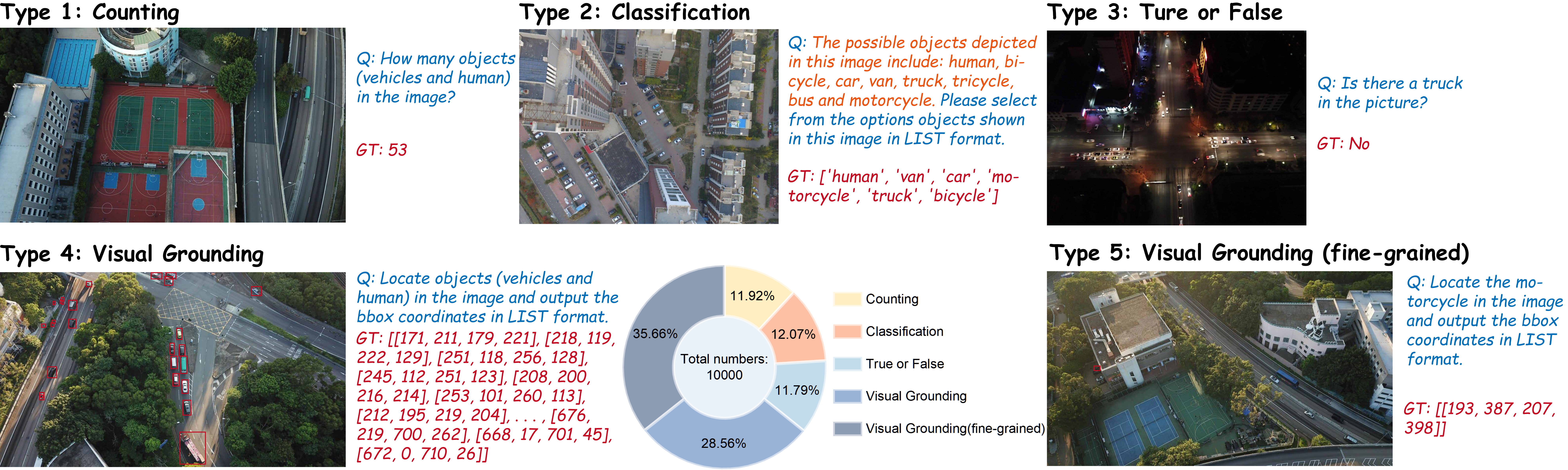}
    \caption{Illustrations of low altitude data.} \label{sup_fig:low}
\end{figure*}

\subsection{Low altitude}
\noindent
The source data of low altitude scenario are from~\cite{zhu2021detection}.

\noindent
\textbf{Data Preprocess}: We mainly execute three operations.
\begin{enumerate}  
    \item Perform adaptive scaling on the image to ensure that its dimensions are less than 1500$\times$1500 pixels.
    
    \item Adjust the coordinates of all targets according to the scaling parameters.
    
    \item Transform coordinate data into a standard format and populate a JSON file accordingly with the designed VQA template.
\end{enumerate}

\noindent
Fig.~\ref{sup_fig:low} presents data from low altitude scenarios, comprising five types of question, i.e., Counting, Classification, True or False, Visual Grounding, and Fine-grained Visual Grounding. 

\noindent
\textbf{Easy visual tasks}:
\begin{enumerate}
    \item \textbf{Counting}: Determine the number of vehicles and human in the input image (Maximum$\ge80$).
    
    \item \textbf{Classification}: Determine the objects presented in the image. A set of objects is provided in the question (\textcolor{orange}{orange text}) as prior knowledge. MLLM is required to comprehend the objects associated with the image from the training data and to indicate the specific vehicles (or human) shown in the image.
    
    \item \textbf{True or False}: Determine whether there is a specific vehicle in the image. MLLM needs to first extract the specific vehicles from the query, then assess whether that vehicle is present in the input image, and finally output a standard response of 'yes' or 'no'.
\end{enumerate}

\noindent
\textbf{Complex visual tasks}:
\begin{enumerate}
    \item \textbf{Visual Grounding}: Locate all the vehicles and human in the input image and then output the rectangular coordinates in LIST format. Target localization in low-altitude scenarios is challenging due to the small size of the targets and frequent obstructions.
    
    \item \textbf{Visual Grounding (fine-grained)}: Locate the specified vehicle in the image and output the bounding box coordinates in LIST format. Fine-grained targets are typically concealed within complex backgrounds and are frequently subject to severe overlap and occlusion, thereby complicating fine-grained localization.
\end{enumerate}

\begin{figure}[t]
    \centering
    \includegraphics[width=3.25in]{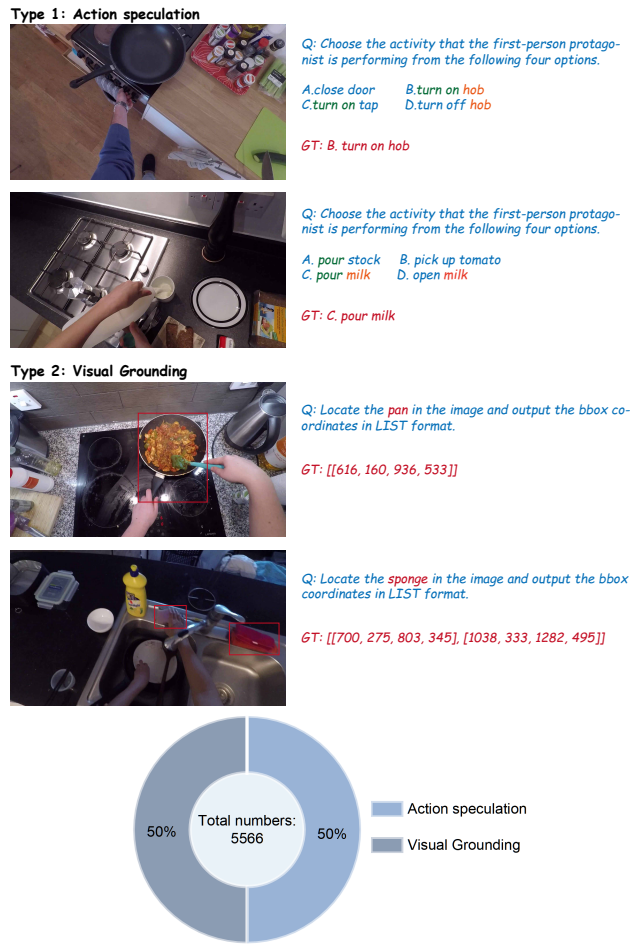}
    \caption{Illustrations of indoor data.} \label{sup_fig:indoor}
\end{figure}

\subsection{Indoor}

\noindent
The source data for the indoor scenario are from a large-scale first-person video dataset~\cite{Damen2018EPICKITCHENS}.

\noindent
\textbf{Data Preprocess}: We mainly execute four operations.
\begin{enumerate}  
    \item Extract key frames from the video stream and remove duplicate frames.
    
    \item Perform adaptive scaling on the image to ensure that its dimensions are less than 1500$\times$1500 pixels.
    
    \item Adjust the coordinates of all targets according to the scaling parameters.
    
    \item Transform coordinate data into a standard format and populate a JSON file accordingly with the designed VQA template.
\end{enumerate}

\noindent
Fig.~\ref{sup_fig:indoor} presents data from indoor scenarios, comprising two types of question, i.e., Action speculation and Visual Grounding. 

\noindent
\textbf{Action speculation}: is conducted through choice questions aimed at deducing the ongoing actions of the first-person protagonist. Each option is formed by pairing a verb (or verbal phrase) with a noun. The set of four options includes: one random action, one correct action, one option that uses the \textcolor{green}{correct verb} but an \textcolor{red}{incorrect noun}, and one option that uses the \textcolor{orange}{correct noun} but an \textcolor{red}{incorrect verb}. The inclusion of these distractors ensures that the model must accurately interpret the character's action in the image, rather than relying solely on object recognition to choose the correct answer.

\noindent
\textbf{Visual Grounding}: locates the common objects in the indoor scenario from the first-person perspective. Due to the constrained field of view inherent in the first-person perspective and the presence of optical distortions such as lens aberration and defocus, object detection within dynamic indoor scenario data streams is a significant challenge.

The data distribution in~Fig.~\ref{sup_fig:high_data} to~\ref{sup_fig:indoor} indicates that MSVQA involves a significant proportion of complex visual grounding tasks, which imposes greater demands on the model's visual reasoning abilities.

\begin{figure*}[htbp]
    \centering
    \begin{subfigure}[b]{0.245\textwidth}
        \centering
        \includegraphics[width=\textwidth]{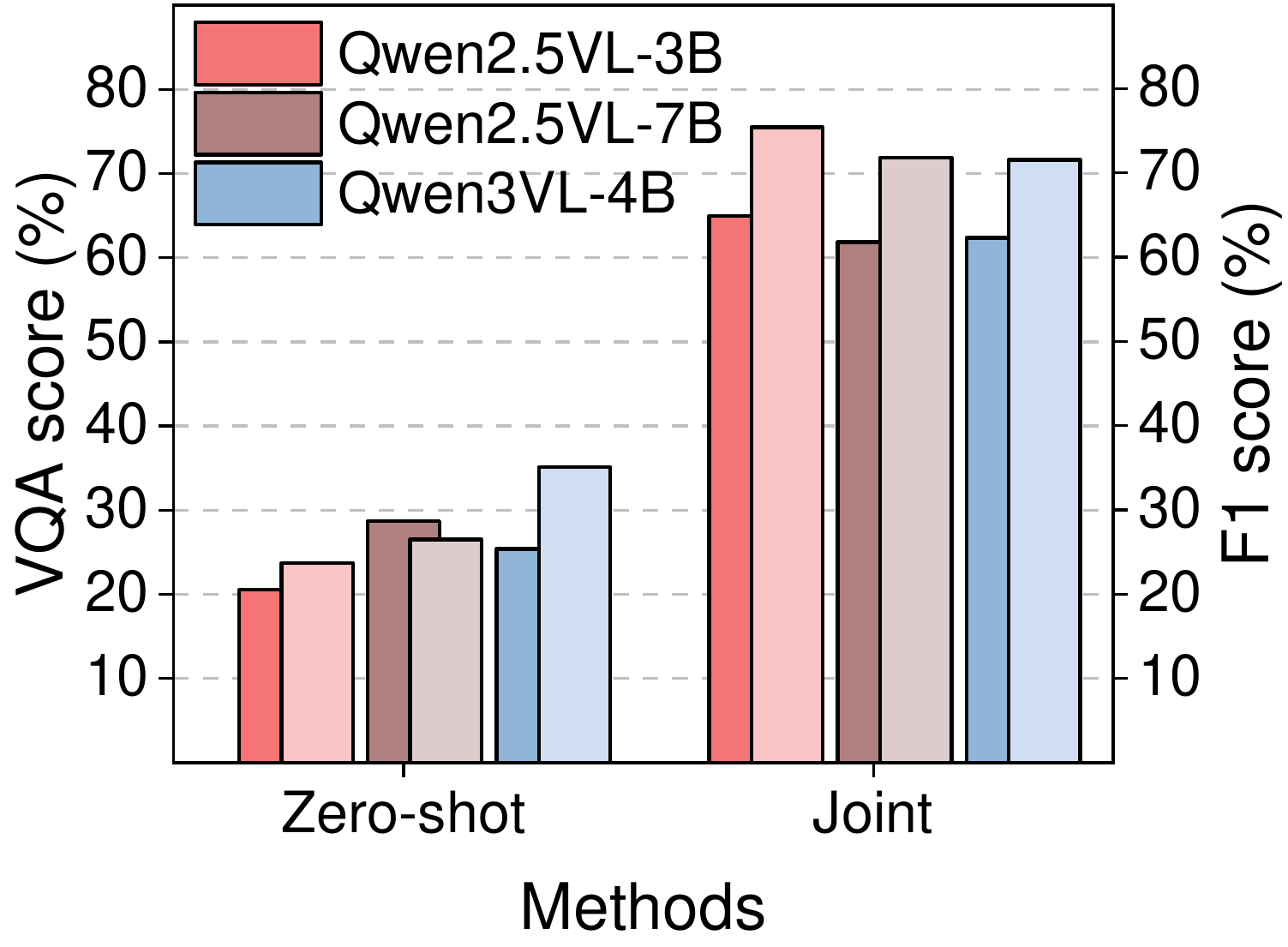}
        \caption{High altitude}
        \label{sub_fig:qwenjoint_high}
    \end{subfigure}
    \begin{subfigure}[b]{0.245\textwidth}
        \centering
        \includegraphics[width=\textwidth]{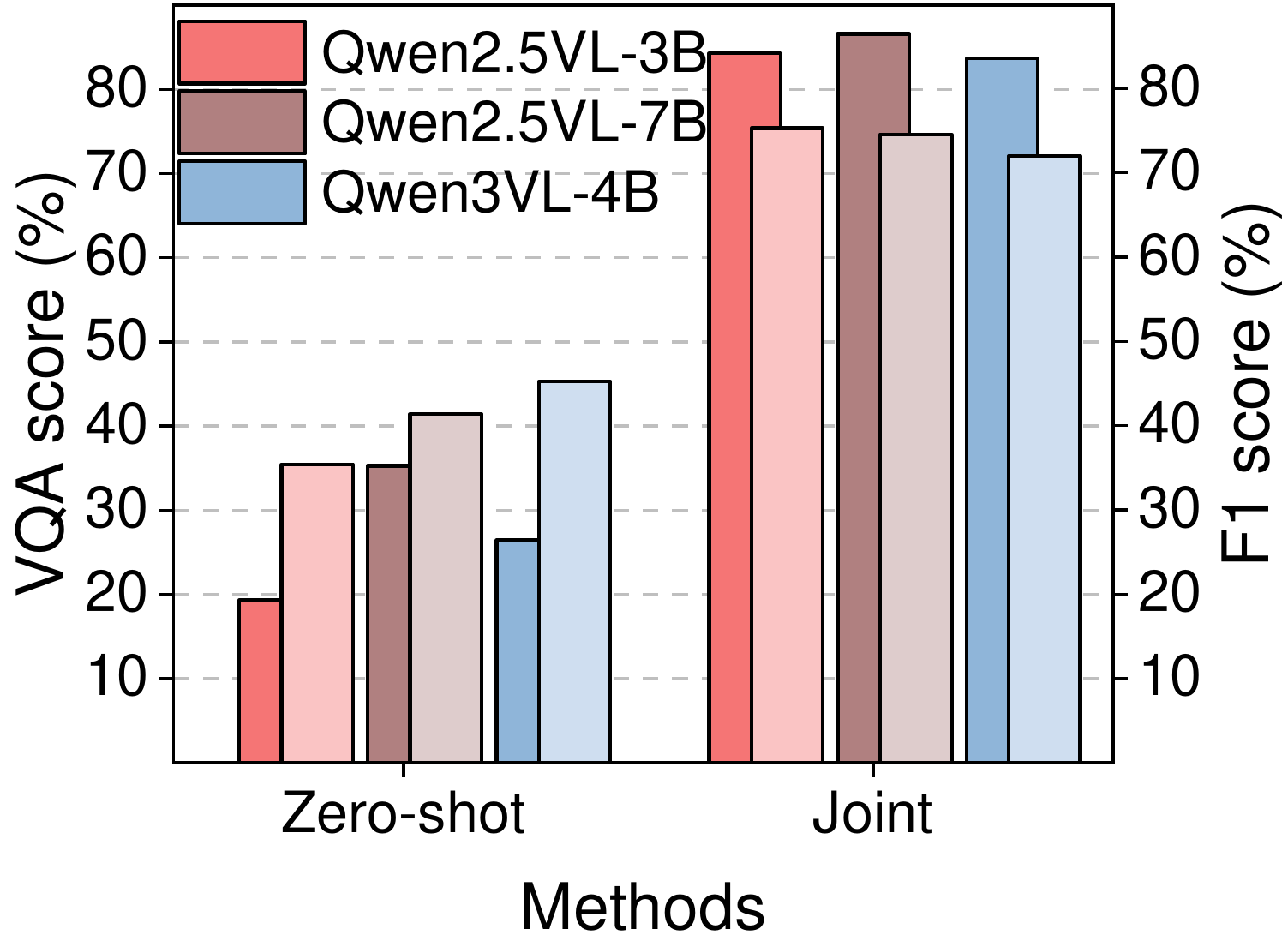}
        \caption{Underwater}
        \label{sub_fig:qwenjoint_underwater}
    \end{subfigure}
    \begin{subfigure}[b]{0.245\textwidth}
        \centering
        \includegraphics[width=\textwidth]{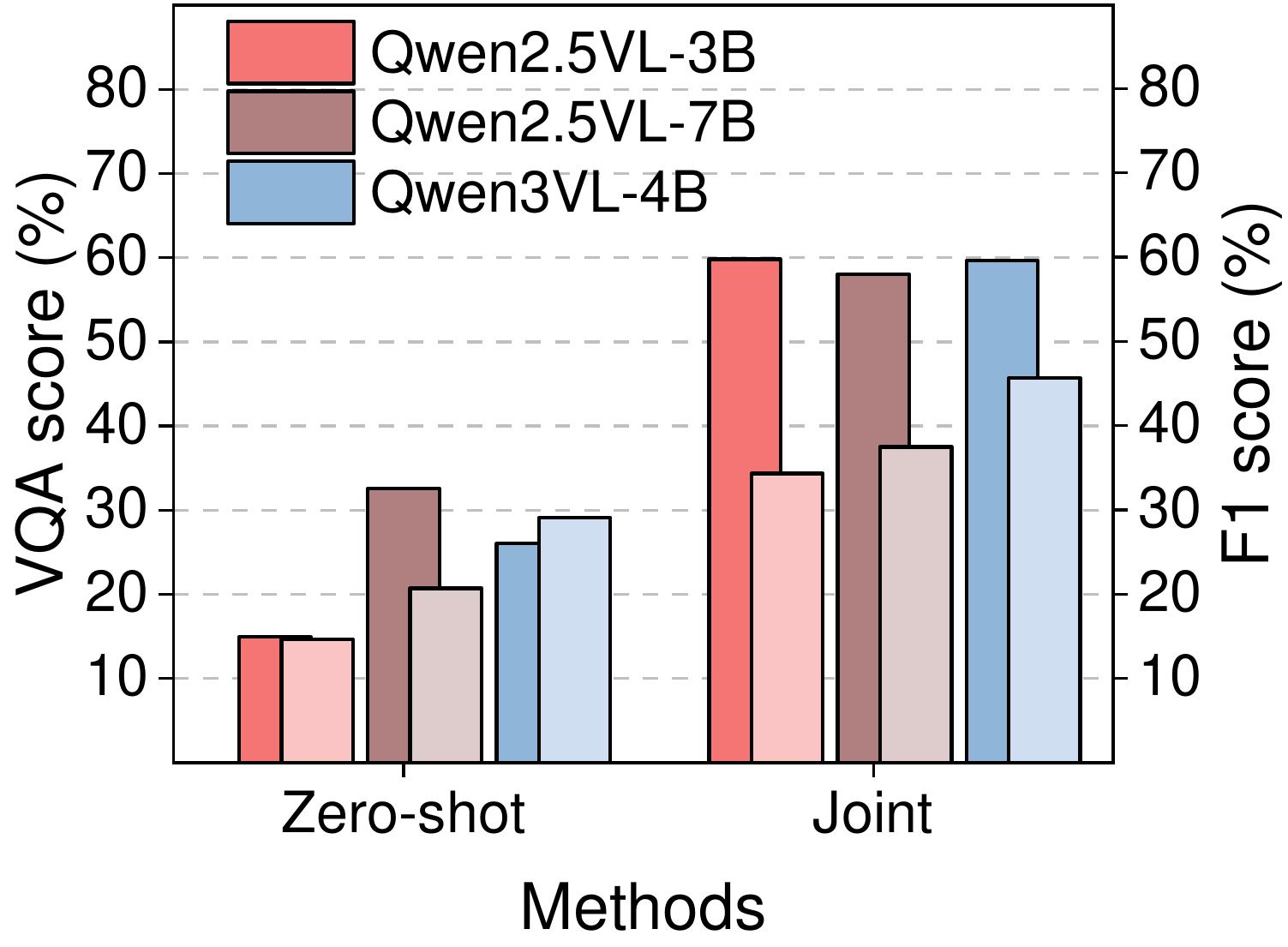}
        \caption{Low altitude}
        \label{sub_fig:qwenjoint_low}
    \end{subfigure}
    \begin{subfigure}[b]{0.245\textwidth}
        \centering
        \includegraphics[width=\textwidth]{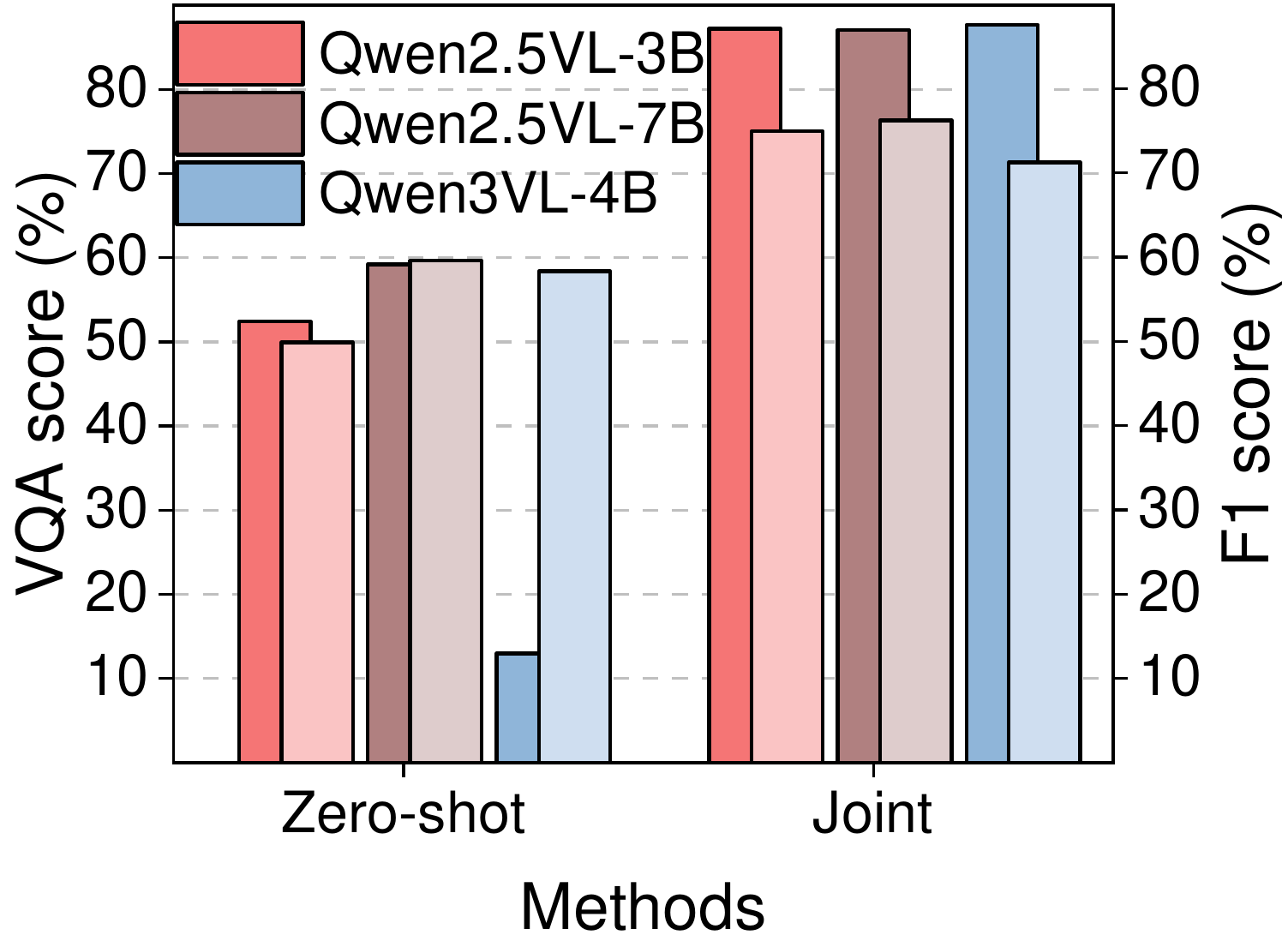}
        \caption{Indoor}
        \label{sub_fig:qwenjoint_indoor}
    \end{subfigure}
    \caption{Comparison of Zero-shot and the upper bound between Qwen2.5VL-3B, Qwen2.5VL-7B and Qwen3VL-4B. The dark color corresponds to the VQA score, and the light color corresponds to the F1 score.}
    \label{fig:qwen_joint_cmp}
\end{figure*}

\section{Evaluation metrics} \label{sup_sec:metric}

\subsection{VQA score}
We compute the VQA score for each scenario by aggregating the scores from the Counting, Classification, True or False, and Action speculation tasks.

\noindent
\textbf{Counting.}\quad Due to the presence of small, numerous, and overlapping targets in certain scenarios, we allow the model to have a certain degree of prediction error. Consequently, 1 score is assigned if the predicted count exactly matches the ground truth, 0.5 score is given if the absolute difference between the predicted count and the ground truth is less than one, and no score is awarded otherwise.

\noindent
\textbf{Classification.}\quad We first compare the predicted results with the ground truth. Since the ground truth comprises multiple category labels, we can obtain the number of correct predictions $n_{\rm correct}$ and incorrect predictions $n_{\rm wrong}$. Assume $n_{\rm all}$ denotes the sequence length of ground truth, the final score can be calculated by
\begin{align}
    \max \left( {\frac{{\left( {{n_{\rm correct}} - 0.5 \times {n_{\rm wrong}}} \right)}}{{{n_{\rm all}}}},0} \right).
\end{align}

\noindent
\textbf{True or False.}\quad The score is determined by directly comparing the predictions with the ground truth, where a correct match yields 1 score.

\noindent
\textbf{Action speculation.}\quad The score is determined by directly comparing the predictions with the ground truth, where a correct match yields 1 score.

For high altitude, underwater, and low altitude scenarios, we compute the total scores for the Counting, Classification, and True or False tasks individually, then normalize them to a percentage scale to derive the VQA score. For the indoor scenario, we calculate the total score for action speculation and similarly convert it to a percentage score as the VQA score.

\subsection{F1 score}
We adopt the F{\footnotesize 1} score to measure the visual grounding capacity of MLLMs. First, we collect the predicted bounding boxes for all images. Subsequently, these predictions are matched with ground truth boxes based on ${\rm IoU=0.5}$, allowing the calculation of true positives (TP), false positives (FP), and false negatives (FN). The precision is calculated by ${\rm TP}/({\rm TP} + {\rm FP})$ and the recall is calculated by ${\rm TP}/({\rm TP} + {\rm FN})$. The F1 score is defined as the harmonic mean of precision and recall,
\begin{align}
    {\rm F1} = \frac{{2 \times {\rm Precision} \times {\rm Recall}}}{{\left( {{\rm Precision} + {\rm Recall}} \right)}}.
\end{align}

In particular, the F{\footnotesize 1} score for visual grounding and fine-grained visual grounding tasks should be calculated separately for high altitude, underwater and low altitude scenarios. Therefore, the average of the F{\footnotesize 1} scores from two tasks is adopted to evaluate the performance of complex visual tasks in each scenario. 

\begin{figure*}[htbp]
    \centering
    \begin{subfigure}[b]{0.495\textwidth}
        \centering
        \includegraphics[width=\textwidth]{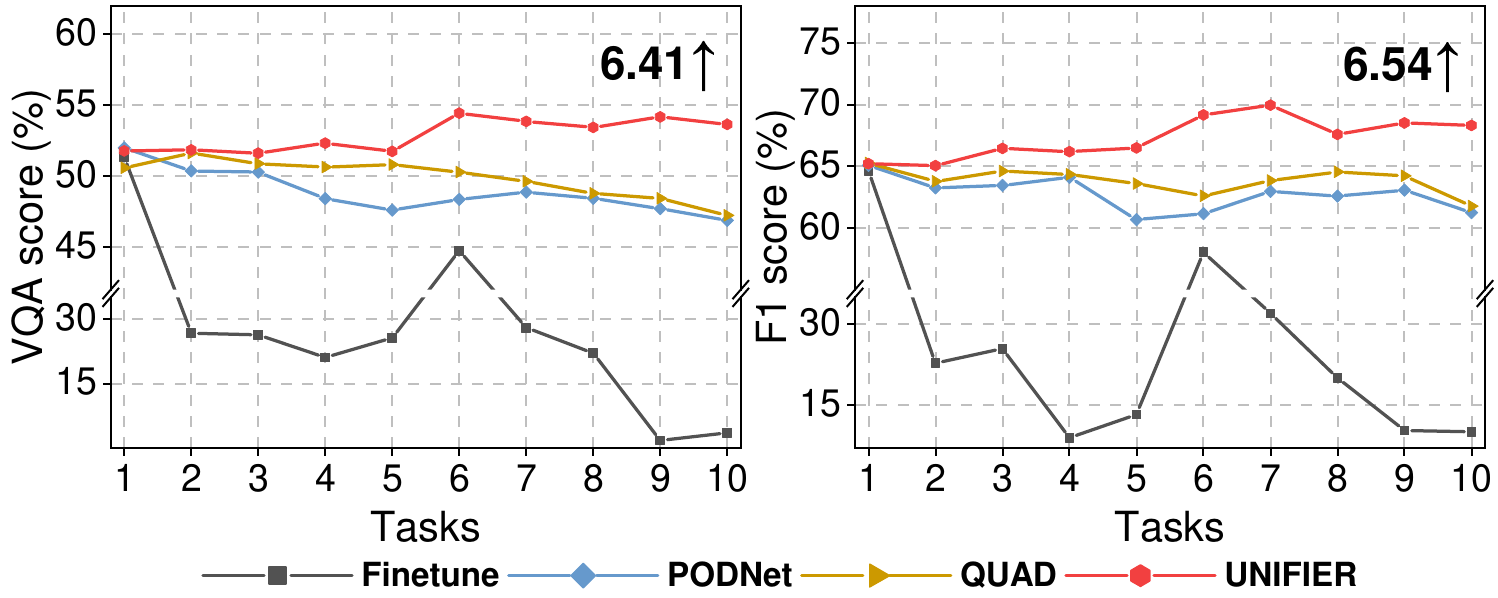}
        \caption{High altitude}
        \label{sub_fig:qwen3_High_10steps}
    \end{subfigure}
    \begin{subfigure}[b]{0.495\textwidth}
        \centering
        \includegraphics[width=\textwidth]{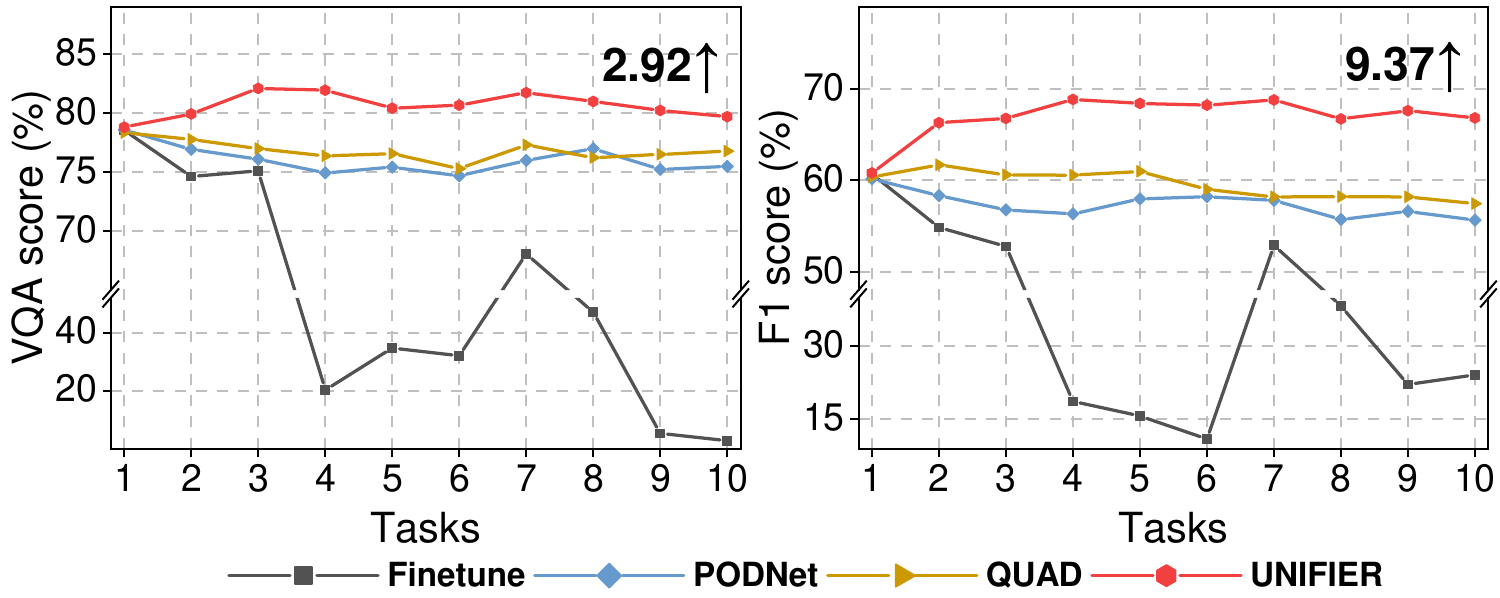}
        \caption{Underwater}
        \label{sub_fig:qwen3_Underwater_10steps}
    \end{subfigure}
    \\
    \begin{subfigure}[b]{0.495\textwidth}
        \centering
        \includegraphics[width=\textwidth]{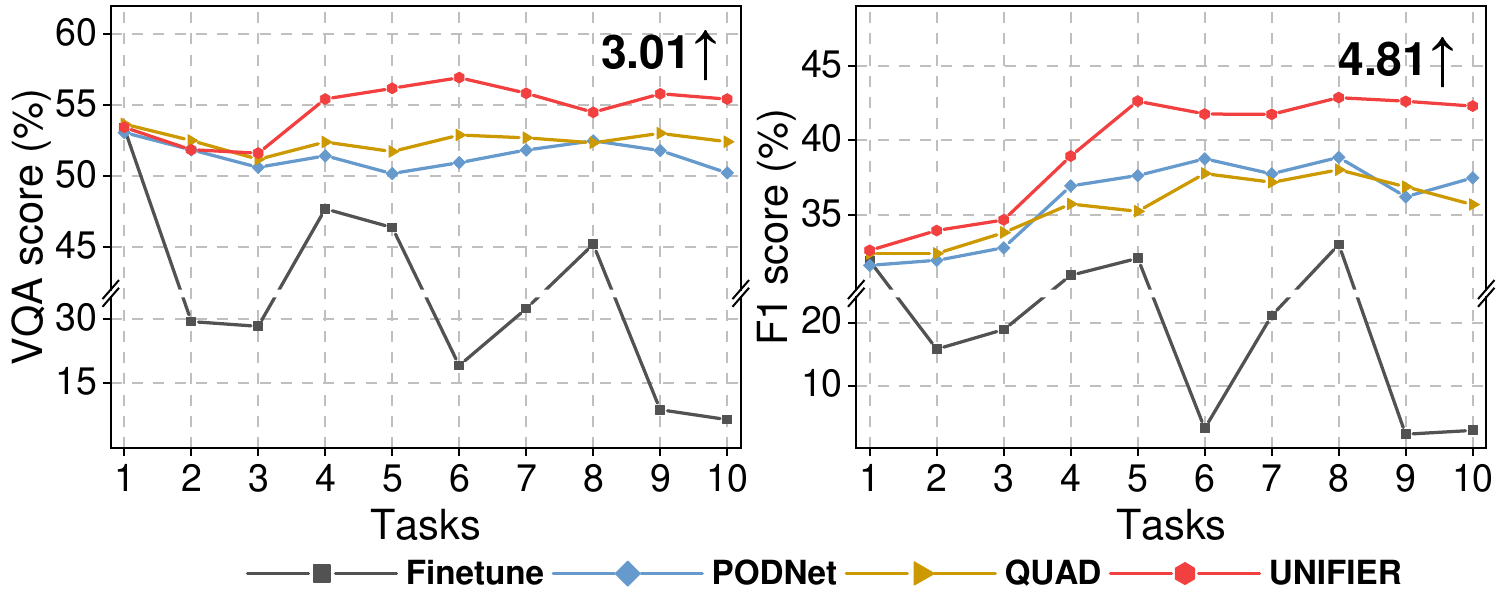}
        \caption{Low altitude}
        \label{sub_fig:qwen3_Low_10steps}
    \end{subfigure}
    \begin{subfigure}[b]{0.495\textwidth}
        \centering
        \includegraphics[width=\textwidth]{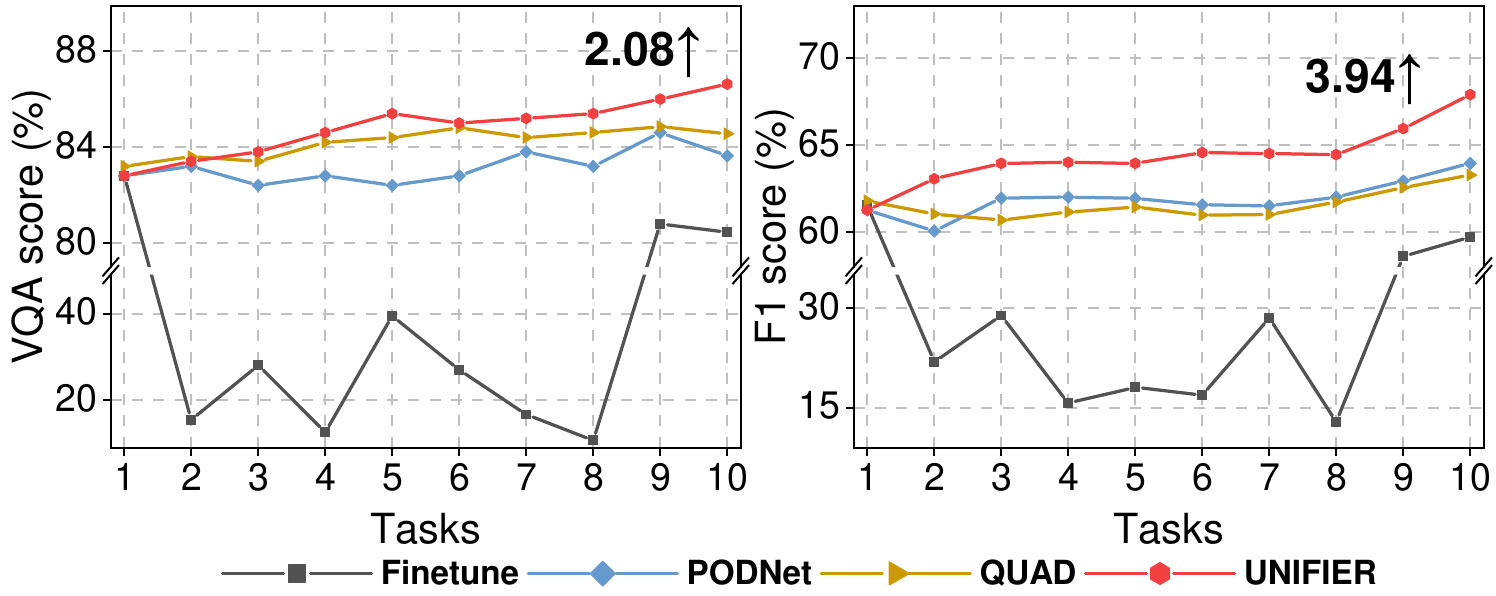}
        \caption{Indoor}
        \label{sub_fig:qwen3_Indoor_10steps}
    \end{subfigure}
    \caption{Incremental trends on 10 steps setting using \textbf{Qwen3-VL}. The performance gap is annotated at the end of each curve.}
    \label{fig:qwen3_comparison_10steps}
\end{figure*}

\section{Comparison methods} \label{sup_sec:cmp}
We compared nine different methods, comprising zero-shot for pre-training capability, the lower bound (Finetune), the upper bound (Joint), two exemplar-free methods, and four exemplar-based methods. The descriptions of these methods are provided below.
\begin{itemize}
    \item \textbf{Zero-shot}: aims to validate the performance from pre-training.~It initializes the model with the pre-trained weight and then is validated on the test set. Due to the non-standard output by an untrained model, We analyzed the inference outputs and optimize the scoring algorithm by incorporating techniques such as fuzzy matching.
    \item \textbf{Joint}: is the upper bound of continual learning. It always learns both previous data and current data in each task jointly. However, it spends more training and storage cost.
    \item \textbf{Finetune}: is the lower bound of continual learning. It only learns current data without other strategies. Therefore, it suffers from severe catastrophic forgetting.
    \item \textbf{EWC}~\cite{ewc}: addresses catastrophic forgetting by selectively slowing the learning of important weights for previous tasks, using a regularization term based on the Fisher Information Matrix to constrain significant parameter changes. Its key advantage is effectively preserving performance on old tasks without storing past data, through a relatively simple and computationally efficient mechanism. However, EWC can be computationally expensive to compute the Fisher matrix for MLLMs.
    \item \textbf{Tailor}~\cite{tailor}: aims to mitigate catastrophic forgetting by identifying a sparse model patch of critical parameters through a fusion of salience and sensitivity analysis, followed by targeted compensation to enhance task adaptation. However, it exhibits limited performance in multi-scenario tasks due to substantial variations in visual features.
    \item \textbf{ER}~\cite{er}: is a rehearsal-based method that employs a fixed-size memory buffer to store and randomly sample visited examples for retraining.
    \item \textbf{PODNet}~\cite{podnet}: is also a rehearsal-based method that employs pooling operations across the dimensions of intermediate features to derive compressed representations for each dimension. It facilitates a trade-off between stability and plasticity by computing the L2 norm between the compressed representations of the old and new models. The original method is based on convolutional networks. We develop it to fit the ViT architecture.
    \item \textbf{VQACL}~\cite{vqacl}: employs a rehearsal-based strategy that integrates a prototype module to capture both task-specific and invariant features, thereby enabling robust and generalizable representations for VQA tasks.
    \item \textbf{QUAD}~\cite{quad}: employs a questions-only rehearsal strategy without storing visual data by leveraging previous task questions for regularization. It integrates question replay to prevent overfitting and attention consistency distillation to preserve cross-modal associations. However, it may underperform in tasks requiring detailed visual reasoning due to the absence of image storage.
\end{itemize}

\begin{figure*}[htbp]
    \centering
    \begin{subfigure}[b]{0.495\textwidth}
        \centering
        \includegraphics[width=\textwidth]{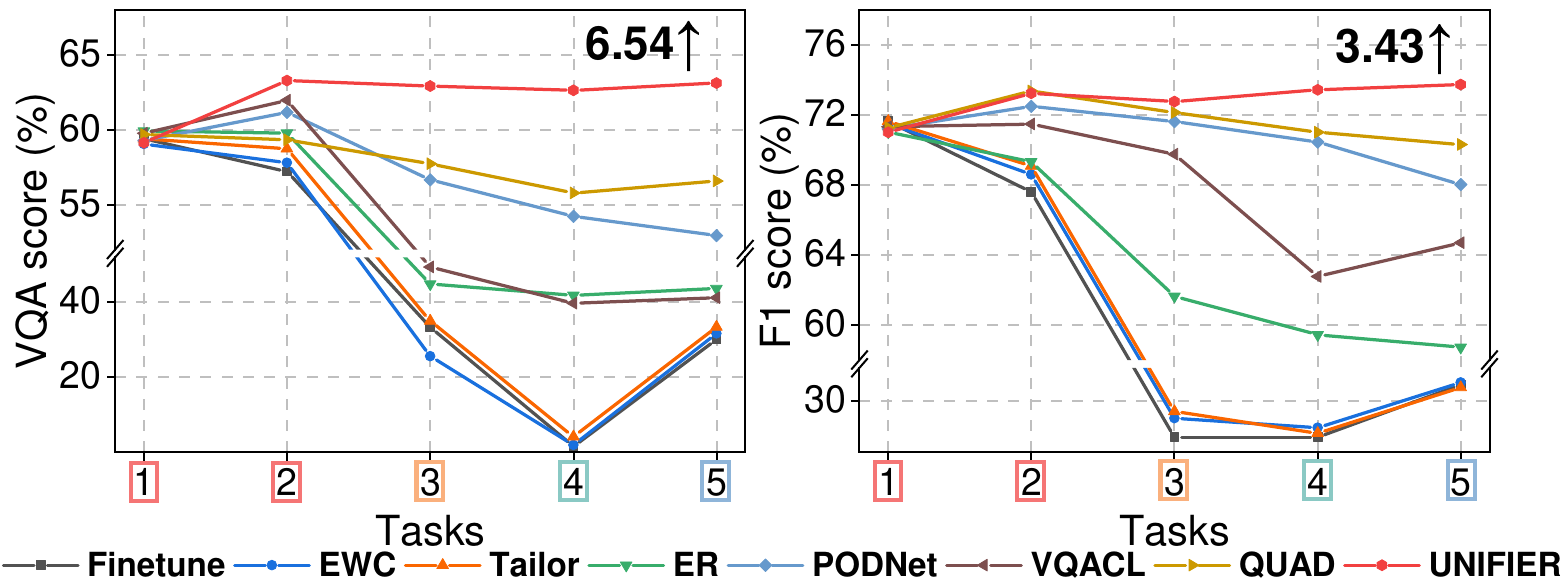}
        \caption{High altitude}
        \label{sub_fig:High_5steps}
    \end{subfigure}
    \begin{subfigure}[b]{0.495\textwidth}
        \centering
        \includegraphics[width=\textwidth]{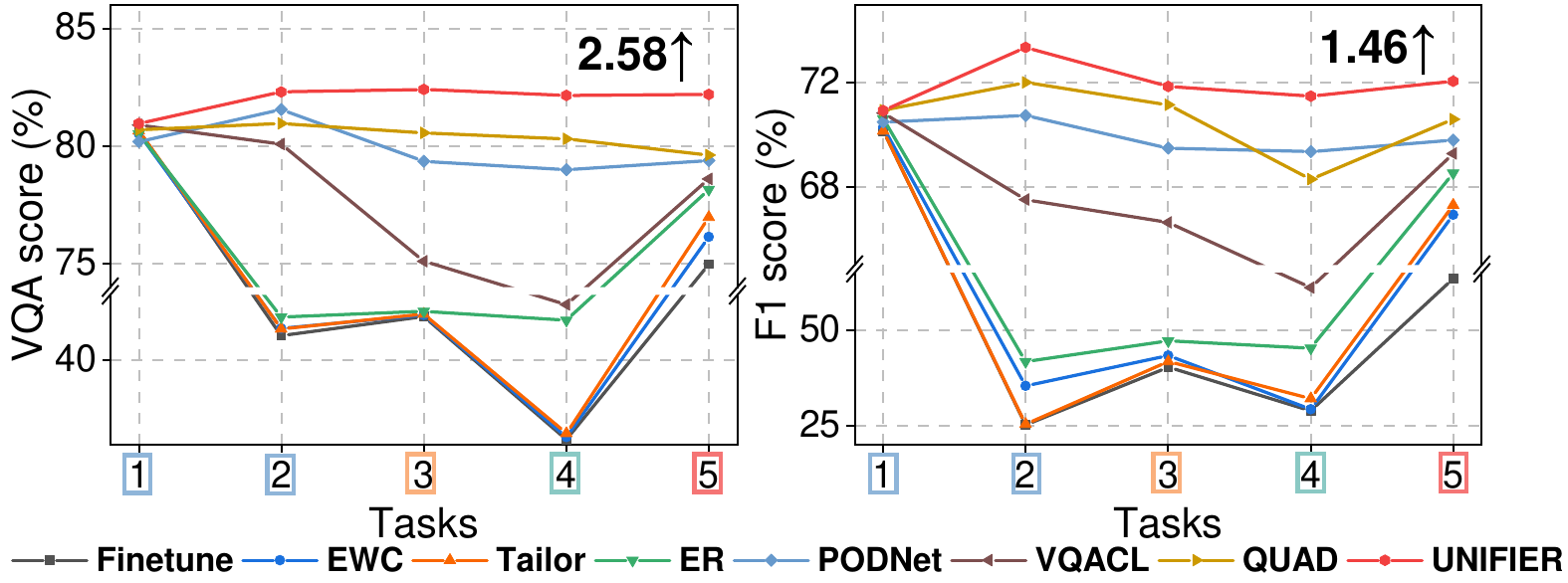}
        \caption{Underwater}
        \label{sub_fig:Underwater_5steps}
    \end{subfigure}
    \\
    \begin{subfigure}[b]{0.495\textwidth}
        \centering
        \includegraphics[width=\textwidth]{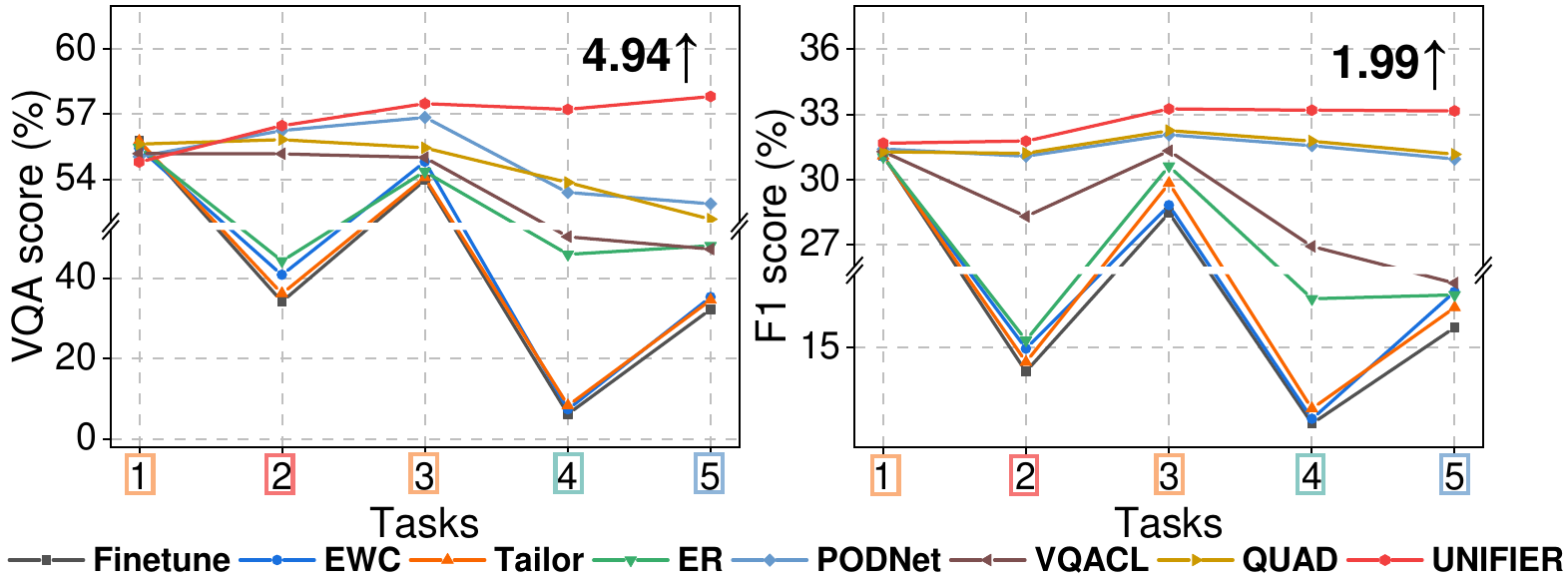}
        \caption{Low altitude}
        \label{sub_fig:Low_5steps}
    \end{subfigure}
    \begin{subfigure}[b]{0.495\textwidth}
        \centering
        \includegraphics[width=\textwidth]{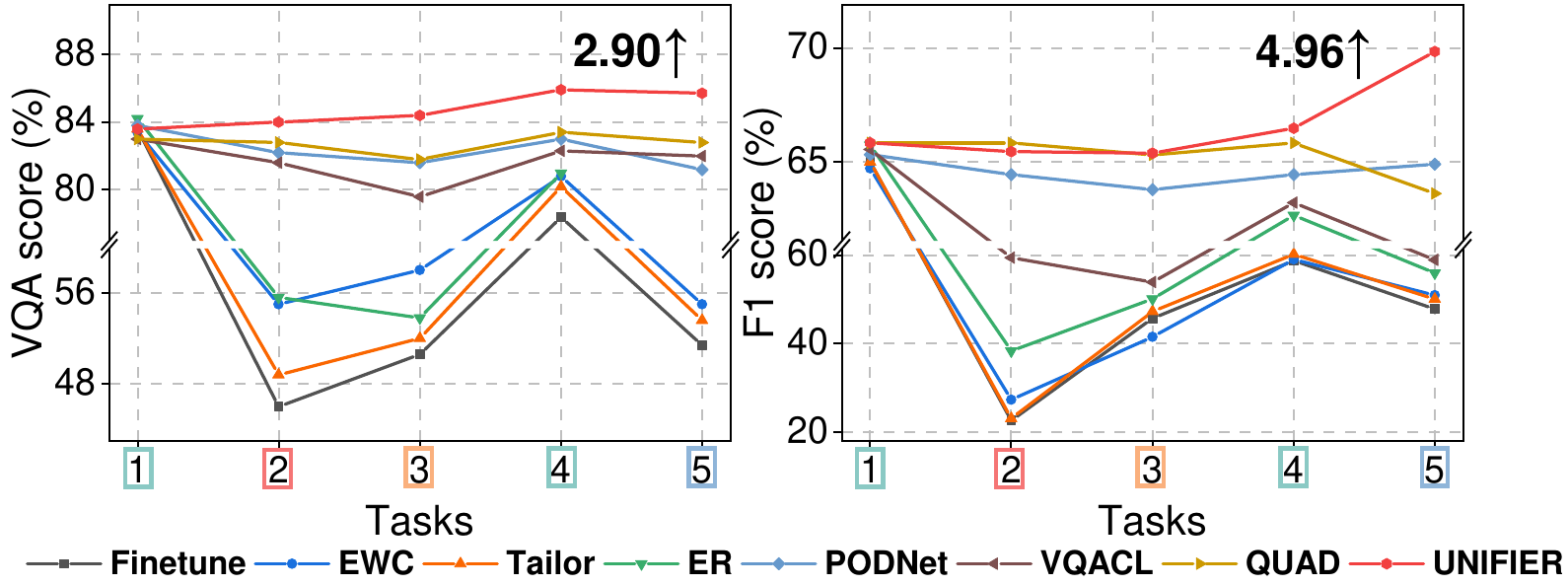}
        \caption{Indoor}
        \label{sub_fig:Indoor_5steps}
    \end{subfigure}
    \caption{Incremental trends on 5 steps setting in different scenarios. The performance gap is annotated at the end of each curve. The color of the boxes on the horizontal axis indicates the task scenario: \textcolor{red!50}{High altitude}, \textcolor{blue!50}{Underwater}, \textcolor{orange!50}{Low altitude} and \textcolor{SeaGreen!50}{Indoor}.}
    \label{fig:comparison_5steps}
\end{figure*}

\begin{figure*}[htbp]
    \centering
    \begin{subfigure}[b]{0.495\textwidth}
        \centering
        \includegraphics[width=\textwidth]{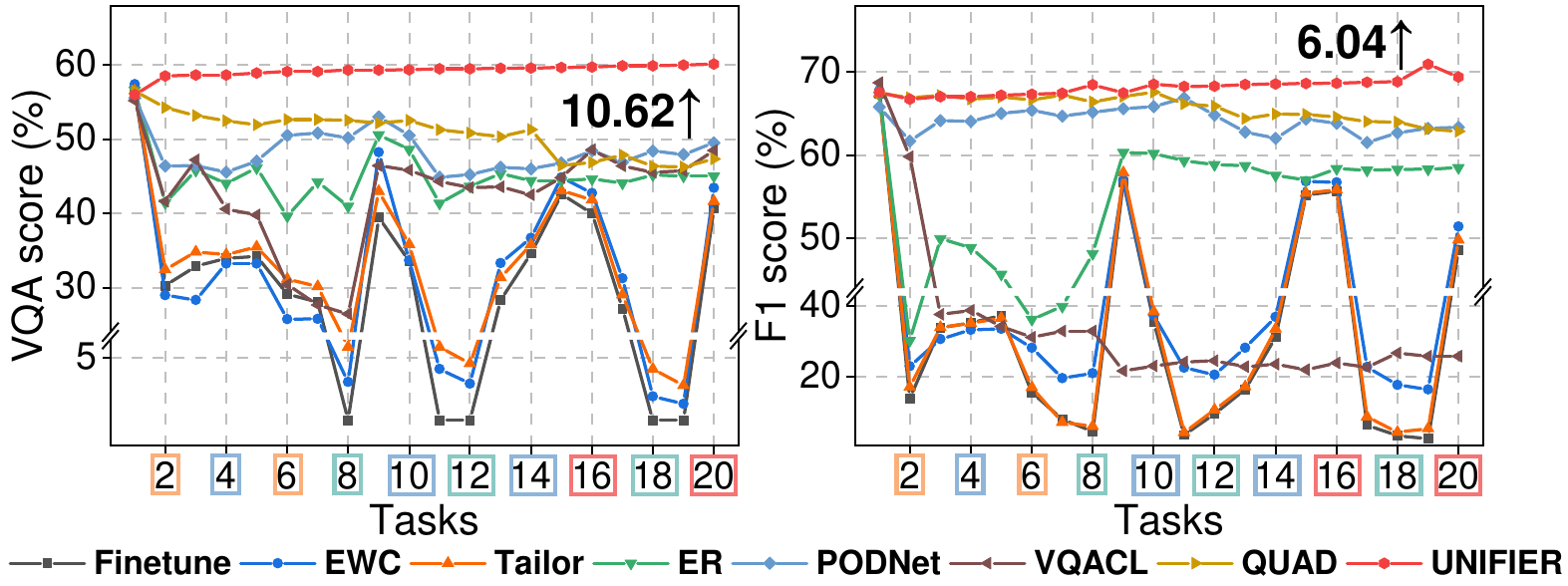}
        \caption{High altitude}
        \label{sub_fig:High_20steps}
    \end{subfigure}
    \begin{subfigure}[b]{0.495\textwidth}
        \centering
        \includegraphics[width=\textwidth]{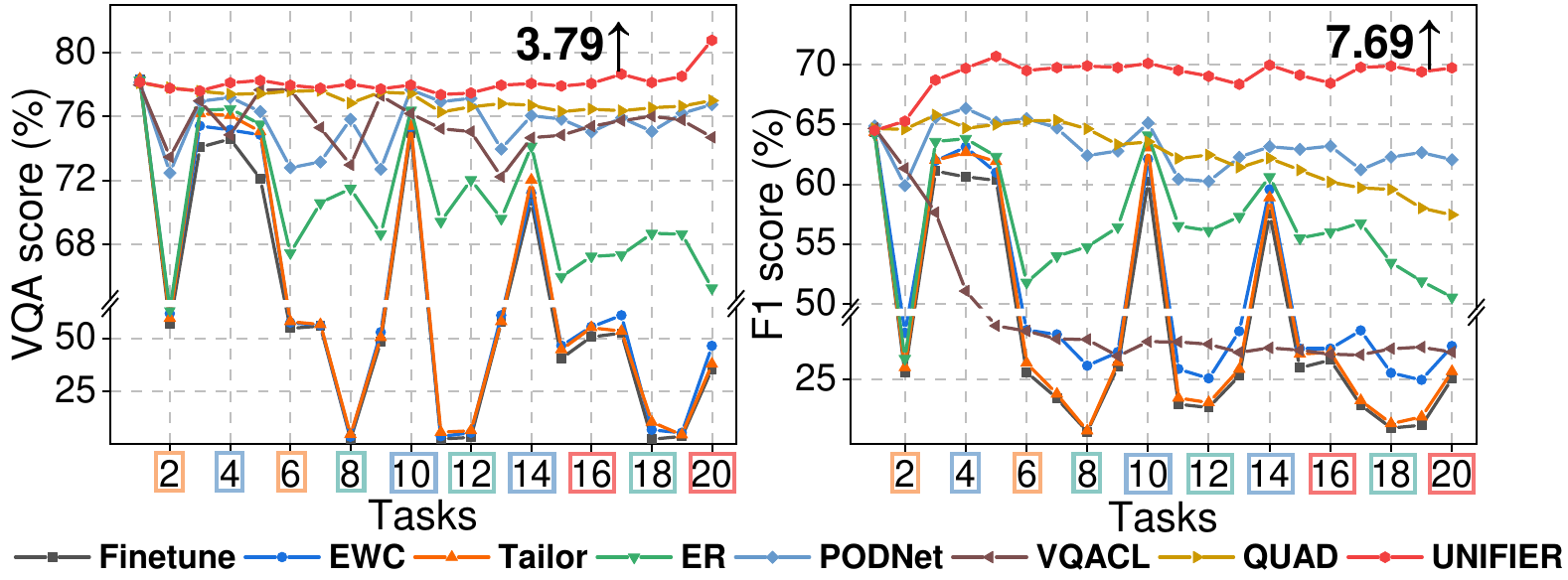}
        \caption{Underwater}
        \label{sub_fig:Underwater_20steps}
    \end{subfigure}
    \\
    \begin{subfigure}[b]{0.495\textwidth}
        \centering
        \includegraphics[width=\textwidth]{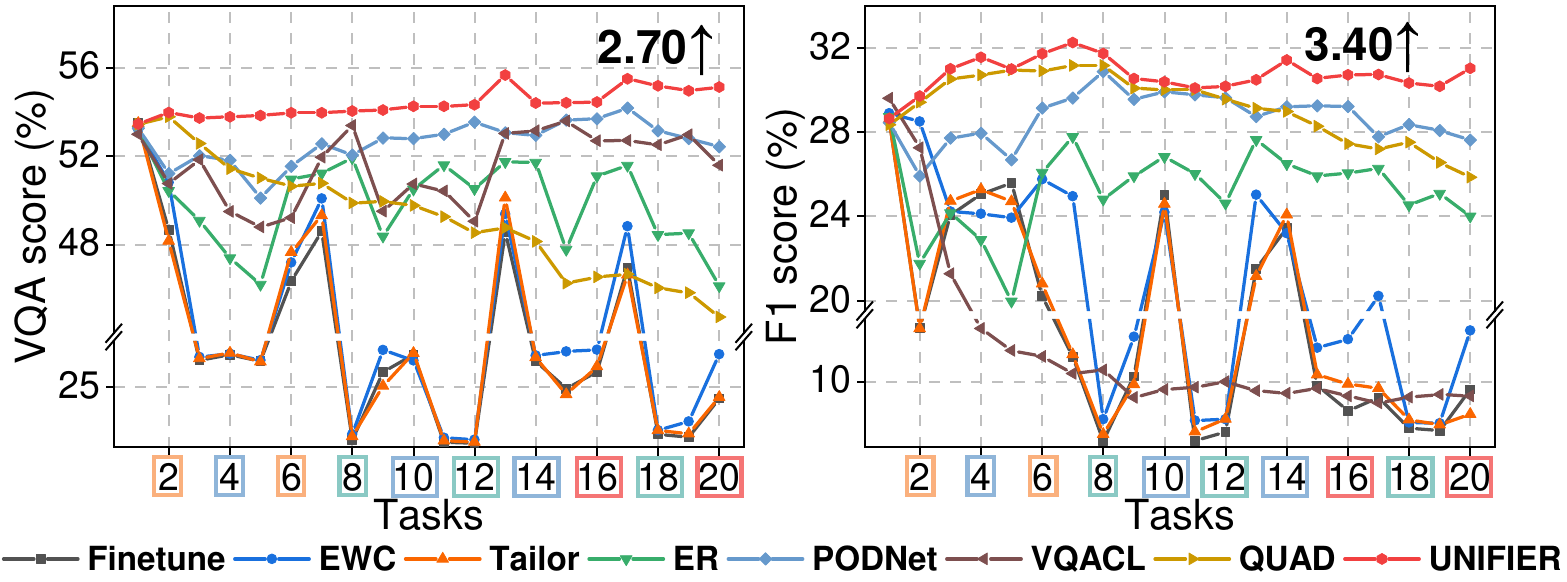}
        \caption{Low altitude}
        \label{fig:Low_20steps}
    \end{subfigure}
    \begin{subfigure}[b]{0.495\textwidth}
        \centering
        \includegraphics[width=\textwidth]{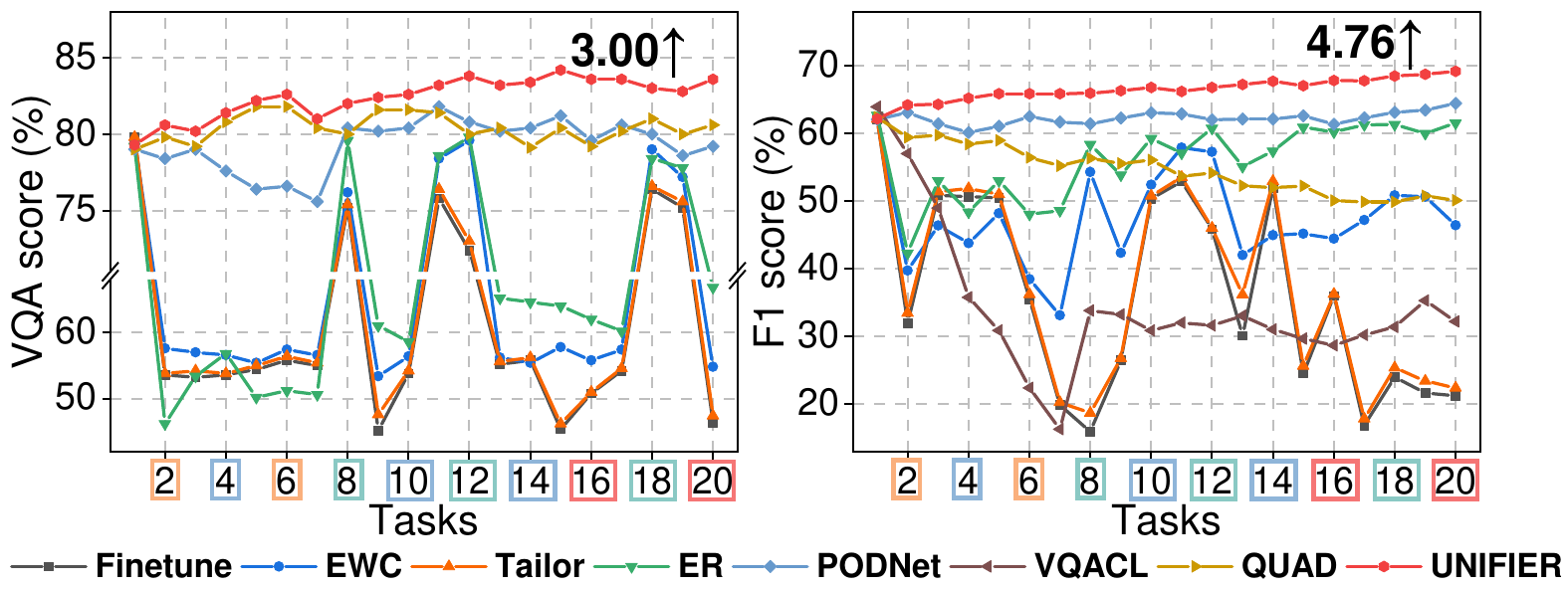}
        \caption{Indoor}
        \label{sub_fig:Indoor_20steps}
    \end{subfigure}
    \caption{Incremental trends on 20 steps setting in different scenarios. The performance gap is annotated at the end of each curve. The color of the boxes on the horizontal axis indicates the task scenario: \textcolor{red!50}{High altitude}, \textcolor{blue!50}{Underwater}, \textcolor{orange!50}{Low altitude} and \textcolor{SeaGreen!50}{Indoor}.}
    \label{fig:comparison_20steps}
\end{figure*}

\section{Hardware information} \label{sup_sec:hardware}
The hardware information is as follows:
\begin{itemize}
    \item \textbf{CPU}: Intel(R) XEON(R) PLATINUM 8558~(96 core)
    \item \textbf{GPU}: 8$\times$NVIDIA H200~(140 GB per device)
    \item \textbf{Mem}: 500 GB
\end{itemize}

\section{Complexity analysis} \label{sup_sec:complexity}
In Tab.~\ref{tab:complex_batch}, we provide a detailed comparison of the inference efficiency of various baseline models before and after incorporating the VRE module, demonstrating that the proposed VRE module does not impose a burden on model inference. We report four metrics to reflect the model’s inference efficiency:
\begin{itemize}
    \item \textbf{Params}: are obtained using the deepspeed Profiler tool, counting all model parameters on a single GPU.
    \item \textbf{TTFT (Time To First Token)}: The average time recorded from input image and text to the model outputting the first token; only this process requires the visual encoder to participate. Timing starts from input image to processor, excluding model loading time, and discarding the first 20 rounds of inference results to eliminate the impact of GPU warm-up.
    \item \textbf{TPOT (Time Per Output Token)}: The average time recorded for the model to generate each subsequent token autoregressively after outputting the first token; this process only requires the text decoder to participate and does not use the visual encoder. We also discard the first 20 rounds of inference results to eliminate the impact of GPU warm-up.
    \item \textbf{Mem.}: records the peak GPU memory usage on a single GPU during inference.
\end{itemize}

This includes the following three types of experiments: 
\begin{itemize}
    \item \textbf{BS=1, 1$\times$H200}: In this mode, we only use one GPU for inference, and only infer one sample at a time. The experimental results show that after adding the VRE module, TTFT increased by 9.82 ms, which is caused by the VRE module. However, TPOT remains unchanged. This indicates that only the generation time of the first token slightly increases, and subsequent token generation is not affected. For generative large language models, the number of generated tokens is usually greater than 100, and can even reach tens of thousands. \textbf{Therefore, the inference cost brought by adding the VRE module is negligible}. \textbf{Note}: Since 1-GPU inference with BS=1 sample yields low utilization and cannot reach peak GPU efficiency, the inference speeds of the three models are similar. We conducted supplementary experiments subsequently.
    \item \textbf{BS=96, 1$\times$H200}: In this mode, we still use 1 GPU for inference, but set the batch size to 96, thereby increasing the GPU efficiency to the peak. Compared to the experimental results with BS=1, the time for the model to generate a single token significantly decreases. The experimental phenomena are consistent with the BS=1 experiment, with the model's TTFT slightly increasing (<20 ms) and TTOP remaining unchanged. The experiment once again proves that \textbf{the proposed method does not introduce additional inference burden}.
    \item \textbf{BS=96, 4$\times$H200}: Since model inference typically utilizes multiple GPUs, we further employed four GPUs with a batch size of 96 for inference. The experimental results remain consistent with the two sets of experiments mentioned above, once again demonstrating that \textbf{the proposed method does not introduce inference overhead}. Note: It is normal for token generation time in multi-GPU experiments to be longer compared to single-GPU token generation, as multi-GPU setups must consider thread safety and secure communication between the CPU and multiple GPUs during inference, thereby reducing generation efficiency.
\end{itemize}

\begin{figure*}[htbp]
    \centering
    \begin{subfigure}[b]{0.25\textwidth}
        \centering
        \includegraphics[width=\textwidth]{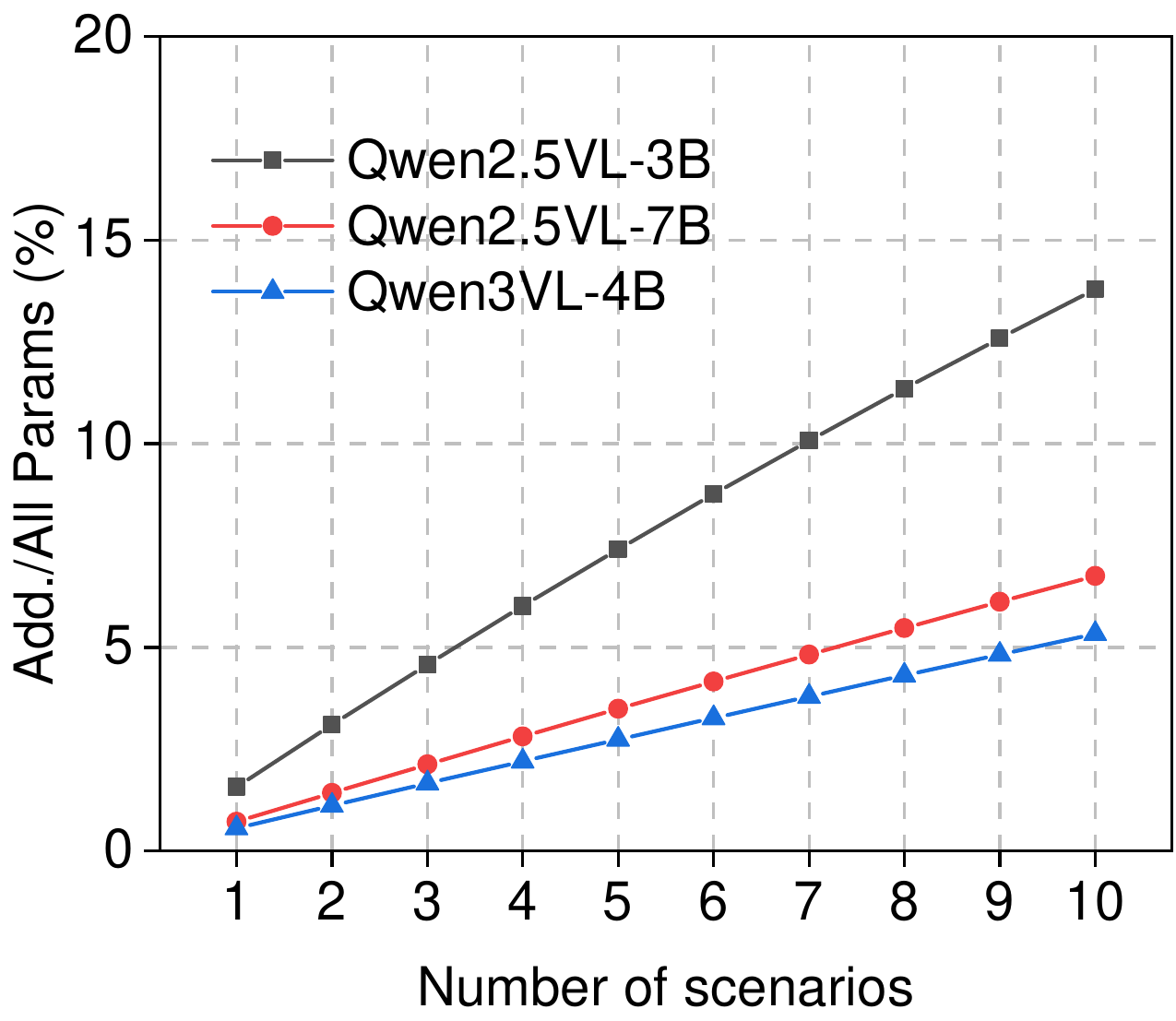}
        \caption{}
        \label{sub_fig:params_tasks}
    \end{subfigure}
    \begin{subfigure}[b]{0.25\textwidth}
        \centering
        \includegraphics[width=\textwidth]{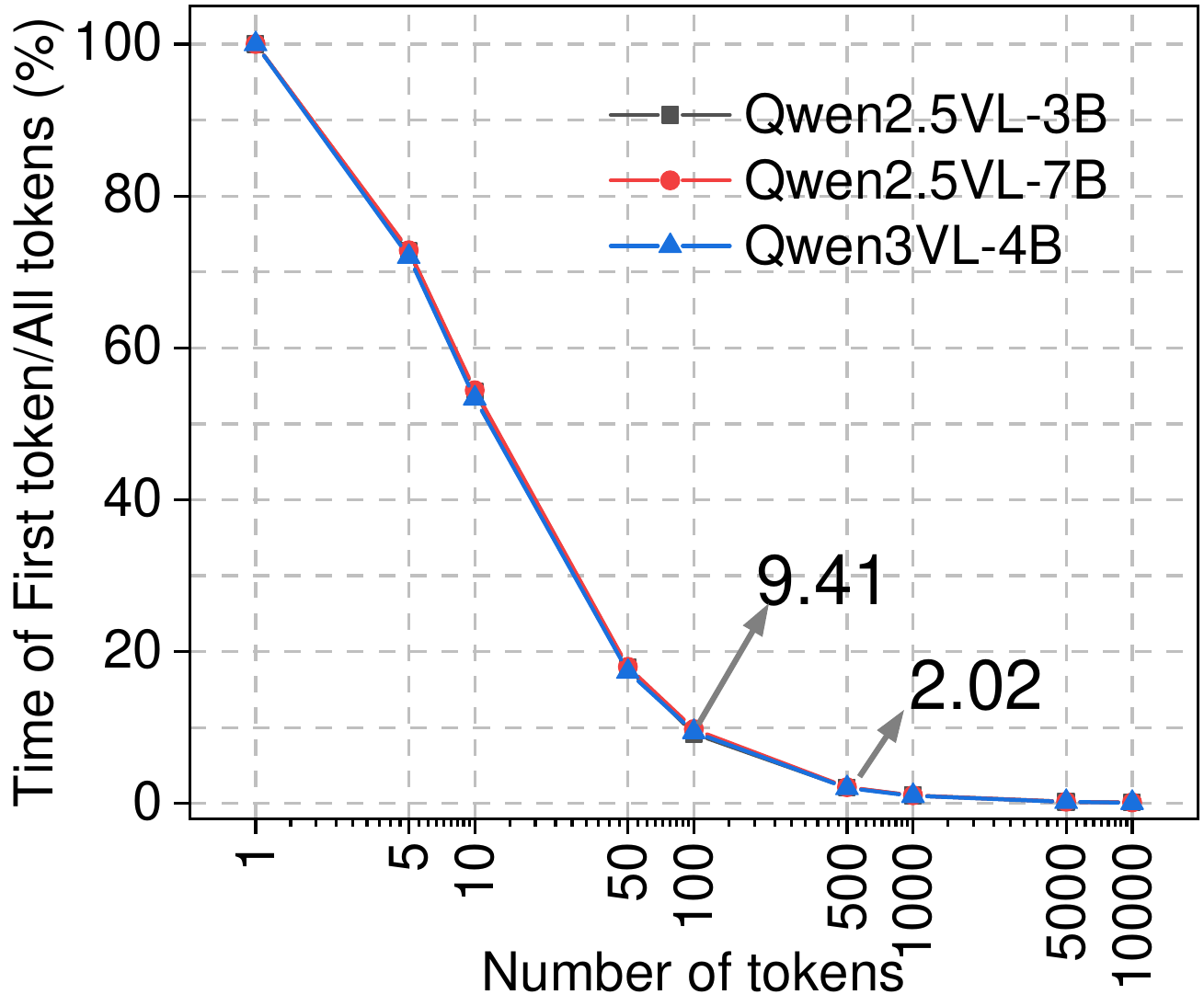}
        \caption{}
        \label{sub_fig:FirstToken_AllTokens}
    \end{subfigure}
    \caption{(a) Curve of the proportion of extra parameters to the total number of parameters. (b) Curve of the proportion of the first token's output time to the total time.}
    \label{fig:param_tokens}
\end{figure*}

The above experiments show that under multiple reasoning modes, adding the VRE module does not bring additional inference burden. Moreover, the parameter count statistics show that after adding four scene branches, the Qwen2.5VL-3B model only increased by 6.4\% in parameters, the Qwem2.5VL-7B model only increased by 2.8\%, and the Qwen3VL-4B model only increased by 2.2\%. We also provided the curve of model parameters increasing with the number of scenes in~Fig.~\ref{sub_fig:params_tasks}. It shows that after adding ten new scenarios, the new parameters of Qwen2.5VL-3B account for less than 15\%, while the other two models have new parameters close to 5\%, proving that \textbf{the parameter growth of VRE is very slow}. In real data streams, similar scenarios occupy the majority, and the main issue in continuous learning is achieving knowledge accumulation of seen scenarios. Therefore, the results show that \textbf{the trend of parameter growth is entirely acceptable}. Fig.~\ref{sub_fig:FirstToken_AllTokens} shows the proportion of TTFT to the total time. The results indicate that when the total number of model output tokens exceeds 100, TTFT accounts for less than 10\% of the total time, and when the total number of tokens exceeds 500, the proportion of TTFT drops to 2\%. Typically, the number of tokens output by the model is greater than 100. For object detection tasks, the number of output tokens is usually greater than 1000. Therefore, \textbf{the additional inference time introduced by the proposed method is negligible}.

\begin{table*}[!t]
\renewcommand{\arraystretch}{1.0}
\centering
\caption{Runtime efficiency comparison. w/o VRE denotes the runtime efficiency of the base model, w/ VRE denotes the runtime efficiency after adding the VRE structure. TTFT (\textit{Time To First Token}) indicates the time to generate the first token, TPOT (\textit{Time Per Output Token}) indicates the generation time for each subsequent token. \textit{Mem.} represents the peak GPU memory usage during the inference. TPOT remains stable after adding VRE, indicating that VRE does not affect inference efficiency.}
\label{tab:complex_batch}
\begin{tabular}{lc|c|cccc}
\toprule
\multicolumn{3}{c|}{Methods} & Params (B) & TTFT (ms) & TPOT (ms) & Mem. (GB)\\
\midrule
\multirow{7}{*}{BS=1, 1$\times$H200}    & \multirow{2}{*}{Qwen2.5VL-3B} & w/o VRE   & 3.75 & 130.10$\pm$1.83  & 35.12$\pm$0.272 & 9.033\\
                                        &                               & w/~~ VRE  & 3.99 & 139.92$\pm$5.02  & 35.34$\pm$0.765 & 9.570\\
\cmidrule(lr){2-7}
                                        & \multirow{2}{*}{Qwen2.5VL-7B} & w/o VRE   & 8.29 & 129.11$\pm$5.32  & 28.21$\pm$0.331 & 17.43\\
                                        &                               & w/~~ VRE  & 8.53 & 139.65$\pm$5.26  & 28.86$\pm$0.475 & 17.92\\
\cmidrule(lr){2-7}
                                        & \multirow{2}{*}{Qwen3VL-4B}   & w/o VRE   & 4.44 & 101.38$\pm$2.77  & 33.04$\pm$0.491 & 9.904\\
                                        &                               & w/~~ VRE  & 4.54 & 113.14$\pm$4.30  & 32.17$\pm$0.854	& 10.15\\
\midrule
\multirow{7}{*}{BS=96, 1$\times$H200}   & \multirow{2}{*}{Qwen2.5VL-3B} & w/o VRE   & 3.75 &  95.26$\pm$0.48  & 8.939$\pm$0.062 & 46.95\\
                                        &                               & w/~~ VRE  & 3.99 & 102.72$\pm$0.71  & 8.974$\pm$0.067 & 52.17\\
\cmidrule(lr){2-7}
                                        & \multirow{2}{*}{Qwen2.5VL-7B} & w/o VRE   & 8.29 & 111.67$\pm$0.58  & 10.40$\pm$0.066 & 64.52\\
                                        &                               & w/~~ VRE  & 8.53 & 124.00$\pm$0.71  & 10.75$\pm$0.055 & 65.17\\
\cmidrule(lr){2-7}
                                        & \multirow{2}{*}{Qwen3VL-4B}   & w/o VRE   & 4.44 &  71.83$\pm$0.89  & 6.980$\pm$0.071 & 37.82\\
                                        &                               & w/~~ VRE  & 4.54 &  78.65$\pm$0.60  & 6.729$\pm$0.055	& 39.80\\
\midrule
\multirow{7}{*}{BS=96, 4$\times$H200}   & \multirow{2}{*}{Qwen2.5VL-3B} & w/o VRE   & 3.75 & 217.83$\pm$29.89 & 20.50$\pm$3.283 & 47.86\\
                                        &                               & w/~~ VRE  & 3.99 & 229.42$\pm$23.40 & 21.59$\pm$3.442 & 53.09\\
\cmidrule(lr){2-7}
                                        & \multirow{2}{*}{Qwen2.5VL-7B} & w/o VRE   & 8.29 & 232.35$\pm$32.33 & 22.10$\pm$3.846 & 65.43\\
                                        &                               & w/~~ VRE  & 8.53 & 252.89$\pm$39.05 & 21.37$\pm$2.285 & 66.08\\
\cmidrule(lr){2-7}
                                        & \multirow{2}{*}{Qwen3VL-4B}   & w/o VRE   & 4.44 & 177.99$\pm$18.38 & 16.04$\pm$2.369 & 38.74\\
                                        &                               & w/~~ VRE  & 4.54 & 218.41$\pm$21.81 & 16.33$\pm$2.137 & 40.72\\
\bottomrule
\end{tabular}
\end{table*}

\begin{figure*}[!t]
    \centering
    \begin{subfigure}[b]{0.495\textwidth}
        \centering
        \includegraphics[width=\textwidth]{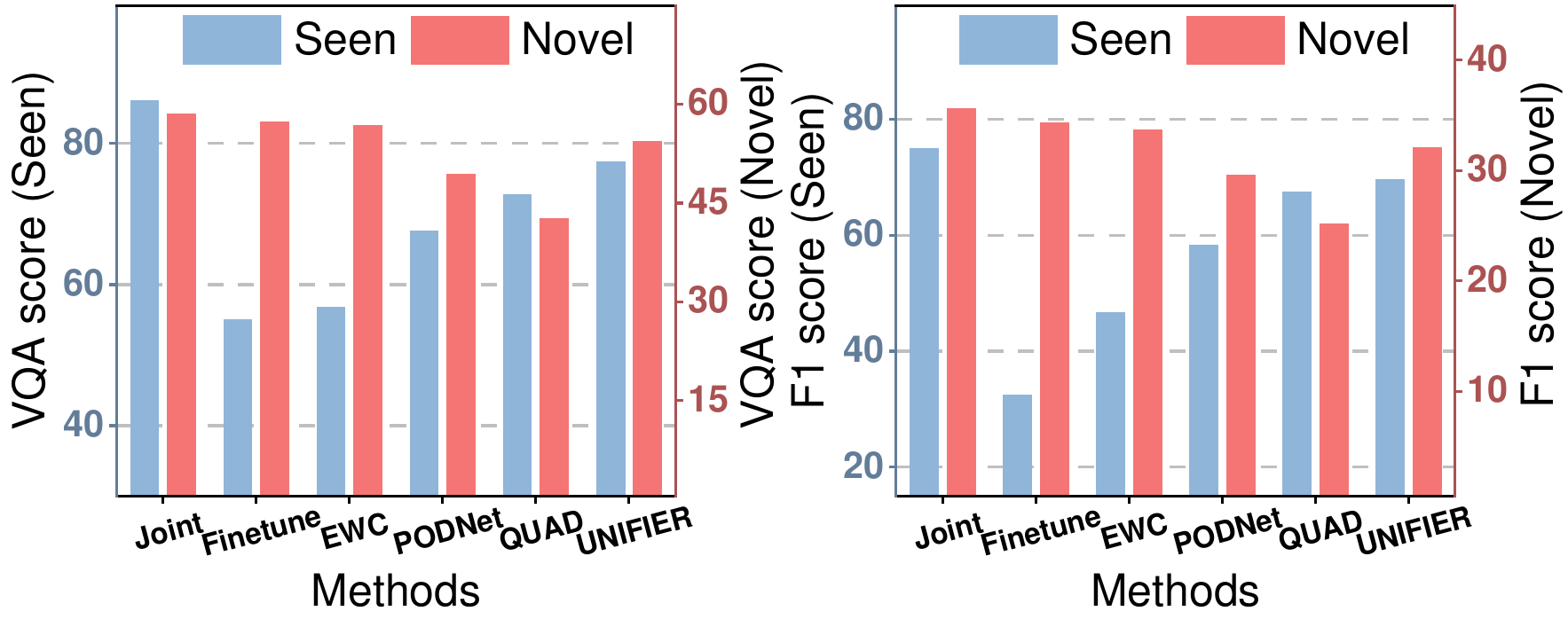}
        \caption{Indoor$\rightarrow$Low altitude}
        \label{sub_fig:indoor2low}
    \end{subfigure}
    \begin{subfigure}[b]{0.495\textwidth}
        \centering
        \includegraphics[width=\textwidth]{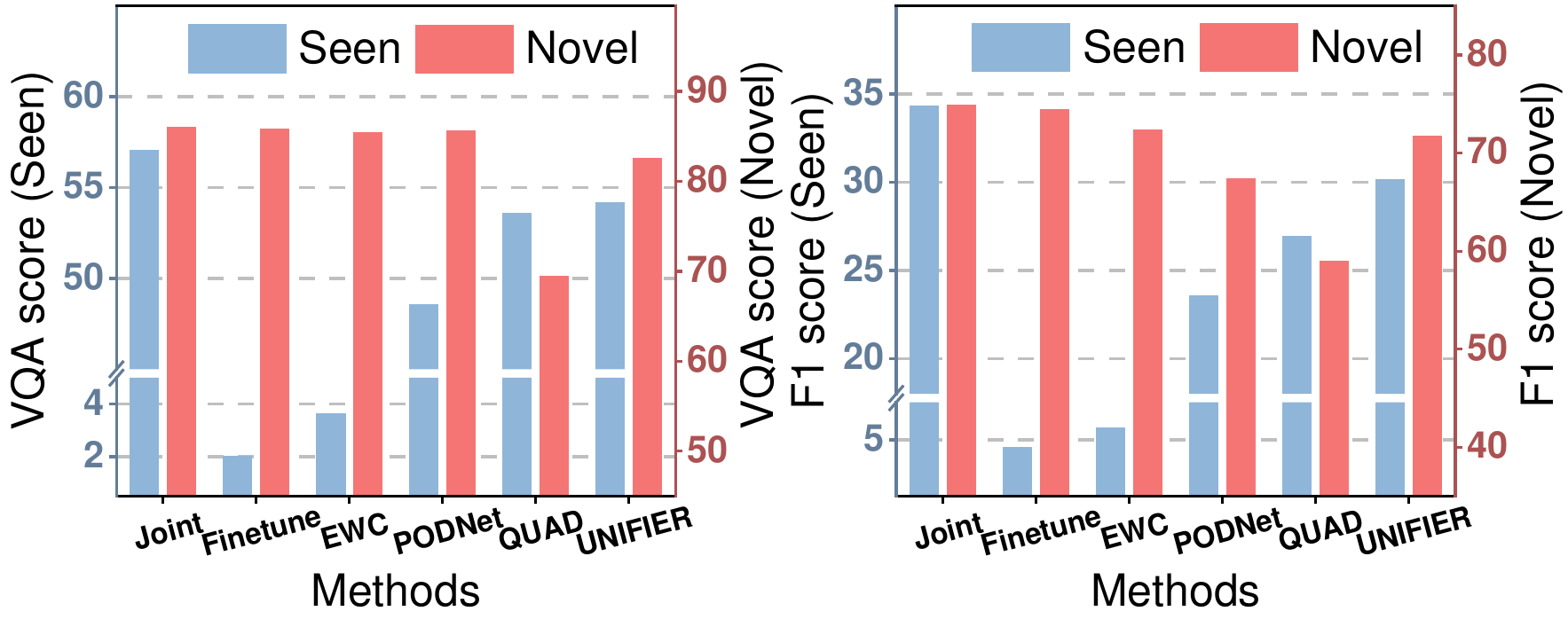}
        \caption{Low altitude$\rightarrow$Indoor}
        \label{sub_fig:low2indoor}
    \end{subfigure}
    \\
    \centering
    \begin{subfigure}[b]{0.495\textwidth}
        \centering
        \includegraphics[width=\textwidth]{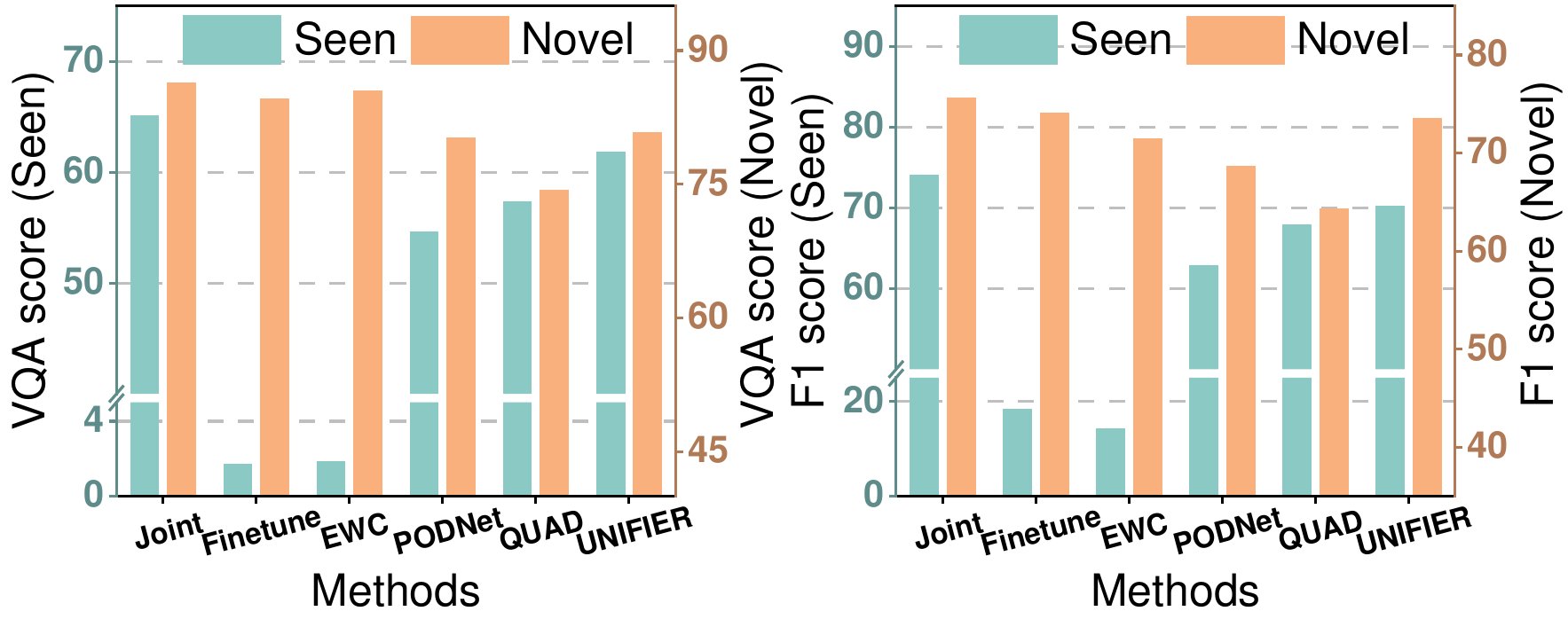}
        \caption{High altitude$\rightarrow$Indoor}
        \label{sub_fig:high2indoor}
    \end{subfigure}
    \begin{subfigure}[b]{0.495\textwidth}
        \centering
        \includegraphics[width=\textwidth]{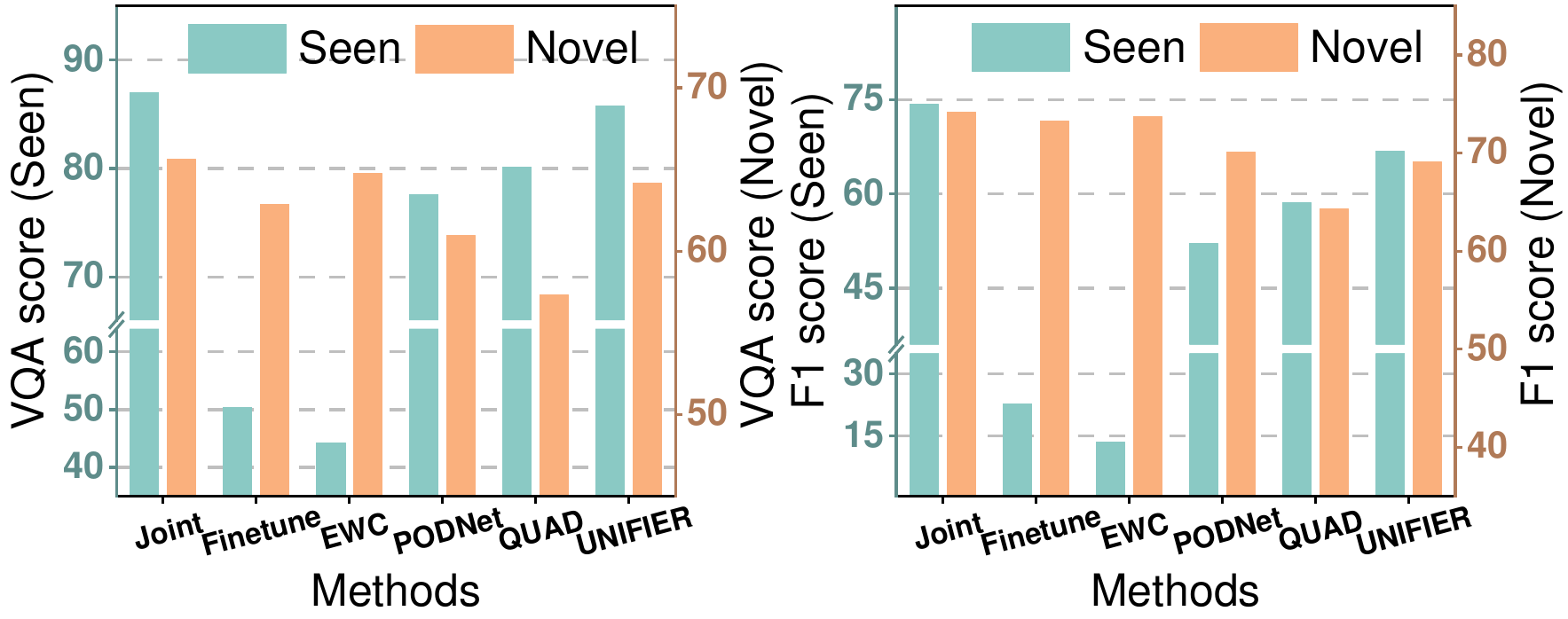}
        \caption{Indoor$\rightarrow$High altitude}
        \label{sub_fig:indoor2high}
    \end{subfigure}
    \\
    \centering
    \begin{subfigure}[b]{0.495\textwidth}
        \centering
        \includegraphics[width=\textwidth]{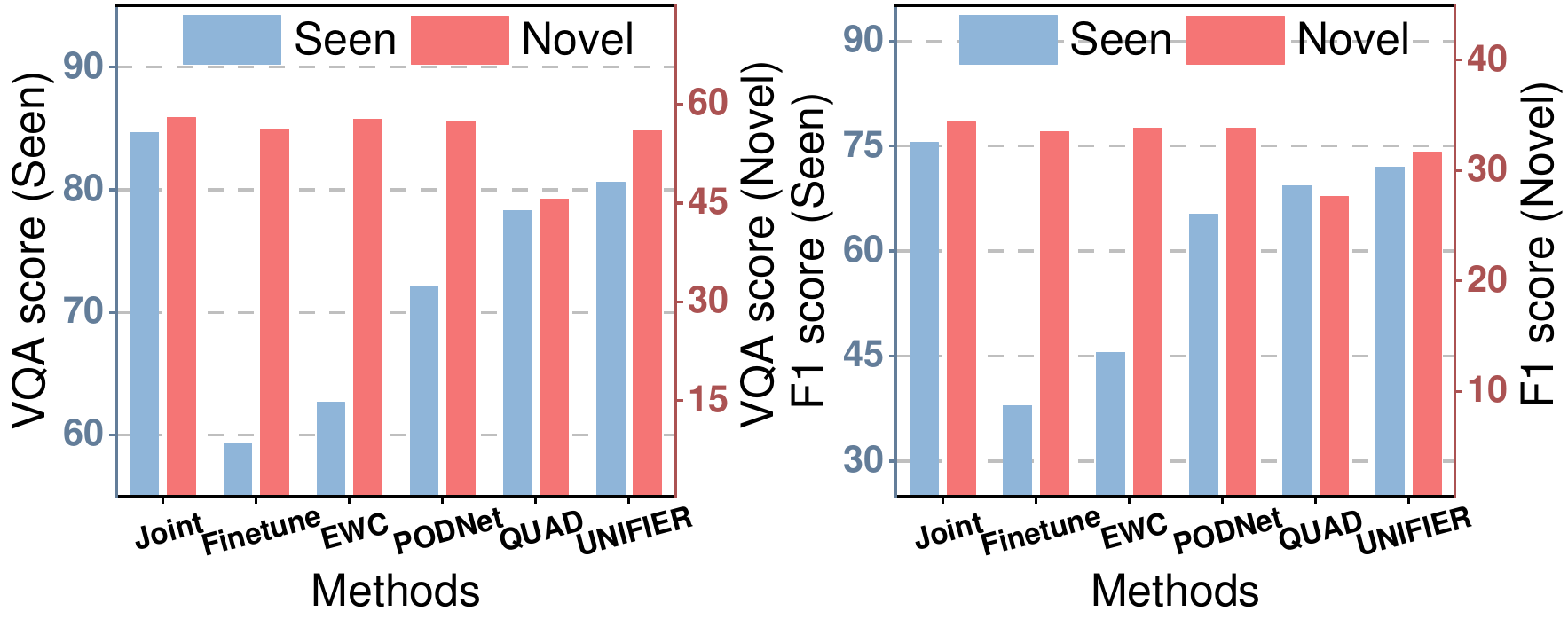}
        \caption{Underwater$\rightarrow$Low altitude}
        \label{sub_fig:underwater2low}
    \end{subfigure}
    \begin{subfigure}[b]{0.495\textwidth}
        \centering
        \includegraphics[width=\textwidth]{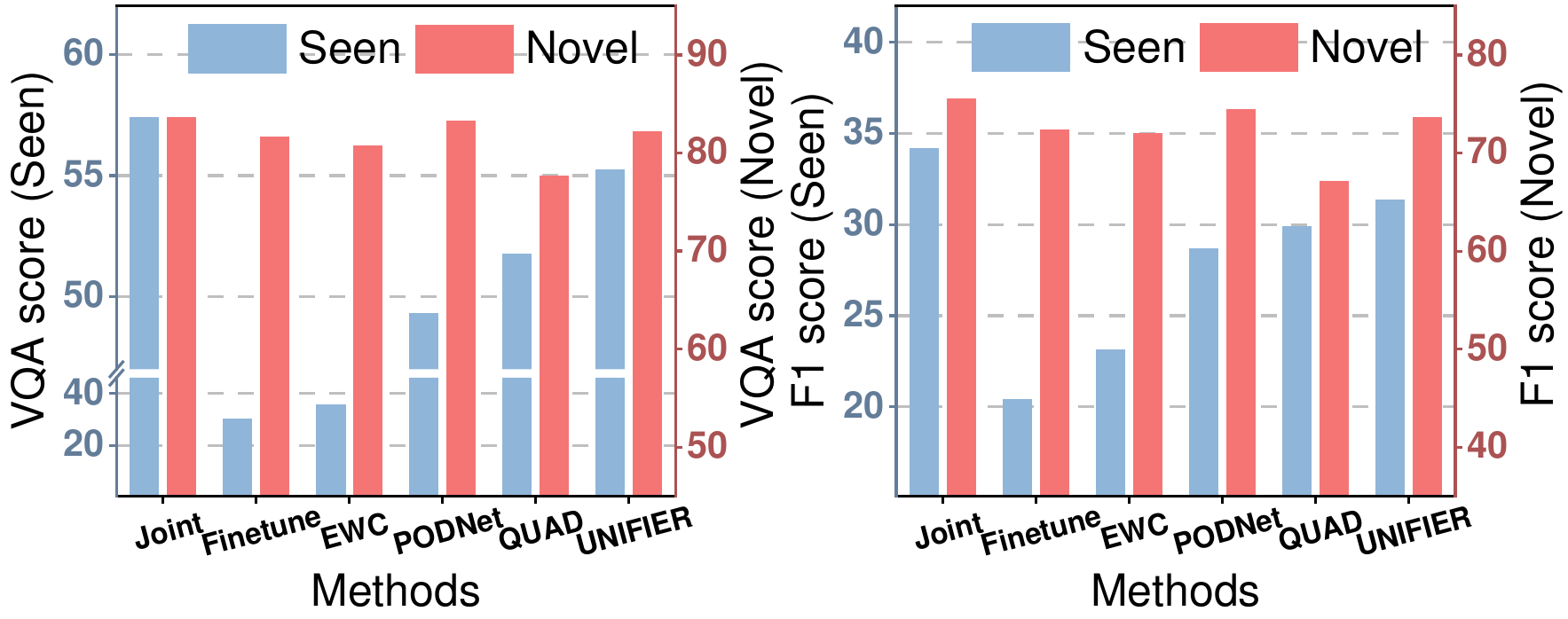}
        \caption{Low altitude$\rightarrow$Underwater}
        \label{sub_fig:low2underwater}
    \end{subfigure}
    \\
    \centering
    \begin{subfigure}[b]{0.495\textwidth}
        \centering
        \includegraphics[width=\textwidth]{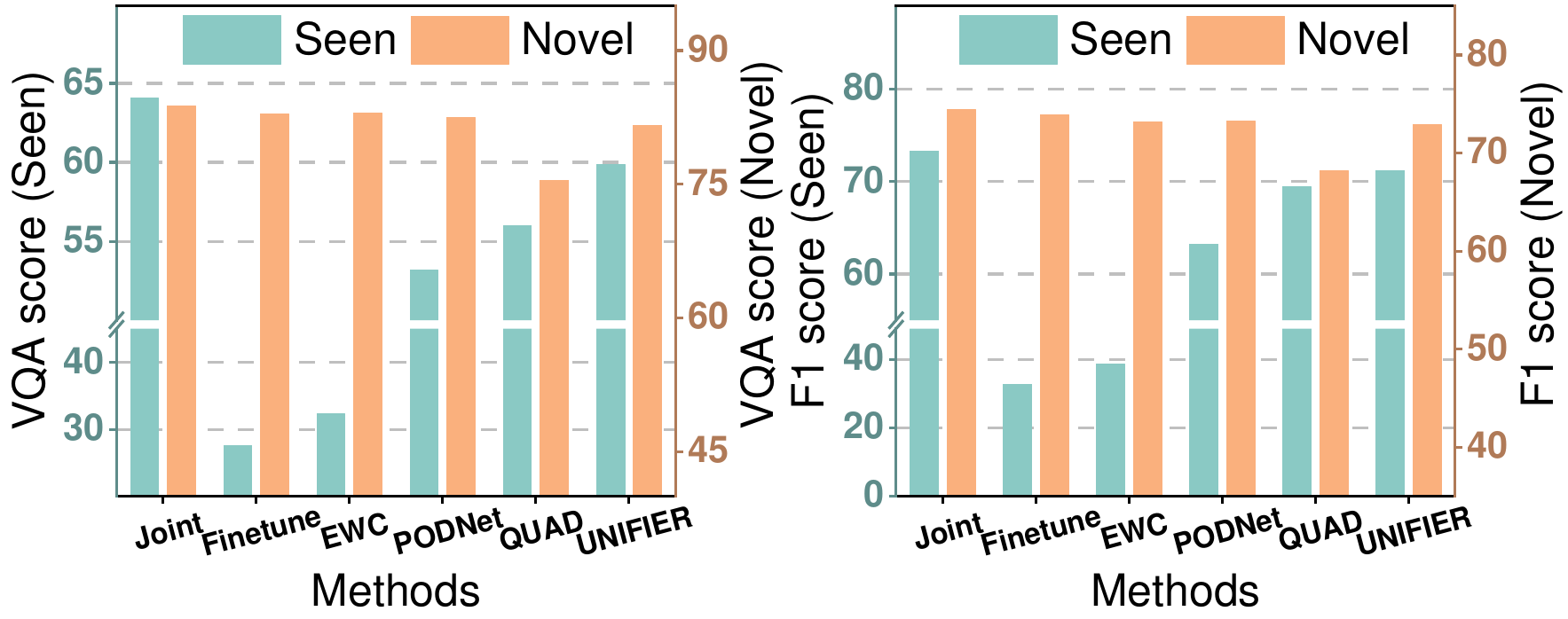}
        \caption{High altitude$\rightarrow$Underwater}
        \label{sub_fig:high2underwater}
    \end{subfigure}
    \begin{subfigure}[b]{0.495\textwidth}
        \centering
        \includegraphics[width=\textwidth]{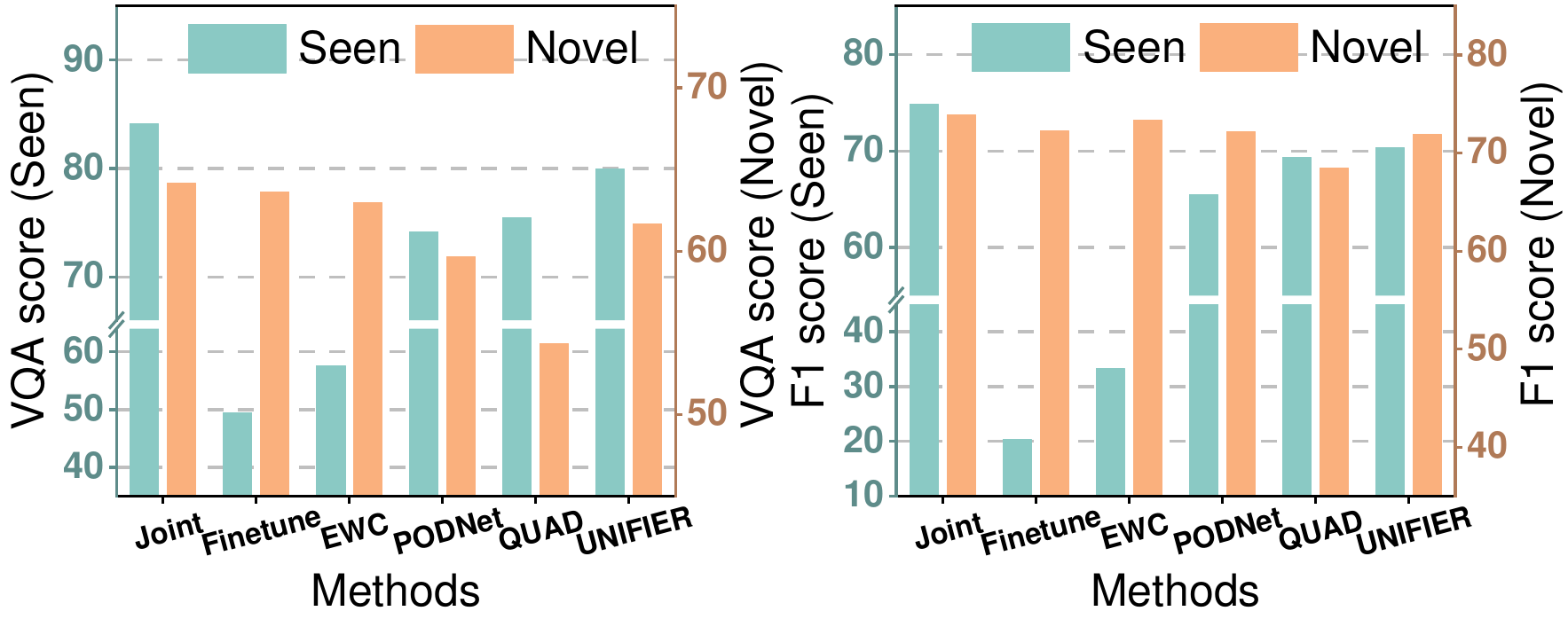}
        \caption{Underwater$\rightarrow$High altitude}
        \label{sub_fig:underwater2high}
    \end{subfigure}
    \caption{More experimental results of performance comparison in scenario alteration.}
    \label{sup_fig:all_alteration}
\end{figure*}
\section{Comparison of different upper bounds} \label{sup_sec:upper}
In this section, we perform additional experiments with Qwen2.5-VL-7B and Qwen3-VL-4B to eliminate the influence of different model types. All models are trained with the same hyper-parameters by 20 epochs. The comparison is shown in~Fig.~\ref{fig:qwen_joint_cmp}. The results indicate that, although there is a slight disparity in the performance of zero-shot using these foundation models, their fine-tuned performance is comparable.

\section{Further experiments with Qwen3-VL} \label{sup_sec:qwen3vl}
To further investigate the generalization ability of the \textsc{Unifier} across various foundation models, we performed additional experiments with \textbf{Qwen3-VL}. The results for the setting of 10 steps are presented in~Fig.~\ref{fig:qwen3_comparison_10steps}. We compare the proposed method with QUAD, PODNet, and the lower bound Finetune, with the last-step VQA score of \textsc{Unifier} outperforming the runner up by 2.08\%$\sim$6.41\%, and the last-step F{\footnotesize 1} score outperforming the runner up by 3.94\%$\sim$9.37\%. It demonstrates that the \textsc{Unifier} is still effective with Qwen3-VL.

\section{Comparison experiments} \label{sup_sec:complete_cmp}
In this section, we provide the results of complete comparison experiments, including 5 steps in~Fig.~\ref{fig:comparison_5steps} and 20 steps in~Fig.~\ref{fig:comparison_20steps}. The performance gap is annotated at the end of each curve. This indicates that the proposed method can effectively prevent catastrophic forgetting in various task settings.

\section{Scenario alteration experiments} \label{sup_sec:scenario_alteration}
In this section, we provide more experiments on scenario alteration, including Indoor$\leftrightarrow$Low altitude (Fig.~\ref{sub_fig:indoor2low} and~\ref{sub_fig:low2indoor}), High altitude$\leftrightarrow$Indoor(Fig.~\ref{sub_fig:high2indoor} and~\ref{sub_fig:indoor2high}), Underwater$\leftrightarrow$Low altitude (Fig.~\ref{sub_fig:underwater2low} and~\ref{sub_fig:low2underwater}) and High altitude$\leftrightarrow$Underwater (Fig.~\ref{sub_fig:high2underwater} and~\ref{sub_fig:underwater2high}). Experimental results demonstrate that the \textsc{Unifier} achieves better performance in continual learning across scenarios than comparison methods.

\section{Learning order} \label{sup_sec:order}
Although we used the same random seed (1993) to determine the learning order in each task setting, the learning order varies across task settings due to different T values. As shown in the Tab.~\ref{tab:order}, the three settings of $T$=5, 10, and 20 already encompass various learning orders. The proposed method demonstrates excellent performance across all three settings, \textbf{eliminating the influence of learning order and proving the effectiveness of the proposed method}. Moreover, we conduct multiple scenario alteration experiments (Fig.~\ref{fig:crs} and~\ref{sup_fig:all_alteration}), which also demonstrate that the effectiveness of \textbf{the proposed method is independent of the learning order}.

\begin{table*}[!t]
\renewcommand{\arraystretch}{1.0}
\centering
\caption{Learning order. \textit{Current} represents the current task scenario. The table is ordered from left to right according to the learning sequence.}
\label{tab:order}
\resizebox{1\linewidth}{!}{
\begin{tabular}{lc|c|c|c|c|c|c|c|c|c|c|c|c|c|c|c|c|c|c|c}
\toprule
\multirow{2}{*}{$T$=5} & \multicolumn{4}{c|}{1} & \multicolumn{4}{c|}{2} & \multicolumn{4}{c|}{3} & \multicolumn{4}{c|}{4} & \multicolumn{4}{c}{5}\\
\cmidrule(lr){2-21}
 & \multicolumn{4}{c|}{Current} & \multicolumn{4}{c|}{\textcolor{red!50}{High altitude}} & \multicolumn{4}{c|}{\textcolor{orange!50}{Low altitude}} & \multicolumn{4}{c|}{\textcolor{SeaGreen!50}{Indoor}} & \multicolumn{4}{c}{\textcolor{blue!50}{Underwater}}\\
\midrule
\multirow{2}{*}{$T$=10} & \multicolumn{2}{c|}{1} & \multicolumn{2}{c|}{2} & \multicolumn{2}{c|}{3} & \multicolumn{2}{c|}{4} & \multicolumn{2}{c}{5} & \multicolumn{2}{c|}{6} & \multicolumn{2}{c|}{7} & \multicolumn{2}{c|}{8} & \multicolumn{2}{c|}{9} & \multicolumn{2}{c}{10}\\
\cmidrule(lr){2-21}
 & \multicolumn{2}{c|}{Cur.} & \multicolumn{2}{c|}{\textcolor{blue!50}{Und.}} & \multicolumn{2}{c|}{\textcolor{blue!50}{Und.}} & \multicolumn{2}{c|}{\textcolor{orange!50}{Low.}} & \multicolumn{2}{c|}{\textcolor{orange!50}{Low.}} & \multicolumn{2}{c|}{\textcolor{red!50}{Hig.}} & \multicolumn{2}{c|}{\textcolor{blue!50}{Und.}} & \multicolumn{2}{c|}{\textcolor{orange!50}{Low.}} & \multicolumn{2}{c|}{\textcolor{SeaGreen!50}{Ind.}} & \multicolumn{2}{c}{\textcolor{SeaGreen!50}{Ind.}}\\
\midrule
\multirow{2}{*}{$T$=20} & 1 & 2 & 3 & 4 & 5 & 6 & 7 & 8 & 9 & 10 & 11 & 12 & 13 & 14 & 15 & 16 & 17 & 18 & 19 & 20\\
\cmidrule(lr){2-21}
 & C. & \textcolor{orange!50}{L.} & \textcolor{blue!50}{U.} & \textcolor{blue!50}{U.} & \textcolor{blue!50}{U.} & \textcolor{orange!50}{L.} & \textcolor{orange!50}{L.} & \textcolor{SeaGreen!50}{I.} & \textcolor{red!50}{H.} & \textcolor{blue!50}{U.} & \textcolor{SeaGreen!50}{I.} & \textcolor{SeaGreen!50}{I.} & \textcolor{orange!50}{L.} & \textcolor{blue!50}{U.} & \textcolor{red!50}{H.} & \textcolor{red!50}{H.} & \textcolor{orange!50}{L.} & \textcolor{SeaGreen!50}{I.} & \textcolor{SeaGreen!50}{I.} & \textcolor{red!50}{H.}\\ 
\bottomrule
\end{tabular}
}
\end{table*}

\end{appendices}

\end{document}